\RequirePackage[svgnames]{xcolor}
\documentclass[11pt,letterpaper,onecolumn]{mystyle}

\usepackage{mathtools}          %
\usepackage{mathrsfs}           %

\usepackage{xspace}
\usepackage{wrapfig}
\usepackage{needspace}
\usepackage{float}              %
\usepackage{soul}               %
\usepackage{enumitem}           %
\usepackage{tabularx}
\usepackage{pifont}             %
\usepackage{soul}

\usepackage{listings}

\usepackage{tikz}
\usepackage{pgfplots}
\usetikzlibrary{arrows.meta,backgrounds}
\pgfplotsset{compat=1.18}
\usepgfplotslibrary{groupplots}

\providecommand{\methodname}{}
\renewcommand{\methodname}{\textsc{SpaceFlow}\xspace}

\makeatletter
\DeclareRobustCommand{\onedot}{\futurelet\@let@token\@onedot}
\def\@onedot{\ifx\@let@token .\else.\null\fi\xspace}
\makeatother

\newcommand{\eg}{e.g\onedot}

\newcommand{\PAR}[1]{\vskip4pt \noindent{\bf #1~}}

\definecolor{figbg}{HTML}{EEF7F1}
\definecolor{highblue}{HTML}{ADC3F0}
\definecolor{loworange}{HTML}{FFA36F}
\definecolor{deltapurple}{HTML}{8A35FF}
\definecolor{grey}{gray}{0.45}

\definecolor{wincol}{HTML}{8BCF9C}
\definecolor{tiecol}{HTML}{D0D0D0}
\definecolor{losscol}{HTML}{E6A09B}

\colorlet{windim}{wincol!35!white}
\colorlet{tiedim}{tiecol!45!white}
\colorlet{lossdim}{losscol!35!white}

\definecolor{promptbg}{RGB}{246,246,248}
\definecolor{promptborder}{RGB}{205,207,213}
\definecolor{promptaccent}{RGB}{174,91,65}

\definecolor{orange}{HTML}{E2920A}
\definecolor{grey}{HTML}{DBDDDF}

\lstdefinestyle{vlmprompt}{
    basicstyle=\ttfamily\small,
    breaklines=true,
    columns=fullflexible,
    keepspaces=true,
    showstringspaces=false,
    breakindent=0pt,
    prebreak=,
    postbreak=,
    breakatwhitespace=true,
    escapeinside={(*@}{@*)}
}

\newtcblisting{promptbox}{
    enhanced,
    listing only,
    listing engine=listings,
    listing options={style=vlmprompt},
    colback=promptbg,
    colframe=promptborder,
    boxrule=0.45pt,
    arc=2mm,
    outer arc=2mm,
    boxsep=0pt,
    left=2.5mm,
    right=2.5mm,
    top=2.0mm,
    bottom=2.0mm,
    width=\linewidth
}

\newcommand{\prompttag}[1]{%
    \begingroup
    \setlength{\fboxsep}{1.2pt}%
    \colorbox{black!10}{%
        \strut\textcolor{black!75}{\bfseries\ttfamily #1}%
    }%
    \endgroup
}

\newcommand{\prompthighlight}[1]{%
    \textcolor{promptaccent}{\bfseries\ttfamily #1}%
}

\usepackage{tikz} %

\DeclareRobustCommand{\roundedhl}[2]{%
  \tikz[baseline=(word.base)]{%
    \node[
      fill=#1,
      rounded corners=2pt,
      inner xsep=2pt,
      inner ysep=1pt,
      text height=1.5ex,
      text depth=0.5pt,
      outer sep=0pt
    ] (word) {#2};
  }%
}

\title{\methodname: Locally Controllable 3D Generation}

\author[1\,*]{Neil De La Fuente}
\author[1\,*]{Joan Lafuente}
\author[1\,*]{Mukhammadali Sayfiddinov}
\author[1\,*]{Felicia Scharitzer}
\author[1,3]{Marc Pollefeys}
\author[1]{Ata Çelen}
\author[2\,\dagger]{Sayan Deb Sarkar}
\author[1\,\dagger]{Elisabetta Fedele}

\affiliation[1]{ETH Z\"urich}
\affiliation[2]{Stanford University}
\affiliation[3]{Microsoft}

\contribution[*]{Equal contribution (ordered alphabetically)}
\contribution[\dagger]{Equal supervision}

\abstract{%
Current 3D generation methods lack explicit local control: geometric adherence is often defined by a global control strength, and appearance cannot be specified locally. We present \methodname, a training-free pipeline for locally controllable 3D generation from text descriptions and a collection of geometric primitives. Each primitive serves as a proxy for an object part and is assigned a local control level, enabling users to specify whether regions should strictly follow the input shape or allow generative completion. During structure generation, we enforce these spatial constraints within the generative flow process. For appearance synthesis, the generated structure is segmented and matched to the primitives. Each generated part is conditioned only on its assigned text or image cue, thereby limiting cross-part leakage. Regional geometry metrics demonstrate that \methodname preserves the specified geometry in high-control regions and enables plausible shape variation in low-control areas. A user study further indicates that the resulting balance between geometric fidelity and generative freedom remains competitive in overall quality. When evaluating appearance on fixed geometry, text-conditioned routing achieves state-of-the-art prompt faithfulness and color/material accuracy. Qualitative results additionally show localized routing of image cues. The project page is available at \textbf{\href{https://spaceflow3d.github.io}{https://spaceflow3d.github.io}}.
}

\teaser{%
  \begin{center}
    \includegraphics[width=\textwidth]{figs/teaser_final_v2.pdf}
    \captionof{figure}{\textbf{Overview of Local Geometric and Appearance Control Capabilities.}
    \textit{Top left (Geometric Control):} \methodname assigns different control strengths to editable geometric inputs: \roundedhl{orange!40}{orange} regions (strong geometric control) preserve the specified geometry, whereas \roundedhl{grey}{grey} regions (weak geometric control) allow prompt-driven completion.
    \textit{Top right (Appearance Control):} Local text or image cues are bound to their target primitives, localizing distinct materials on specified regions, and combining visual references across different parts.
    \textit{Bottom (Joint Geometric and Appearance Control):} Additional examples show localized shape and appearance control.}
    \label{fig:teaser}
  \end{center}
}
\begin{document}
\maketitle

\section{Introduction}

Recent text- and image-conditioned 3D generative models can synthesize high-quality textured assets across increasingly diverse object categories~\cite{10.1145/3658146,hunyuan3d22025tencent,Xiang_2025_CVPR, 10.1007/978-3-031-73235-5_1, poole2022dreamfusion, xu2024instantmesh, jun2023shapegeneratingconditional3d, Liu_2024_CVPR}. These advances dramatically lower the barrier from a visual idea or reference to an initial 3D candidate asset, with applications in games, immersive media, and digital content creation~\cite{Lin_2023_CVPR, Chen_2024_CVPR, Tang_2023_ICCV, chen2023fantasia3d, sun2024dreamcraft3d}. 
Although the quality of assets generated from a single image or text prompt continues to improve, integrating such generators into artistic workflows requires more interactive and controllable guidance mechanisms~\cite{Li_2025_CVPR,sanghi2023sketchashapezeroshotsketchto3dshape, instructnerf2023, Li20223DDesignerTP}. 

In practice, asset creation is iterative~\cite{li2025meshpad, Yuan_2026_CVPR} and part-specific~\cite{Chen2025PartGen, mo2019structurenethierarchicalgraphnetworks, 10.1145/3528223.3530084, koo2023salad, jones2020shapeAssembly}, rather than one-shot: creators establish coarse structure, inspect generated candidates, retain some design decisions, and revise others~\cite{Ctrl-Room}. Some parts may have to follow an intended shape, whereas others are left under-constrained, relying on the generative prior~\cite{nakayama2023difffacto}.
Supporting this process requires control not only over \emph{what} is generated, but also over \emph{where} each instruction applies and \emph{how strictly} it should be followed~\cite{Chen2025PartGen,cheng2024progressive3d,fedele2026spacecontrol}.

To support this need, recent methods augment general-purpose 3D generators~\cite{Chen_2026_CVPR,Xiang_2025_CVPR} with controllable geometric or appearance guidance~\cite{fedele2026spacecontrol,sayandsarkar_2025_guideflow3d}. On one hand, proxy-guided approaches~\cite{fedele2026spacecontrol,10.1145/3641519.3657461} enable \textit{geometric} conditioning by guiding 3D generation with an input 3D shape. However, they apply this conditioning \textit{globally}, such that all regions are equally constrained by the input geometry. On the other hand, other approaches introduce test-time, part-aware guidance for \textit{appearance} transfer~\cite{sayandsarkar_2025_guideflow3d} or localized editing of implicit representations~\cite{Song_2023_ICCV}, while leaving geometry generation unconstrained or operating post-hoc on pre-existing representations. Appearance guidance is also limited to a single global condition, preventing different cues from being assigned to different regions. Existing approaches therefore lack a unified mechanism for local control over both geometry and appearance. 

We propose \methodname, a unified approach for local control of both geometry and appearance during 3D generation, as shown in Fig.~\ref{fig:teaser}. In practice, the user specifies the geometry and scene layout as an explicit geometric input, decomposed into an arbitrary set of local parts that can be addressed independently. For geometric conditioning, we assign a local control level to each part: higher levels enforce closer adherence to the specified geometry, whereas lower levels leave more freedom to the generative prior. The formulation is agnostic to how individual parts are represented, ranging from coarse geometric primitives to detailed meshes. Following~\cite{fedele2026spacecontrol}, we represent parts as superquadric primitives~\cite{Paschalidou2019CVPR}, which provide a compact, expressive, and easily editable representation. These can either be defined manually or initialized from existing geometry using shape decomposition methods~\cite{fedele2025superdec, superflex}.

We use the same part decomposition for appearance conditioning. Specifically, the user can assign a distinct appearance condition to each part, allowing different regions to be controlled independently. We achieve this by restricting cross-attention to the condition associated with each part, while retaining global self-attention across all tokens. This enables distinct local appearances while preserving coherence and smooth blending across the generated asset.

We evaluate the geometric and appearance components separately and demonstrate their combined capabilities on diverse multipart assets. Regional geometry metrics show that \methodname preserves strongly constrained regions more faithfully than weak global conditioning, while allowing greater variation in weakly constrained regions than strong global conditioning. A user study further supports this balance between geometric fidelity and generative freedom. On fixed geometry, our VLM-as-a-judge evaluation of localized text-conditioned appearance shows higher prompt faithfulness and color/material accuracy than the evaluated TRELLIS and GuideFlow3D variants~\cite{Xiang_2025_CVPR,sayandsarkar_2025_guideflow3d}. We additionally demonstrate localized image-conditioned appearance through qualitative results.

In summary, our technical contributions are:
\begin{itemize}
    \item An approach to locally control the strength of \textit{geometric} conditioning for 3D generation.
    \item An approach to locally condition the \textit{appearance} of generated 3D assets.
    \item A \textit{unified framework} combining both approaches to enable local geometric and appearance control for 3D generation.
\end{itemize}

\section{Related Work}

\PAR{3D Asset Generation.}
\label{sec:rw_3d_generation} Generative 3D modeling has advanced rapidly in output fidelity and representation flexibility, from explicit point clouds~\cite{nichol2022pointegenerating3dpoint} and implicit functions~\cite{jun2023shapegeneratingconditional3d} to compact latent-space approaches~\cite{NEURIPS2022_40e56dab, 10.1145/3592442, Cheng_2023_CVPR}. Recently, large-scale generators have achieved strong fidelity: CLAY~\cite{10.1145/3658146} leverages multi-resolution latent spaces, Hunyuan3D~\cite{hunyuan3d22025tencent} scales diffusion for high-resolution textured assets, SAM 3D~\cite{Chen_2026_CVPR} enables single-image 3D reconstruction, and TRELLIS~\cite{Xiang_2025_CVPR} introduces structured latents factoring geometry and appearance on a sparse voxel grid. However, these models condition on a global text or image prompt and lack fine-grained, localized part control.

\PAR{Spatially Controllable 3D Generation.} 
\label{sec:rw_spatial_control} Spatially controllable generation conditions on explicit 3D inputs, such as voxels, primitives, proxy shapes, or meshes, rather than ambiguous text or images~\cite{Cheng_2023_CVPR,Metzer_2023_CVPR,10.1145/3641519.3657425,fedele2026spacecontrol}. Existing frameworks either fine-tune generative models to accept structural conditions~\cite{10.1145/3641519.3657461,10.1145/3641519.3657425,li2025cmd} at additional training cost, or enforce test-time constraints via optimization and latent projection~\cite{Metzer_2023_CVPR,fedele2026spacecontrol} without retraining. Closest to our structural setting, SpaceControl~\cite{fedele2026spacecontrol} injects user-specified geometry into a pretrained generator's latent space with a control strength that trades geometric fidelity for generative realism. However, since this trade-off is applied uniformly across the entire asset, it cannot handle scenarios where designated regions must strictly preserve input geometry while others are left unconstrained for generative completion~\cite{fedele2026spacecontrol}.

\PAR{Appearance Control for 3D Assets.} 
\label{sec:rw_appearance_control} Appearance control has been widely explored across mesh texturing~\cite{text2mesh, 10.1145/3588432.3591503, perla2024easitex,zeng2024paint3d, Cohen-Bar_2025_CVPR}, radiance fields, and 3D Gaussian splats~\cite{liu2023stylerf, liu2023stylegaussian, 10.1007/978-3-031-85187-2_15}. These methods primarily formulate appearance control as post-hoc texturing or stylization of an existing 3D asset, whereas our focus is localized appearance conditioning during the generative flow process~\cite{text2mesh,10.1145/3588432.3591503,liu2023stylerf,liu2023stylegaussian}.
GuideFlow3D~\cite{sayandsarkar_2025_guideflow3d} provided training-free guidance of a pretrained rectified-flow model using a self-similarity loss for robust appearance transfer. However, it transfers a single appearance source globally across the generated asset. In contrast, \methodname routes independent visual conditions to specific semantic parts, eliminating cross-part leakage.

\PAR{Part-Aware Generation and Editing.}
\label{sec:rw_part_primitive_editing}A complementary line of work introduces part-aware generators, which incorporate explicit part structure through hierarchical graphs and programs~\cite{mo2019structurenethierarchicalgraphnetworks,jones2020shapeAssembly}, implicit or radiance-field representations, and diffusion over point or latent spaces~\cite{10.1145/3528223.3530084,Tertikas2023CVPR,koo2023salad,nakayama2023difffacto}, including recent part-level reconstruction and synthesis methods~\cite{Chen2025PartGen,li2025cmd,nam2025parte,NEURIPS2025_32cc6132,slim2026composecompositionalsynthesisediting}. In contrast, \methodname does not require the generated asset to be represented as parts; it uses primitives and part assignments only as localized control signals for a pretrained generator.
\vspace{-0.1cm}

\section{Method}

\subsection{Problem Formulation}
\label{sec:problem formulation}

\begin{wrapfigure}{r}{0.48\textwidth}
\vspace{-10pt}
    \centering
    \includegraphics[width=\linewidth]{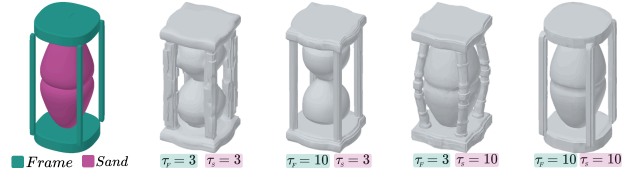}
    \caption{\textbf{Local spatial control.} Starting from the same "hourglass" superquadric primitive set and prompt, high control \((\tau\!=\!10)\) preserves the corresponding part's input geometry, while low control \((\tau\!=\!3)\) allows generative variation. The four settings showcase independent control of the frame and sand.}
    \label{fig:local-spatial-control}
\end{wrapfigure}

Given a text prompt $p$ and a user-edited set of deformable primitives $\mathcal{S}\!=\!\{S_i\}_{i=1}^{N}$, our goal is to generate a textured 3D asset $X\!=\!(G,T)$ whose geometry and appearance adhere to localized spatial controls. Each primitive $S_i$ is parameterized by a control strength $\tau_i\!\in\!\{\tau_{\text{low}}, \tau_{\text{high}}\}$ and an optional appearance cue $a_i\!\in\!\mathcal{A}$ such as an image or text prompt. The control strengths $\tau_i$ determine the local trade-off between geometric fidelity and generative freedom. $\tau_{\text{high}}$ promotes strict adherence to the input, whereas $\tau_{\text{low}}$ relaxes the spatial constraints to allow for flexible modeling using a generative prior. As shown in Fig.~\ref{fig:local-spatial-control}, assigning control independently per primitive enables the trade-off to vary spatially across the asset. Furthermore, when an appearance cue $a_i$ is available, it only influences the appearance of its corresponding primitive region $S_i$.

\begin{figure}[t]
  \centering
  \includegraphics[width=\textwidth]{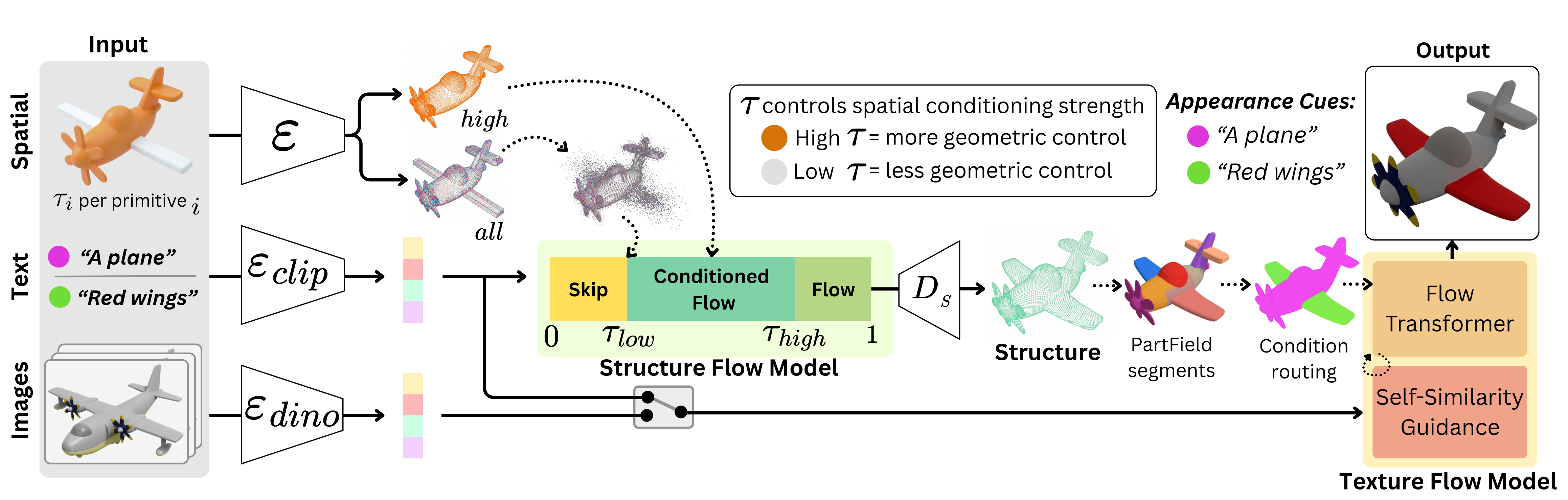}
  \vspace{-15pt}
  \caption{\textbf{\methodname pipeline.} Given editable geometric controls with local geometric conditioning strengths $\tau_i$, \methodname encodes the high-control regions and injects them into the sparse-structure flow process over a controlled time interval. The resulting structure is segmented with PartField~\cite{liu2025partfield}, then per-part text or image conditions are routed to each part, and the latent is finally passed to an appearance flow model and refined with self-similarity guidance, producing a 3D asset with localized geometric and appearance control.}
  \label{fig:overview}
\end{figure}

\subsection{Preliminaries}

\noindent\textbf{TRELLIS}~\cite{Xiang_2025_CVPR} is a pretrained 3D generator based on structured latents (SLAT), which attach local feature vectors to the active voxels of a sparse 3D grid. It generates assets in two stages using rectified flow transformers: first establishing the coarse sparse voxel structure, and then synthesizing the local geometry and appearance latents on the active voxels. Finally, different decoders are applied to map SLAT to diverse 3D representations~\cite{mildenhall2020nerf,kerbl3Dgaussians,flexicube} of high quality.

\noindent\textbf{SpaceControl}~\cite{fedele2026spacecontrol} is a training-free method built on TRELLIS that enables conditioning asset generation on geometric inputs. Given a user-specified geometric control input $\mathcal{S}$ (primitives or meshes), it encodes the geometry into the structure latent space via TRELLIS's pretrained VAE encoder to obtain a clean latent $\mathbf{z}_1$. Instead of sampling from pure Gaussian noise, SpaceControl perturbs $\mathbf{z}_1$ to an intermediate flow timestep $\tau$, producing a partially noisy latent $\mathbf{z}_\tau$, and starts denoising from this state. The timestep $\tau$ serves as a global control parameter governing the trade-off between geometric adherence and generative freedom across the entire asset.

\Needspace{16\baselineskip}
\subsection{Method Overview}
\label{sec:model_overview}

We propose \methodname, a training-free framework built upon TRELLIS~\cite{Xiang_2025_CVPR}. As shown in Fig.~\ref{fig:overview}, our method takes as input a set of editable superquadric control shapes with per-primitive geometric conditioning strengths $\tau_i$, alongside optional appearance cues $a_i$ (text descriptors or reference images) for localized appearance control.

The primitives and their corresponding control levels $\tau_i$ first guide the Structure Flow Model (Sec.~\ref{sec:spatial_local_control}) during sparse-structure generation. The resulting sparse structure is then passed to the appearance stage. We extract and cluster PartField~\cite{liu2025partfield} features on this structure to segment the geometry, match each segment to its respective input primitive, and route the local appearance conditions to their target regions in the Appearance Flow Model (Sec.~\ref{sec:appearance_local_control}). Finally, self-similarity guidance~\cite{sayandsarkar_2025_guideflow3d} is applied to ensure coherent appearance and smooth cross-part transitions.

\subsection{Local Geometric Control}
\label{sec:spatial_local_control}

To implement local geometric control during structure generation, we partition the user-defined primitives \(\mathcal{S}\) according to their assigned control levels \(\tau_i\):

\begin{equation}
\mathcal{S}_r = \{S_i \in \mathcal{S} \mid \tau_i = \tau_r\}, \qquad r \in \{\mathrm{low}, \mathrm{high}\},
\label{eq:partition}
\end{equation}

where \(\tau_{\mathrm{low}}\) and \(\tau_{\mathrm{high}}\)  denote the endpoints of the flow interval over which spatial guidance is enforced.

First, we convert the complete set $\mathcal{S}$ and the high-control subset $\mathcal{S}_{\text{high}}$ into meshes independently. Both  meshes are encoded into the structure latent space via the TRELLIS VAE to obtain clean latents $\mathbf{z}_1^{\text{all}}$ and $\mathbf{z}_1^{\text{high}}$. Additionally, we construct a binary spatial mask $\mathbf{M}_{\text{low}}$ by generating a bounding box around $\mathcal{S}_{\text{low}}$ and downsampling the region to match the dimensions of the latent grid. Following~\cite{fedele2026spacecontrol}, we perturb $\mathbf{z}_1^{\text{all}}$ with noise at timestep $\tau_{\text{low}}$ to obtain $\mathbf{z}_{\tau_{\text{low}}}^{\text{all}}$ and begin the denoising process, thus establishing the coarse layout of the more flexible, low-control components.

To preserve the precise geometry and structural details of high-control regions, we inject the clean high-control latent $\mathbf{z}_1^{\text{high}}$ at each denoising step $t$. Using the low-control mask $\mathbf{M}_{\text{low}}$, we compute a spatially masked weighted average to update the latent state $\mathbf{z}'_t$:

\begin{equation}
\mathbf{z}'_t = (1 - \mathbf{M}_{\text{low}}) \odot \left( \alpha \mathbf{z}_1^{\text{high}} + (1 - \alpha) \mathbf{z}_t \right) + \mathbf{M}_{\text{low}} \odot \mathbf{z}_t
\label{eq:high_control_injection}
\end{equation}
where $\mathbf{z}_t$ is the intermediate denoised prediction, $\odot$ denotes the element-wise product, and $\alpha$ is the blending weight.

We leverage a RePaint-inspired resampling strategy~\cite{lugmayr2022repaint} over $K$ iterations for a smooth transition between high- and low-control regions. In each iteration, we  perturb the blended state from $t_{i+1}$ back to step $t_i$ following the noise schedule, denoise forward to $t_{i+1}$ in a single solver step, and re-apply the masked blending in 
Eq.~\ref{eq:high_control_injection}. This iterative resampling is performed across the interval $[\tau_{\text{low}},\tau_{\text{high}}]$, corresponding to the \textit{Conditioned Flow} phase shown in Fig.~\ref{fig:overview}.

\begin{figure*}[t]
    \centering

    \begin{minipage}[t]{0.465\textwidth}
        \vspace{0pt}
        \centering
        \includegraphics[width=\linewidth]{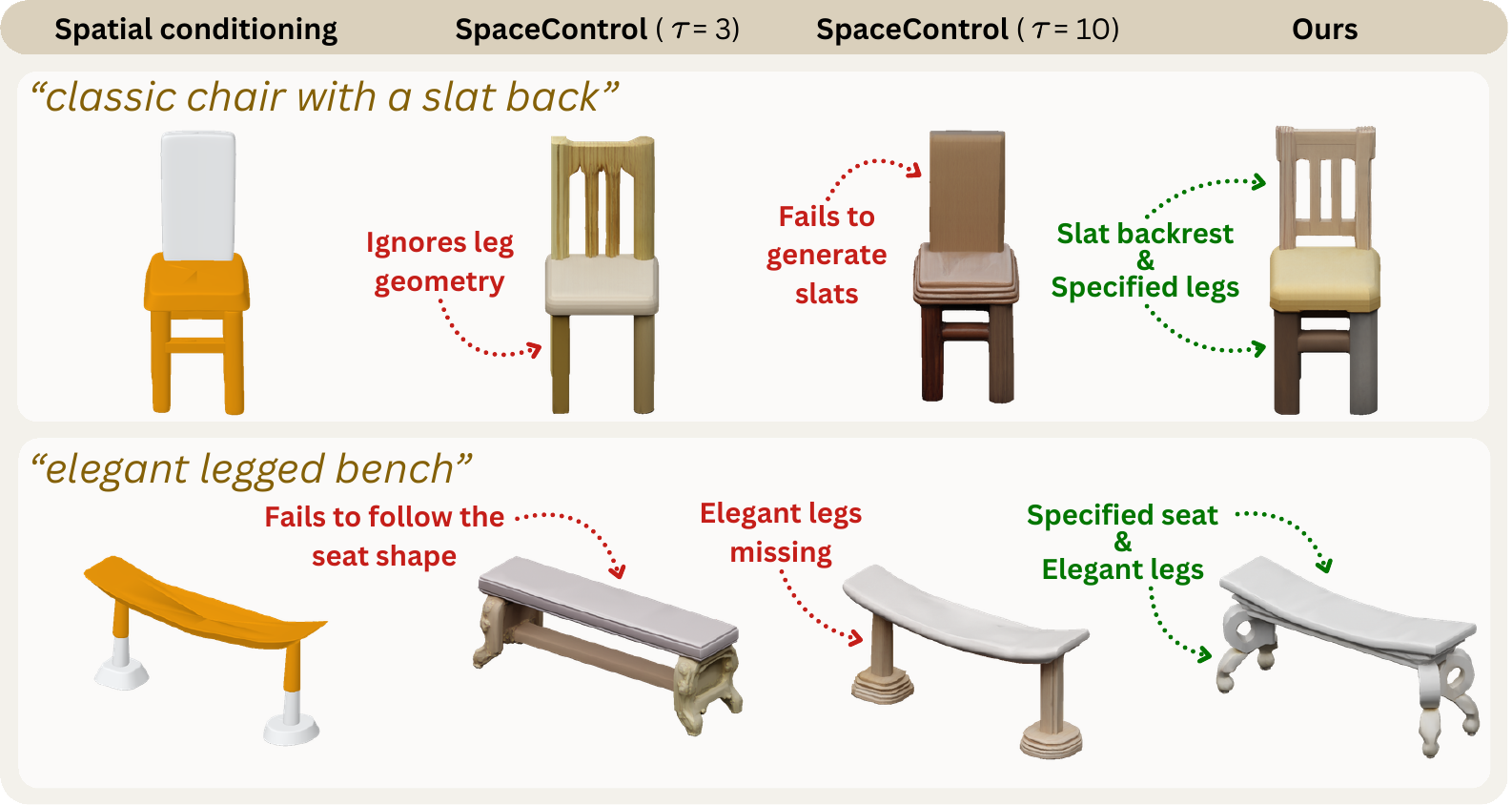}
        \textbf{(a) Spatial control in isolation}
    \end{minipage}
    \hfill
    \begin{minipage}[t]{0.52\textwidth}
        \vspace{0pt}
        \centering        \includegraphics[width=\linewidth]{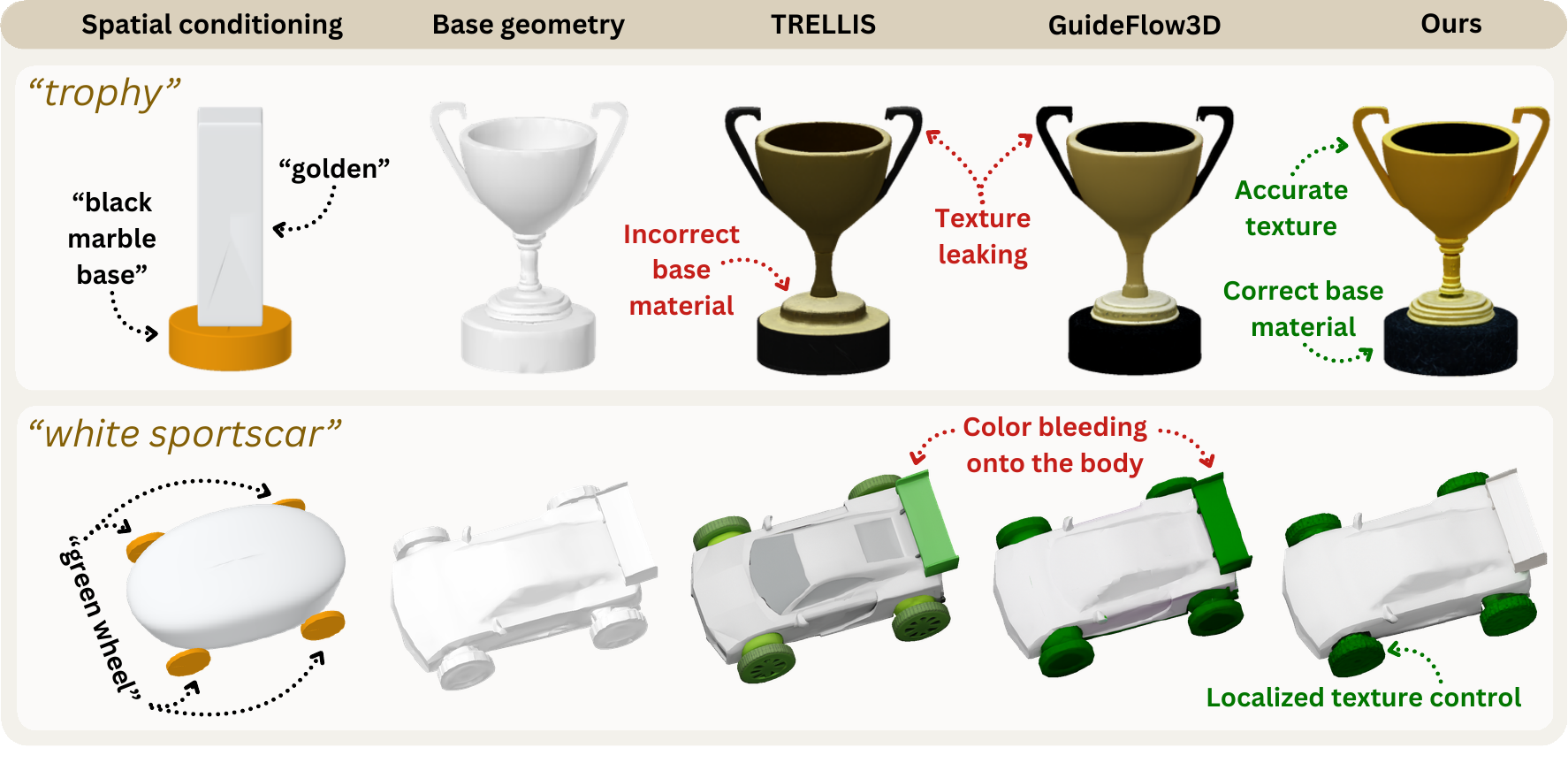}
        \textbf{(b) Appearance control in isolation}
    \end{minipage}
    \vspace{-3pt}
    \caption{
        \textbf{Spatial and appearance control in isolation.}
        \textbf{(a) Spatial control.}
        Uniform control with SpaceControl either loses specified structure
        ($\tau\!=\!3$) or constrains the full object ($\tau\!=\!10$), whereas
        \methodname preserves the high-control regions and allows the generative
        prior to complete the low-control regions.
        \textbf{(b) Appearance control on fixed geometry.}
        Global conditioning can miss part-level cues or spread them beyond their
        targets; \methodname routes each color or material cue to its target part
        while retaining coherent textures.
        }
    \label{fig:control_isolation}
    \vspace{-10pt}
\end{figure*}

\subsection{Local Appearance Routing and Guidance}
\label{sec:appearance_local_control}

The appearance stage takes as input the generated sparse structure and the local appearance cues $\mathcal{A}\!=\!\{a_i\}_{i=1}^{N}$, which default to the global text prompt $p$ when unspecified. To ensure each local cue affects only its designated primitive $S_i$, we propose \emph{local appearance routing}. Specifically, we define a routing volume $\Phi$ that maps every appearance-latent voxel to its corresponding condition $a_i$, dictating which prompt guides each spatial region along the flow trajectory.

Constructing $\Phi$ requires establishing semantic correspondences between the generated structure and initial primitives. Direct per-voxel geometric assignment is unreliable as low-control regions can be reshaped by the generative prior, causing drift from their original boundaries. To resolve this, we
extract PartField~\cite{liu2025partfield} descriptors from the generated
structure, cluster the active voxels via $K$-means, and assign each semantic cluster as a unit to its closest input primitive. This produces a geometrically adaptive part-consistent routing volume, accommodating for structural reinterpretation. Details on cluster-to-primitive assignment and small-part handling are
described in Supp.\

We build on the optimization-guided rectified flow mechanism in~\cite{sayandsarkar_2025_guideflow3d}, extending its global appearance guidance with the routing volume $\Phi$. Each cue $a_i$ is encoded into an embedding $\mathbf{e}_i$. Within every cross-attention layer, each latent voxel attends only to the embedding selected by $\Phi$. This confines each cue to its user-intended region.

Cross-attention routing alone does not completely isolate local cues because subsequent self-attention can inadvertently propagate and blend appearance features across distinct parts~\cite{wang2024compositional, park2024shape}. We inject
a soft self-attention bias modulated by $\Phi$. For any query-key pair sharing the same assigned condition, we add a positive bias of $\log\beta$ ($\beta > 1$) to their raw attention logit~\cite{ying2021transformers}. This prioritizes intra-part feature propagation without blocking interactions across the rest of the asset.

To further encourage a coherent appearance within semantic parts, we optimize a test-time supervised contrastive loss over the PartField semantic clusters $\{\ell_j\}$:
\vspace{-0.2cm}
\begin{equation}
\mathcal{L} =
-\frac{1}{|\mathcal{V}|}
\sum_j
\log
\frac{
    \sum_{k:\,\ell_k=\ell_j,\,k\neq j}\exp(s_{jk}/\tau_c)
}{
    \sum_{k\neq j}\exp(s_{jk}/\tau_c)
}\, ,
\label{eq:contrastive}
\vspace{-0.2cm}
\end{equation}

where $\mathcal{V}$ is the set of structure voxels, $s_{jk}$ is the cosine similarity between voxel latents, and $\tau_c$ is the temperature. During inference, we perform test-time optimization directly on the intermediate appearance latents without modifying the pretrained network weights. Specifically, at each guidance step, we compute $\nabla_{\mathbf{z}} \mathcal{L}$ and update the latent state. This pulls together features within the same semantic part while pushing distinct parts apart. To maintain optimal visual fidelity, we track the latents along the guidance trajectory and select the lowest-loss state for final decoding. Full implementation details, routing hyperparameters, and optimizer configurations are provided in Supp.\

\section{Experimental Results}
\label{sec:results}

We evaluate each control component separately to isolate their effects: Sec.~\ref{sec:geometric_control} analyzes local \textit{geometric} control, while Sec.~\ref{sec:appearance_control} evaluates localized \textit{appearance} conditioning.

\PAR{Dataset.}
To evaluate geometric and appearance control, we constructed a dataset of $83$ geometric and appearance conditionings covering diverse object categories. Geometric conditioning is provided through superquadric primitives with designated high-control ($\tau_i\!=\!10$) and low-control ($\tau_i\!=\!3$) regions. Appearance conditioning is defined by local text prompt associated to single or multiple superquadrics. We refer to Supp.\  for more details.

\subsection{Geometric Control}
\label{sec:geometric_control}

We evaluate our localized geometric control formulation ($\tau_{\text{high}}\!=\!10$ and $\tau_{\text{low}}\!=\!3$) in direct comparison to SpaceControl with uniform global guidance scales ($\tau\!=\!3$ or $\tau\!=\!10$). Specifically, $\tau\!=\!3$ allows the generative model substantial freedom to generate natural, plausible geometry while retaining a small level of structural guidance, whereas $\tau\!=\!10$ enforces strict geometric adherence to the input shapes.

\subsubsection{Quantitative Results}

We evaluate the performance of \methodname in local geometric control using \textit{GPT 5.6 Sol} as an automated VLM judge. We assess prompt and geometric fidelity (PGF), which captures high-control shape adherence and low-control prompt
completion, alongside realism and overall preference. Each comparison is evaluated in three passes with random ordering and aggregated by majority vote.

Tab.~\ref{tab:llm_geometric_eval_winrate} shows the head-to-head win rates against global guidance baselines. These results illustrate the trade-off of uniform guidance: it forces a single compromise across the entire asset. A high global
scale ($\tau\!=\!10$) enforces geometry at the expense of realism, whereas a low scale ($\tau\!=\!3$) allows generative freedom but weakens input adherence.

\textit{PGF} results demonstrate that \methodname reliably follows the high-control input geometry while using the generative prior to match the prompt in low-control regions. The \textit{Overall} metric confirms that, when evaluated on the complete input, our method is preferred against both baselines.

\begin{table}[!htbp]
\centering
\caption{\textbf{Localized geometric control.} Win rates (\%) of \methodname{} ($\tau_{\mathrm{low}}\!=\!3$, $\tau_{\mathrm{high}}\!=\!10$) against SpaceControl~\cite{fedele2026spacecontrol} using a uniform global control strength. Brackets report 95\% Wilson CIs over 83 assets.}
\label{tab:llm_geometric_eval_winrate}
\small
\setlength{\tabcolsep}{4pt}
\renewcommand{\arraystretch}{1.15}
\begin{tabular*}{\textwidth}{@{\extracolsep{\fill}}lcccccc@{}}
\toprule
& \multicolumn{2}{c}{Prompt and geometric fidelity} & \multicolumn{2}{c}{Realism} & \multicolumn{2}{c}{Overall} \\
\cmidrule(lr){2-3}\cmidrule(lr){4-5}\cmidrule(l){6-7}
Baseline & Win rate & 95\% CI & Win rate & 95\% CI & Win rate & 95\% CI \\
\midrule
SpaceControl ($\tau=3$) & 78.3\% & [68.3, 85.8] & 43.4\% & [33.2, 54.1] & 75.9\% & [65.7, 83.8] \\
\addlinespace[3pt]
SpaceControl ($\tau=10$) & 66.3\% & [55.6, 75.5] & 85.5\% & [76.4, 91.5] & 68.7\% & [58.1, 77.6] \\
\bottomrule
\end{tabular*}
\end{table}

\subsubsection{Geometric and Feature-Space Evaluation}
To complement the VLM evaluation, we quantify spatial fidelity, regional feature deviation, and semantic alignment using three metrics:

\PAR{(a) Regional Chamfer Distance.} We evaluate spatial fidelity to the input primitives by computing the Chamfer Distance (CD) separately on the designated high-control ($\text{CD}_{\text{high}}$) and low-control ($\text{CD}_{\text{low}}$) regions. Specifically, we partition mesh faces by mapping each face centroid to the nearest input primitive surface, inheriting its assigned control level. All meshes are normalized to a unit cube, and we report the symmetric mean-squared Chamfer distance ($\times10^3$).
To quantify the difference in control enforcement across regions, we compute the regional delta $\Delta\text{CD}\!=\!\text{CD}_{\text{low}} - \text{CD}_{\text{high}}$, where a higher positive value indicates that the model effectively differentiates control strengths, preserving specified geometry more strictly in high-control regions than in low-control ones.

\PAR{(b) Feature-Based Deviation.}
As a complementary measure of local control, we compare each source
voxel structure $V^s$, for
$s\in\{\text{local-}\tau,\tau\!=\!3\}$, with a uniformly guided
$\tau\!=\!10$ reference $V^{10}$. This reference remains close to the input while providing an object-like surface on which PartField descriptors are meaningful; direct adherence to the superquadric proxy is measured separately by Chamfer distance. Each active voxel $v\in V^s$ has a spatial coordinate $x_v$ and an $\ell_2$-normalized PartField feature $z_v^s$ sampled at $x_v$. We first match it to its nearest reference voxel:
\begin{equation}
u^*(v)=\underset{u\in V^{10}}{\operatorname{argmin}}\|x_v-x_u\|_2 .
\end{equation}
We then compute the cosine distance in PartField space:
\begin{equation}
d_{\mathrm{feat}}^{s\rightarrow 10}(v)
= 1-\mathbf{z}_v^s\cdot\mathbf{z}_{u^*(v)}^{10}.
\end{equation}
As for regional Chamfer distance, each voxel inherits the high- or
low-control label of its nearest input primitive. Let $V_r^s$ denote the source voxels assigned to region $r\in\{\mathrm{high},\mathrm{low}\}$. Their mean feature distance is
\begin{equation}
D_r^{s\rightarrow 10}
=\frac{1}{|V_r^s|}
\sum_{v\in V_r^s}d_{\mathrm{feat}}^{s\rightarrow 10}(v).
\end{equation}
Finally, the feature delta
$\Delta D^{s\rightarrow 10}
\!=\!D_{\mathrm{low}}^{s\rightarrow 10}
-D_{\mathrm{high}}^{s\rightarrow 10}$
is positive when low-control features deviate more from the reference
than high-control features, enabling direct comparison with
SpaceControl.

\PAR{(c) Text Alignment.} In order to verify semantic fidelity to the text prompt, we compute the CLIP similarity between rendered views of the generated assets and the input text.

\begin{wraptable}{r}{0.55\textwidth}
\vspace{-\intextsep}
\centering
\caption{\textbf{Quantitative evaluation of spatial control selectivity and semantic alignment.} We report the regional Chamfer Distance delta ($\Delta\text{CD}\!=\!\text{CD}_{\text{low}} - \text{CD}_{\text{high}}$, $\times 10^3$), the feature deviation delta to the reference ($\Delta D^{s \rightarrow 10}\!=\!D_{\text{low}}^{s \rightarrow 10} - D_{\text{high}}^{s \rightarrow 10}$), and CLIP text-image similarity. Positive $\Delta$ values indicate effective localized control selectivity between high- and low-control regions. $^\dagger\Delta D$ is $0.00$ by definition for $\tau\!=\!10$ as $D^{10 \rightarrow 10}\!=\!0$.}
\label{tab:quantitative_spatial}
\small
\setlength{\tabcolsep}{3.5pt}
\begin{tabular}{lcccc}
\toprule
Method & $\text{CD}_{\text{high}} \downarrow$ & $\Delta\text{CD} \uparrow$ & $\Delta D^{s \rightarrow 10} \uparrow$ & CLIP $\uparrow$ \\
\midrule
SpaceControl ($\tau\!=\!3$)    & 15.99          & -0.14         & 0.01                 & 0.232 \\
SpaceControl ($\tau\!=\!10$)    & \textbf{0.94}  & 0.11          & 0.00$^\dagger$       & 0.226 \\
\methodname~(\textit{Ours}) & 3.65           & \textbf{6.05} & \textbf{0.15}        & \textbf{0.233} \\
\bottomrule
\end{tabular}
\end{wraptable}

\PAR{Results.} 
Tab.~\ref{tab:quantitative_spatial} reports additional metrics evaluating our method against SpaceControl baselines under uniform global guidance ($\tau\!=\!3$ and $\tau\!=\!10$). The regional Chamfer Distance on high-control areas ($\text{CD}_{\text{high}}$) demonstrates that \methodname provides significantly stronger geometric adherence than $\tau\!=\!3$, approaching the strict compliance of the $\tau\!=\!10$ baseline.

The regional deltas show that while uniform SpaceControl baselines exhibit virtually no difference in the level of control across regions ($\Delta\text{CD}\!\approx\!0$ and $\Delta D^{s \rightarrow 10}\!\approx\!0$), \methodname achieves a substantial positive delta in both physical geometry ($\Delta\text{CD}$) and feature space ($\Delta D^{s \rightarrow 10}$). This confirms that our method effectively leverages the two control levels, performing more constrained generation in high-control than low-control regions.
Furthermore, CLIP similarity remains comparable across methods, suggesting that this selectivity does not reduce semantic alignment. Together, these objective metrics ground the VLM judge's findings, demonstrating that localized control successfully bypasses the rigid trade-off of uniform guidance across the entire asset.

\begin{figure}[!htbp]
  \centering
  \includegraphics[width=\linewidth]{figs/asset_gallery_compressed_transparent.pdf}
  \vspace{-5pt}
  \caption{\textbf{Qualitative comparison of spatial and appearance control across diverse assets.}
  \emph{Control (top):} Input superquadric proxies with designated local control strengths (orange: high control $\tau_i\!=\!10$; gray: low control $\tau_i\!=\!3$) and localized part-level appearance prompts.
  \emph{SpaceControl~\cite{fedele2026spacecontrol}:}  Baselines under uniform control strengths ($\tau\!=\!3$ vs.\ $\tau\!=\!10$), illustrating the trade-off between generative reinterpretation and strict adherence to the proxy geometry.
  \emph{Texture Baselines:} Appearance baselines (TRELLIS~\cite{Xiang_2025_CVPR} and GuideFlow3D~\cite{sayandsarkar_2025_guideflow3d}) built directly on our generated structure to isolate texture synthesis.
  \emph{Local Control (Ours):} \methodname provides localized geometric control (generating text-aligned geometric details in low-control regions while preserving high-control shapes) alongside accurate part-level texture routing.}
  \label{fig:examples}
  \vspace{-10pt}
\end{figure}

\subsubsection{Human Evaluation}

\begin{wrapfigure}{r}{0.48\textwidth}
\vspace{-\intextsep}
\captionsetup{skip=4pt}
\centering
\begin{tikzpicture}[x=1cm,y=1cm]
\definecolor{wincol}{HTML}{8CCB9B}
\definecolor{tiecol}{HTML}{D6D6D6}
\definecolor{losscol}{HTML}{E7AAA5}
\def\barw{5.8}
\def\barh{0.48}
\foreach \p in {0,20,40,60,80,100} {
  \pgfmathsetmacro{\xx}{\p*\barw/100}
  \draw[gray!20,line width=0.25pt] (\xx,-0.04) -- (\xx,1.45);
  \node[font=\scriptsize,anchor=north] at (\xx,-0.12) {\p\%};
}
\newcommand{\studybar}[5]{%
  \pgfmathsetmacro{\xw}{#2*\barw/100}
  \pgfmathsetmacro{\xt}{#3*\barw/100}
  \draw[fill=wincol,draw=black,line width=0.35pt] (0,#1) rectangle (\xw,{#1+\barh});
  \draw[fill=tiecol,draw=black,line width=0.35pt] (\xw,#1) rectangle ({\xw+\xt},{#1+\barh});
  \draw[fill=losscol,draw=black,line width=0.35pt] ({\xw+\xt},#1) rectangle (\barw,{#1+\barh});
  \node[font=\scriptsize,anchor=east] at (-0.12,{#1+\barh/2}) {#5};
  \node[font=\scriptsize] at ({\xw/2},{#1+\barh/2}) {#2\%};
  \node[font=\scriptsize] at ({\xw+\xt/2},{#1+\barh/2}) {#3\%};
  \node[font=\scriptsize] at ({(\xw+\xt+\barw)/2},{#1+\barh/2}) {#4\%};
}
\node[font=\small\bfseries] at (2.9,1.83) {Overall preference};
\studybar{0.88}{48}{13}{39}{$\tau=3$}
\studybar{0.12}{71}{10}{19}{$\tau=10$}
\foreach \col/\lab/\xx in {wincol/Wins/0.9,tiecol/Ties/2.7,losscol/Losses/4.3} {
  \draw[fill=\col,draw=black,line width=0.3pt] (\xx,-0.98) rectangle ({\xx+0.22},-0.76);
  \node[font=\scriptsize,anchor=west] at ({\xx+0.28},-0.87) {\lab};
}
\end{tikzpicture}
\vspace{3mm}
\caption{\textbf{User study} showing overall preference for \methodname over SpaceControl~\cite{fedele2026spacecontrol} with uniform $\tau\!=\!3$ and $\tau\!=\!10$ guidance across 337 evaluation trials.}
\label{fig:user_study_local_tau}
\vspace{-0.5\intextsep}
\end{wrapfigure}

To complement our quantitative evaluation, we conducted a pairwise user study assessing the perceptual quality of the generated assets. We compared \methodname{} against SpaceControl baselines under uniform guidance ($\tau\!=\!3$ and $\tau\!=\!10$). For each scene, three variants (\methodname{}, $\tau\!=\!3$, and $\tau\!=\!10$) were generated from identical text prompts and superquadric inputs. Participants inspected interactive 3D model pairs side-by-side and evaluated them based on shape fidelity, realism, and overall preference, yielding a total of 337~trials across 29~evaluators.

As shown in Fig.~\ref{fig:user_study_local_tau}, \methodname remains competitive in overall preference against low guidance ($\tau\!=\!3$), with a $48\%$ win rate, and is strongly preferred to the high-guidance baseline ($\tau\!=\!10$), with a $71\%$ win rate. Overall, the study shows that local control avoids the rigid trade-off of uniform global guidance while maintaining overall quality.

\subsection{Appearance Control}
\label{sec:appearance_control}

We assess localized appearance control using \textit{GPT 5.6 Sol} as an automated vision-language judge, evaluating multiple rendered views of each asset. 

\noindent\textbf{Text-conditioned appearance.}
We evaluate on $83$ text-conditioned assets across three variants sharing an identical base structure: 
(i)~\textbf{TRELLIS}, applying the appearance flow model on fixed geometry without guidance or routing; 
(ii)~\textbf{GuideFlow3D}, employing self-similarity guidance without routing; and 
(iii)~\textbf{\methodname} (ours), adding part-wise condition routing to self-similarity guidance (Sec.~\ref{sec:appearance_local_control}).

\begin{table}[!htbp]
\centering
\caption{\textbf{Pairwise appearance evaluation.}
Win rates (\%) of \methodname{} against fixed-structure appearance
baselines. Brackets report 95\% Wilson CIs over the set of test assets.}
\label{tab:vlm_appearance_eval_winrate}
\small
\setlength{\tabcolsep}{4pt}
\renewcommand{\arraystretch}{1.15}
\begin{tabular*}{\textwidth}{@{\extracolsep{\fill}}lcccccc@{}}
\toprule
& \multicolumn{2}{c}{Prompt faithfulness}
& \multicolumn{2}{c}{Texture detail}
& \multicolumn{2}{c}{Overall} \\
\cmidrule(lr){2-3}\cmidrule(lr){4-5}\cmidrule(l){6-7}
Baseline & Win rate & 95\% CI & Win rate & 95\% CI & Win rate & 95\% CI \\
\midrule
TRELLIS
& 63.9\% & [53.1, 73.4]
& 31.3\% & [22.4, 41.9]
& 59.0\% & [48.3, 69.0] \\
\addlinespace[3pt]
GuideFlow3D
& 58.5\% & [47.7, 68.6]
& 37.5\% & [27.7, 48.5]
& 57.3\% & [46.5, 67.5] \\
\bottomrule
\end{tabular*}
\end{table}

The input specifies per-primitive requests, such as an orange backrest or a red cushion, consumed by all methods. We perform two complementary evaluations:

First, an \emph{absolute rating} on a $1$--$10$ scale across four metrics: \emph{prompt faithfulness} (evaluating localized per-primitive adherence), \emph{color and material accuracy} (evaluating the correctness of colors and surface finishes relative to the prompt), \emph{texture detail}, and \emph{overall preference} (see Supp.\ for details). We report the mean scores in Tab.~\ref{tab:quantitative_appearance}.
Second, analogous to our geometric evaluation, we conduct a \emph{pairwise head-to-head evaluation} to measure win rates across prompt faithfulness, texture detail, and overall preference.

\begin{wraptable}{r}{0.48\textwidth}
\vspace{-\intextsep}
\centering
\caption{\textbf{Quantitative evaluation of appearance control.} Mean scores ($1$--$10$) averaged over 83 cases: prompt faithfulness (PF), color/material accuracy (CM), texture detail (TD), and overall. All variants share the same structure; metrics are appearance-only.}
\label{tab:quantitative_appearance}
\vspace{2.5mm}
\small
\setlength{\tabcolsep}{3.5pt}
\begin{tabular}{lcccc}
\toprule
Method & PF\,$\uparrow$ & CM\,$\uparrow$ & TD\,$\uparrow$ & Overall\,$\uparrow$ \\
\midrule
TRELLIS & 6.13 & 5.79 & \textbf{6.87} & 6.42 \\
GuideFlow3D & 6.40 & 5.96 & 6.76 & 6.46 \\
\methodname (ours)                                & \textbf{7.07} & \textbf{6.43} & 6.49 & \textbf{6.78} \\
\bottomrule
\end{tabular}
\end{wraptable}

In pairwise comparisons (Tab.~\ref{tab:vlm_appearance_eval_winrate}), \methodname shows an advantage on prompt-faithfulness and overall preference (considering the input requirements) over TRELLIS~\cite{Xiang_2025_CVPR} and GuideFlow3D~\cite{sayandsarkar_2025_guideflow3d}.  
This shows that part-wise routing improves prompt fidelity by ensuring that localized appearance cues are faithfully reflected in the final asset and eliminating cross-part color bleeding. Conversely, both baselines are favored on texture detail, indicating that localized routing prioritizes part-level prompt adherence at some cost to unconstrained texture
quality.

The complementary absolute ratings in Tab.~\ref{tab:quantitative_appearance} show that \methodname’s gains in localized prompt fidelity and material accuracy compensate for its lower unconstrained texture uniformity, yielding the highest overall score. Further, since all three variants share the same base structure, Tab.~\ref{tab:quantitative_appearance} isolates the improvement from self-similarity guidance from TRELLIS~\cite{Xiang_2025_CVPR} to GuideFlow3D~\cite{sayandsarkar_2025_guideflow3d}, and the additional effect of localized condition routing in \methodname.

\noindent\textbf{Image-conditioned appearance.}
In the \emph{image} regime, appearance is dictated by reference style images. Because baseline methods cannot process per-part image conditions, \methodname is scored standalone on \emph{local routing accuracy}, \emph{style fidelity}, \emph{texture detail}, and \emph{overall preference} (see Supp.\ for details). \methodname obtains a style fidelity of $6.53$, a local routing accuracy of $7.07$, texture detail of $6.33$, and an overall score of $6.73$, confirming that our framework effectively transfers visual style references to their designated regions.

\subsection{Qualitative Results}

As shown in Fig.~\ref{fig:control_isolation}, existing methods use global control, forcing the entire object to follow either geometric fidelity ($\tau\!=\!10$) or generative freedom ($\tau\!=\!3$), while global appearance cues may leak across parts. \methodname avoids this compromise by assigning control strengths per region and routing appearance cues to their corresponding parts.

Fig.~\ref{fig:examples} presents a diverse gallery of multi-part objects. Under our joint control, high-control primitives (orange) strictly guide the asset geometry in the specified area, while low-control proxies (gray) are completed with text-aligned geometric details (\eg, the rocket tail fins). To benchmark each modality without confounding factors, the figure isolates both axes: SpaceControl baselines demonstrate the structural failure modes of uniform $\tau$ under identical texturing, whereas TRELLIS and GuideFlow3D show texture leakage across our fixed geometry. In contrast, \methodname  consistently preserves user-specified geometry while cleanly localizing textures to their target parts (see Supp.\ for more qualitative results).

\FloatBarrier
\section{Conclusion}

We introduced \methodname, an inference-time framework for localized geometry and appearance control in 3D asset generation. Editable geometric inputs define the controlled regions, each associated with a geometric control level and a text or image appearance cue. During structure generation, the control levels modulate spatial guidance, allowing constrained regions to follow the input geometry more closely while leaving others unconstrained, relying on the generative prior. During appearance synthesis, the framework matches the input regions to generated parts and routes each cue only to its corresponding part, supporting localized conditioning while limiting cross-part leakage.

Evaluations combining vision-language judge assessments, user studies, and regional metrics demonstrate that \methodname successfully overcomes the limitations of uniform global guidance, enabling localized geometric control. Furthermore, localized appearance routing consistently outperforms existing unrouted baselines in prompt faithfulness. Overall, \methodname enables users to specify not only \emph{what} to generate, but precisely \emph{where} each spatial constraint and appearance cue should apply.

\section{Acknowledgements}
NF is funded by the Rafael del Pino Foundation. EF is supported by the
ETH AI Center doctoral fellowship and by the Swiss National Science Foundation (SNSF) Advanced Grant 216260 (\textit{Beyond Frozen Worlds: Capturing Functional 3D Digital Twins from the Real World}). SDS is partly supported by the Stanford Doerr School of Sustainability.

{
    \small
    \bibliographystyle{unsrtnat}
    \bibliography{main}
}

\clearpage
\appendix
\section*{Supplementary Material}

\setcounter{section}{0}
\setcounter{figure}{0}
\setcounter{table}{0}
\setcounter{equation}{0}
\renewcommand{\thesection}{S\arabic{section}}
\renewcommand{\thefigure}{S\arabic{figure}}
\renewcommand{\thetable}{S\arabic{table}}
\renewcommand{\theequation}{S\arabic{equation}}

\begin{abstract}
\noindent\textit{This supplementary material contains:}
\begin{itemize}
    \item \textit{Structure generation and guidance architecture (Sec.~\ref{sec:supp_structure_architecture}) and the appearance generation pipeline (Sec.~\ref{sec:supp_appearance_architecture}).}
    \item \textit{Superquadric primitives and their mathematical formulation (Sec.~\ref{sec:superquadric_primitives}).}
    \item \textit{Details of the 83-asset evaluation dataset (Sec.~\ref{sec:dataset_details}).}
    \item \textit{Vision-language-model evaluation, including structure-control evaluation, pairwise and standalone appearance evaluation, judge configuration, and validation (Sec.~\ref{sec:supp_vlm_evaluation}).}
    \item \textit{User-study details and protocol (Sec.~\ref{sec:supp_user_study}).}
    \item \textit{Additional analyses, including sensitivity to local control strength, spatial feature-distance visualization, and primitive-to-part assignment visualizations (Sec.~\ref{sec:additional_analyses}).}
    \item \textit{Limitations (Sec.~\ref{sec:limitations}).}
    \item \textit{Additional qualitative results for image-conditioned and text-conditioned appearance control (Sec.~\ref{sec:supp_qualitative}).}
    \item \textit{Implementation details (Sec.~\ref{sec:supp_implementation}).}
    \item \textit{Interactive examples in an accompanying HTML gallery, with rotatable 3D input geometries and generated results that can be inspected from arbitrary viewpoints.}
\end{itemize}
\end{abstract}

\section{Structure Generation and Guidance Architecture}
\label{sec:supp_structure_architecture}

\begin{figure*}
    \centering
    \includegraphics[width=0.62\linewidth]{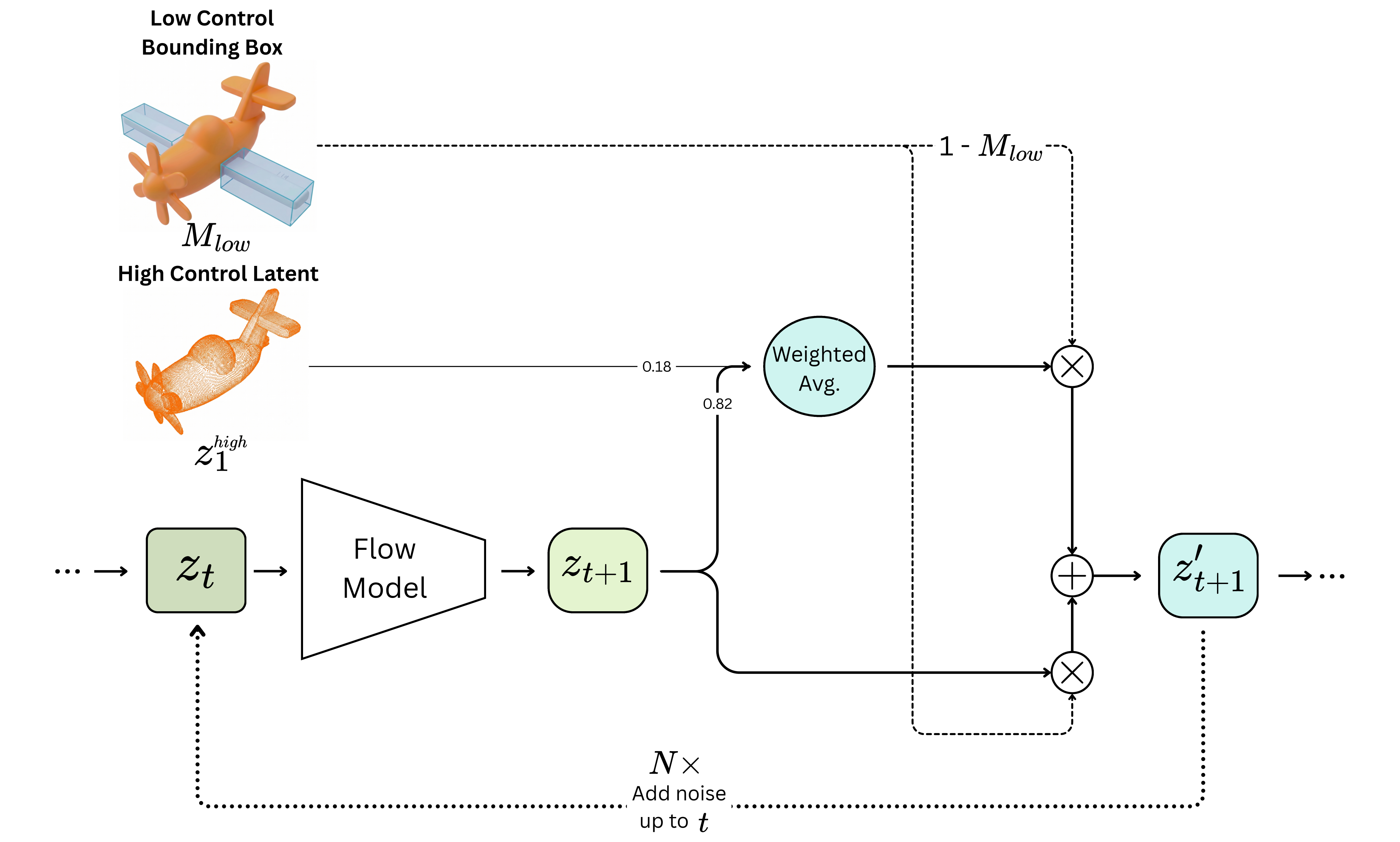}
    \caption{
        \textbf{Illustration of a single Conditioned Flow iteration.} The process receives the low-control region mask $M_{\text{low}}$ (derived from the bounding boxes), the high-control target latent $z_1^{\text{high}}$, and the current latent state $z_t$. The flow model predicts the next state $z_{t+1}$. Inside the high-control region (masked by $1 - M_{\text{low}}$), a weighted average of the target latent $z_1^{\text{high}}$ and the predicted latent $z_{t+1}$ is computed. This is combined with the unconstrained prediction $z_{t+1}$ in the low-control region (masked by $M_{\text{low}}$) to form $z'_{t+1}$. Finally, noise is added to $z'_{t+1}$ up to timestep $t$ (repeated $K$ times) to update the latent $z_t$ for the subsequent step.
    }
    \label{fig:supp_structure_architecture}
\end{figure*}

Tab.~\ref{tab:supp_structure_settings} details the hyperparameters used across our spatial control evaluations. In addition, we present a schematic of our Conditioned Flow formulation, which is applied between $\tau_{\text{low}}$ and $\tau_{\text{high}}$, in Fig.~\ref{fig:supp_structure_architecture}. As illustrated, this mechanism blends the flow-predicted latent with the high-control target latent inside the designated control regions while allowing the low-control regions to generate freely. The combined latent is then updated through a feedback loop that adds noise back to the target timestep. This refinement process is repeated $K$ times before the model proceeds to the next generation step. At each of the $K$ resampling iterations, a fresh noise sample is drawn independently.

To match the notation of the main paper, $K$ denotes the number of conditioned-flow resampling iterations. We use $K_{\mathrm{PF}}$ to denote the number of PartField semantic clusters.

\PAR{Latent Space Mask Generation.}
To spatially constrain the conditioning mechanism, we construct a discrete 3D binary mask from the input primitives. For low-control regions, a single bounding box is fitted around all low-control superquadrics with $7\%$ diagonal padding. This continuous representation is first voxelized at a high resolution of $64^3$, then downsampled to the $16^3$ latent space resolution via $4 \times 4 \times 4$ average pooling, and finally binarized with a threshold of $> 0$ (marking any latent cell with non-zero occupancy). The resulting binary mask directly dictates where the latent flow blending is applied.

\begin{figure*}[h!]
    \centering
    \includegraphics[width=0.9\linewidth]{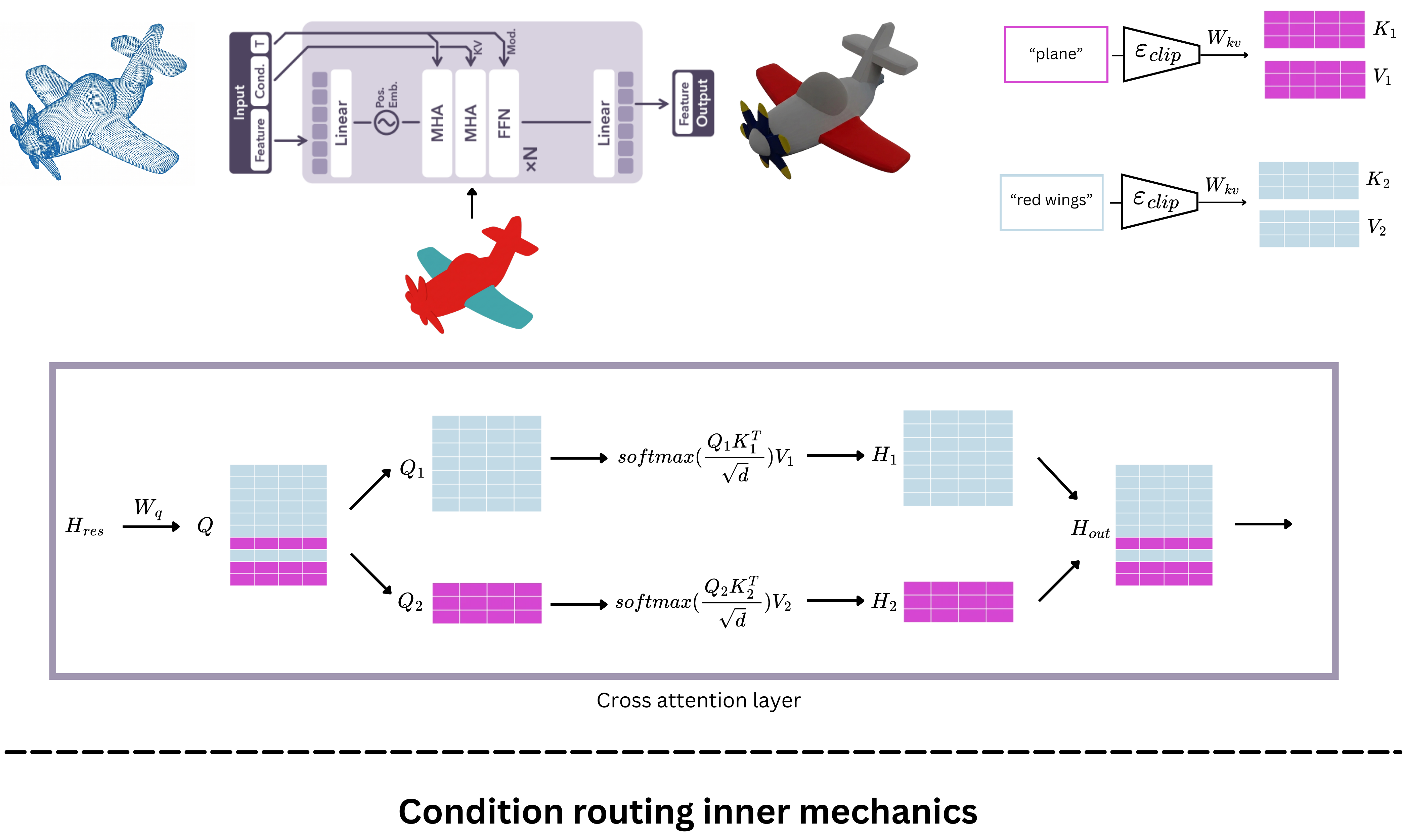}
    \caption{
        \textbf{Routed cross-attention.} Each text or image cue is encoded by a
        frozen CLIP (text) or DINOv2 (image) encoder into its own key/value pair
        $K_i, V_i$. Within every cross-attention layer of the appearance
        transformer, the latent queries $Q$ are partitioned by the routing
        volume so that the voxels of a part attend only to their assigned
        cue---\eg \emph{``red wings''} routes to the wings while the global
        \emph{``plane''} prompt covers the remaining voxels---and the per-region
        outputs $H_i$ are recombined into $H_{\mathrm{out}}$. This confines each
        appearance cue to its target part and reduces cross-part texture
        leakage.
    }
    \label{fig:supp_appearance_routing}
\end{figure*}

\clearpage
\section{Appearance Generation Pipeline}
\label{sec:supp_appearance_architecture}

\begin{wrapfigure}[29]{r}{0.49\textwidth}
    \vspace{-\intextsep}
    \captionsetup{skip=4pt}
    \centering
    \includegraphics[width=\linewidth]{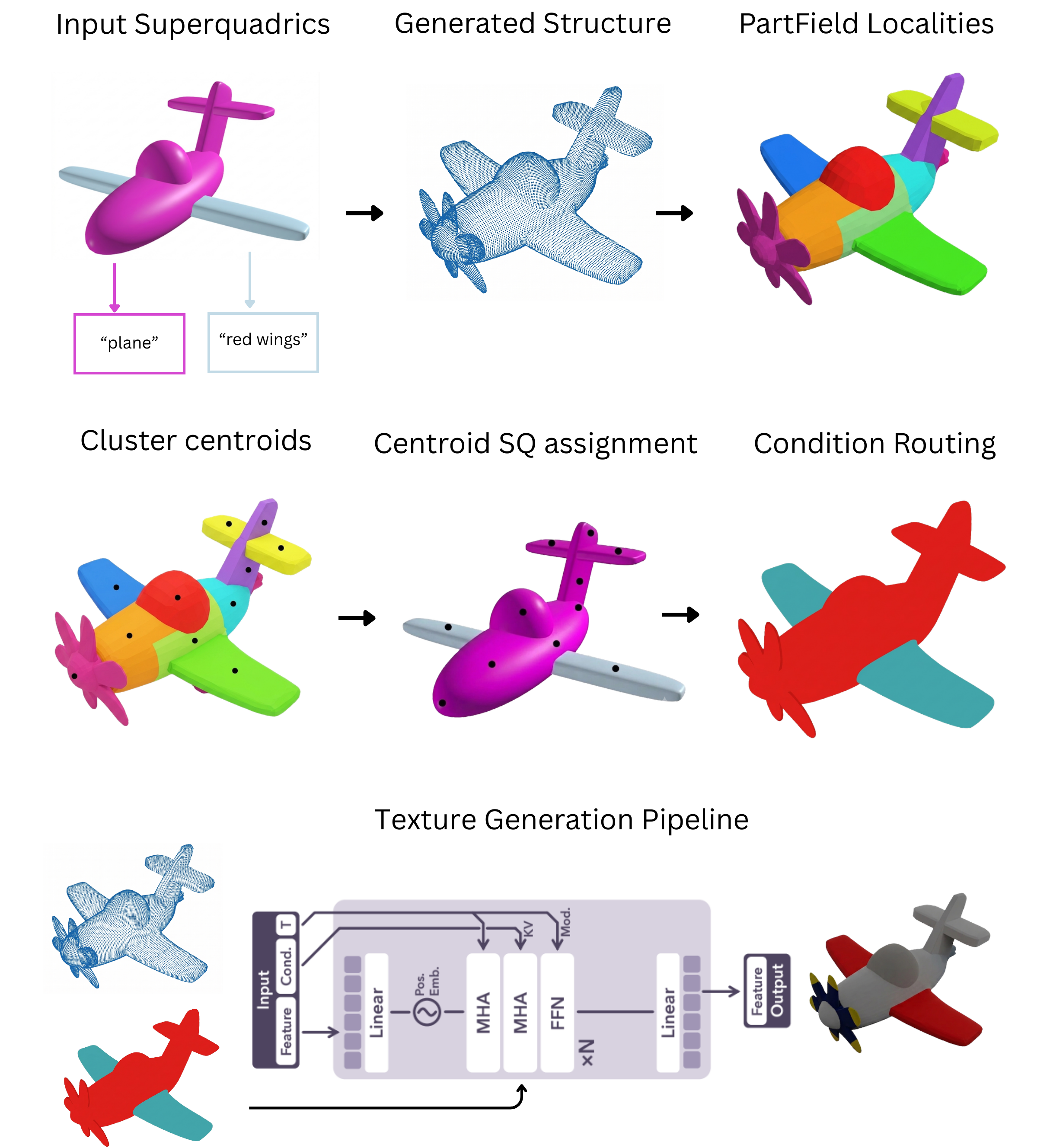}
    \caption{
        \textbf{Part-wise appearance pipeline.} From the generated sparse
        structure and the part-wise text or image cues, we decode the structure
        into a mesh and extract a PartField triplane feature field. Active
        structure voxels are clustered into semantic localities; each cluster
        centroid is assigned to its nearest input superquadric, and the
        resulting map is rasterized into a condition-routing volume that stays
        part-consistent even where low-control geometry drifts from the
        scaffold. The appearance rectified-flow transformer then synthesizes the
        SLAT appearance latents under this routing.
    }
    \label{fig:supp_appearance_pipeline}
    \vspace{-0.5\intextsep}
\end{wrapfigure}

Tab.~\ref{tab:supp_appearance_settings} details the hyperparameters used across our appearance control evaluations, and Fig.~\ref{fig:supp_appearance_pipeline} gives an overview of the part-wise appearance pipeline.

The appearance stage operates on the sparse structure produced by the first stage. We decode the structure into a mesh and extract a PartField feature representation. Features sampled at active structure voxels are clustered into semantic regions, and each cluster is matched to its closest input primitive. This produces a routing volume that remains part-consistent even when low-control geometry deviates from the original superquadric geometric input.

Each global or part-specific text or image condition is encoded into a
separate embedding. Within the appearance-flow transformer, every latent
voxel attends only to the embedding assigned by the routing volume;
unassigned voxels attend to the global condition. This routed cross-attention, illustrated in Fig.~\ref{fig:supp_appearance_routing}, restricts each cue
to its intended region. During sampling, we additionally optimize a
part-aware self-similarity objective that encourages coherent appearance
within each semantic part and separation across different parts.

\PAR{Routing-volume construction.}
We decode the generated sparse structure into a mesh and extract a PartField%
triplane feature field $\mathcal{P}$. At each active structure voxel $\mathbf{v}_j$ on the $64^3$ feature-sampling grid, we sample a descriptor $\mathbf{f}_j$ and cluster the descriptors with $K_{\mathrm{PF}}$-means, obtaining semantic labels $\ell_j$. For each cluster $C_m$, we select the active voxel $\mathbf{c}_m$ nearest the cluster mean as its representative, where $m\in\{1,\ldots,K_{\mathrm{PF}}\}$. We voxelize the surface of each input primitive $S_i$ into a set $\mathcal{B}_i$ and assign the complete cluster to the closest primitive:
\begin{equation}
a(m)=\arg\min_i\min_{\mathbf{b}\in\mathcal{B}_i}
\left\lVert\mathbf{c}_m-\mathbf{b}\right\rVert_2.
\label{eq:supp_cluster_assignment}
\end{equation}
Each active voxel inherits the primitive assignment $a(\ell_j)$ of its cluster, so all voxels in the same semantic cluster are routed consistently. We rasterize these inherited assignments on the structure voxel grid and aggregate them onto the coarser $32^3$ appearance-latent grid to obtain the routing volume $\Phi$. After downsampling, we count how many latent cells are assigned to each primitive. Thin or small primitives, such as wheels, handles, or chair legs, may receive too few cells because multiple fine voxels are merged into a single coarse latent cell. When this support falls below a minimum threshold, we recompute the affected mask directly on the latent grid by assigning each latent-cell center to its nearest voxelized primitive. We then dilate the recovered mask by one-cell neighborhood step, ensuring that the corresponding local appearance cue has enough latent cells to influence the generated texture while remaining spatially close to the intended primitive.

\PAR{Attention routing.}
The global condition and every primitive-specific text or image cue are encoded separately, using the encoders listed in Tab.~\ref{tab:supp_appearance_settings}. In each cross-attention layer, $\Phi$ selects the condition embedding attended to by every latent voxel; voxels without a local assignment use the global condition. For the self-attention bias, we add $\log\beta$ to the attention logit of each query--key pair whose voxels share the same routed condition. This is equivalent to multiplying its pre-softmax attention weight by $\beta$ and preserves soft interactions with the rest of the object.

\PAR{Guidance optimization.}
At each conditioned appearance-flow step, we evaluate the supervised contrastive objective in Equation~\ref{eq:contrastive}. To avoid materializing the full $|\mathcal{V}|\times|\mathcal{V}|$ similarity matrix, pairwise similarities are computed in chunks, with the chunk size selected according to the number of active voxels. We then take one AdamW step on the appearance latents. The lowest-loss latent state encountered along the guidance trajectory is retained, denormalized, and decoded into the final textured asset. Setting the guidance weight to zero disables the contrastive update and recovers part-routed sampling alone. Tab.~\ref{tab:supp_appearance_settings} reports the grid resolutions, $K_{\mathrm{PF}}$, $\beta$, chunk sizes, optimizer settings, and number of appearance-flow steps.

\vspace{-0.2cm}
\begin{equation}
\mathcal{L} =
-\frac{1}{|\mathcal{V}|}
\sum_j
\log
\frac{
    \sum_{k:\,\ell_k=\ell_j,\,k\neq j}\exp(s_{jk}/\tau_c)
}{
    \sum_{k\neq j}\exp(s_{jk}/\tau_c)
}\, ,
\label{eq:contrastive}
\vspace{-0.2cm}
\end{equation}
where $\mathcal{V}$ is the set of structure voxels, $s_{jk}$ is the cosine similarity between voxel latents, and $\tau_c$ is the temperature. Voxels with no same-cluster neighbors are excluded from the mean through a validity mask.

\begin{table}[h!]
    \centering
    \caption{
        \textbf{Structure-generation settings used in our spatial-control experiments.}
    }
    \label{tab:supp_structure_settings}
    \small
    \begin{tabular}{p{0.48\linewidth}p{0.4\linewidth}}
        \toprule
        \textbf{Setting} & \textbf{Value} \\
        \midrule
        Structure backbone
            & TRELLIS text sparse-structure flow model \\
        Structure latent tensor shape
            & $8 \times 16 \times 16 \times 16$ \\
        Local control strengths
            & $\tau_{\mathrm{low}} = 3$, $\tau_{\mathrm{high}} = 10$ \\
        High-control blending weight ($\alpha$)
            & $0.18$ for local-$\tau$ \\
        Number of structure flow steps
            & $12$ \\
        Flow solver
            & Euler flow-matching sampler with classifier-free guidance interval \\
        CFG strength and interval
            & $7.5$, active for $t \in [0.5, 0.95]$ \\
        Time rescaling
            & $3.0$ \\
        Resampling iterations $K$
            & $10$ \\
        \bottomrule
    \end{tabular}
\end{table} 
\begin{table}[h!]
    \centering
    \caption{
        \textbf{Appearance-generation and self-similarity guidance settings.}
        Text-conditioned appearance is evaluated pairwise and with standalone VLM scores;
        image-conditioned appearance is evaluated standalone using the criteria in
        Fig.~\ref{fig:supp_image_prompt} and is additionally illustrated qualitatively.
    }
    \label{tab:supp_appearance_settings}
    \small
    \begin{tabular}{p{0.46\linewidth}p{0.42\linewidth}}
        \toprule
        \textbf{Setting} & \textbf{Value} \\
        \midrule
        Appearance backbone
            & TRELLIS SLAT flow model and decoders \\
        Conditioning used in quantitative evaluation
            & Global and part-specific text prompts \\
        Text-condition encoder
            & CLIP ViT-L/14 \\
        Image-condition encoder
            & DINOv2 ViT-L/14, used for image-conditioned standalone evaluation and qualitative examples \\
        Part representation
            & PartField triplanes \\
        PartField descriptor dimension
            & $448$ \\
        Feature-sampling grid
            & Generated sparse voxels on a $64 \times 64 \times 64$ grid \\
        Local routing volume
            & $32 \times 32 \times 32$, only for local appearance prompts \\
        Clustering method
            & $K_{\mathrm{PF}}$-means on PartField voxel descriptors \\
        Number of PartField clusters $K_{\mathrm{PF}}$
            & $30$ \\
        Self-attention boost $\beta$
            & $2.5$ \\
        Guidance optimizer
            & AdamW \\
        Guidance weight
            & $1.0$ \\
        Optimizer learning rate
            & $5 \times 10^{-4}$ \\
        Similarity-computation chunk size
            & Automatic: $1024$, $512$, or $256$ depending on voxel count \\
        Number of appearance-flow steps
            & $300$ \\
        Contrastive temperature $\tau_c$ & 1.0 \\
        \bottomrule
    \end{tabular}
\end{table}

\section{Superquadric Primitives}
\label{sec:superquadric_primitives}
Superquadrics are compact geometric primitives capable of representing a continuous spectrum of shapes, such as ellipsoids, cylinders, rounded boxes, and cuboids.
A standard superquadric is defined by its scale $\mathbf{s} = (s_x, s_y, s_z)^\top \in \mathbb{R}^3_+$ and shape exponents $\boldsymbol{\epsilon} = (\epsilon_1, \epsilon_2)^\top \in \mathbb{R}^2_+$, which govern its roundness and squareness along the vertical and horizontal cross-sections.

For a 3D point $\mathbf{x} \in \mathbb{R}^3$, let $\mathbf{u} = (u_x, u_y, u_z)^\top = \mathbf{R}^\top(\mathbf{x}-\mathbf{t})$ represent its coordinates in the primitive's local frame, transformed by rotation $\mathbf{R} \in \mathrm{SO}(3)$ and translation $\mathbf{t} \in \mathbb{R}^3$. The volume bounded by the superquadric is implicitly defined by the inside-outside function $F(\mathbf{x}) \leq 1$:
\begin{equation}
F(\mathbf{x}) = \left( \lvert \frac{u_x}{s_x}\rvert^{\frac{2}{\epsilon_2}} + \lvert \frac{u_y}{s_y}\rvert^{\frac{2}{\epsilon_2}} \right)^{\frac{\epsilon_2}{\epsilon_1}} + \lvert \frac{u_z}{s_z}\rvert^{\frac{2}{\epsilon_1}} ,
\label{eq:superquadric_implicit}
\end{equation}
where $F(\mathbf{x}) < 1$ corresponds to points inside the primitive, $F(\mathbf{x}) = 1$ on its surface, and $F(\mathbf{x}) > 1$ outside.

\noindent \textbf{Deformable Superquadrics.}
We use the deformable superquadric representation of SuperFlex.\footnote{
\url{https://github.com/GabrielTavernini/superflex}
}
For a world-space point $\mathbf{x}$, let
$\mathbf{q}=\mathbf{R}^{\top}(\mathbf{x}-\mathbf{t})$ denote its coordinates
in the primitive's local frame. Let
$F_{\mathrm{can}}(\mathbf{v};\mathbf{s},\boldsymbol{\epsilon})$ denote the
right-hand side of Eq.~\eqref{eq:superquadric_implicit} with
$\mathbf{u}$ replaced by $\mathbf{v}$. The deformed inside--outside
function is
\begin{equation}
F_{\mathrm{def}}(\mathbf{x})
=
F_{\mathrm{can}}\!\left(
    \mathcal{D}^{-1}_{\boldsymbol{\tau},\boldsymbol{\beta}}(\mathbf{q});
    \mathbf{s},\boldsymbol{\epsilon}
\right),
\label{eq:deformable_superquadric}
\end{equation}
where
$\boldsymbol{\tau}=(\tau_x,\tau_y)$ controls tapering and
$\boldsymbol{\beta}_a=(k_a,\alpha_a)$ controls bending about local axis
$a\in\{x,y,z\}$. The inverse deformation follows the composition
\begin{equation}
\mathcal{D}^{-1}_{\boldsymbol{\tau},\boldsymbol{\beta}}
=
\mathcal{T}^{-1}_{\boldsymbol{\tau}}
\circ
\mathcal{B}^{-1}_{y,\boldsymbol{\beta}_y}
\circ
\mathcal{B}^{-1}_{x,\boldsymbol{\beta}_x}
\circ
\mathcal{B}^{-1}_{z,\boldsymbol{\beta}_z}.
\label{eq:deformation_composition}
\end{equation}
Thus, inverse bending is applied about $z$, then $x$, then $y$, followed
by inverse tapering. Equivalently, the forward map tapers the canonical
primitive first and then bends it about $y$, $x$, and $z$. We use
$\boldsymbol{\tau}\in(-1,1)^2$, $k_a\geq 0$, and
$\alpha_a\in[0,2\pi)$, with angles measured in radians. The exact
tapering and bending maps, axis conventions, parameterization, and
admissible curvature range are specified by the linked implementation.

In our pipeline, these deformable primitives serve as compact, editable
spatial control regions rather than final surface meshes.

\section{Dataset Details}
\label{sec:dataset_details}
We constructed a dataset of $83$ 3D assets covering diverse object categories. Each asset is defined by superquadric primitives with designated high-control and low-control regions, alongside local appearance prompts $a_i$. 

The benchmark combines manually authored configurations designed to reflect realistic use cases with assets generated via a semi-automated pipeline in which an LLM (\textit{ChatGPT 5.6 Sol}) proposed candidate primitive layouts, local control assignments, and appearance cues. All assets were validated and, if needed, refined by human annotators. The validation criteria were the following:

(i)~\emph{meaningful control contrast}, ensuring high-control regions specify well-defined structural elements while low-control regions remain intentionally coarse to test generative completion rather than trivial generation.
(ii)~\emph{spatial validity}, confirming that low-control primitives are not geometrically contained by high-control ones.
(iii)~\emph{appearance fidelity}, verifying that per-part appearance prompts are coherent, unambiguous, and accurately mapped to target primitives.

\PAR{Object Categories.}
In Tab.~\ref{tab:categories_dataset}, we report the distribution of object categories in the benchmark along with representative examples. The dataset spans $7$ broad semantic categories covering everyday manufactured goods, furniture, vehicles, electronics, and natural/decorative objects, with primitive complexity ranging from $2$ to $18$ parts per asset.

\begin{table*}[t]
  \centering
  \caption{\textbf{Object category distribution in the evaluation benchmark ($N=83$).} The benchmark spans $7$ broad semantic categories designed to evaluate multi-part geometric control and local appearance conditioning.}
  \label{tab:categories_dataset}
  \small
  \resizebox{\linewidth}{!}{%
  \begin{tabular}{lccl}
    \toprule
    Category & Count & Share & Representative Examples \\
    \midrule
    Vehicles \& Transport        & 15 & 18.1\% & Airplane, Space Shuttle, Satellite, Sailboat, F1 Car, Truck \\
    Electronics, Audio \& Optics & 15 & 18.1\% & Telescope, Gramophone, Piano, Radio, Microscope, Clocks \\
    Toys, Nature \& Decor        & 14 & 16.9\% & Tin Robot, Wooden Dolls, Bonsai, Cactus, Elephant, Trophy \\
    Furniture \& Seating         & 13 & 15.7\% & Office Chair, Bar Stool, Armchairs, Park Bench, Pool Table \\
    Tools \& Utility             & 10 & 12.0\% & Fire Hydrant, Hammer, Sewing Machine, Toolbox, Iron, Bucket \\
    Lighting \& Lanterns         &  8 &  9.6\% & Desk Lamps, Chandelier, Stone Lantern, Candle Lantern \\
    Kitchenware \& Tableware     &  8 &  9.6\% & Teapot, Moka Pot, Mug, Stand Mixer, Frying Pan, Spoon \\
    \midrule
    \textbf{Total}               & \textbf{83} & \textbf{100.0\%} & \\
    \bottomrule
  \end{tabular}%
  }
\end{table*}

\begin{figure}[!htbp]
    \centering
    \includegraphics[width=1\linewidth]{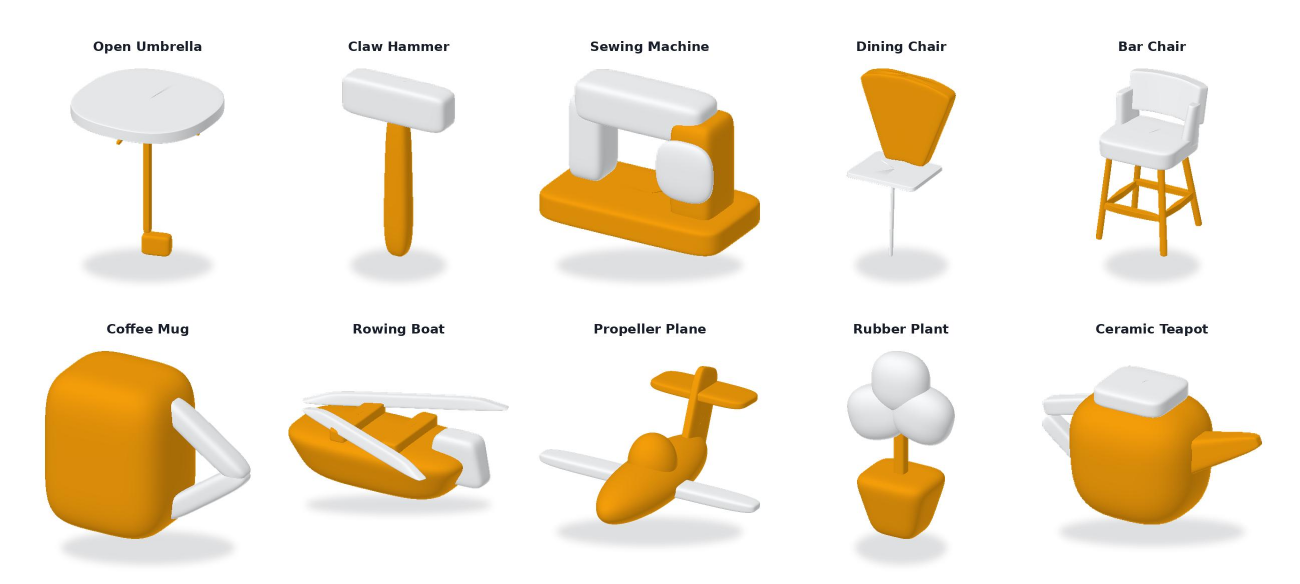}
    \caption{\textbf{Dataset assets gallery.} Ten representative assets from our 83-asset benchmark. Orange primitives designate high-control regions requiring strict geometric preservation, while white/gray primitives designate low-control regions intended for generative completion. Subtitles denote the target object concept.}
    \label{fig:supp_dataset_samples}
\end{figure}

\PAR{Dataset Samples.}
Fig.~\ref{fig:supp_dataset_samples} shows a representative selection of assets from our 83-asset benchmark across diverse categories.

\section{Vision-Language-Model Evaluation}
\label{sec:supp_vlm_evaluation}

We evaluate localized geometric structure control and appearance generation using vision-language-model judges (\textit{GPT-5.6 Sol}). Each evaluation instance presents the judge with two fixed views of a generated asset ($\sim 210^\circ$ apart) alongside the corresponding input conditions (superquadric proxies or text/image references). All evaluations are blinded: judges receive no method identifiers, filenames, or cross-sample scores.

For reproducibility, we detail below the exact prompts, rendering procedures, inference parameters used across our structure and appearance evaluations.

\subsection{Structure Control Evaluation}
\label{sec:supp_structure_judge_prompt}

To isolate geometric control from appearance quality, we render the generated assets as untextured shapes at the same scale and alignment as the input superquadrics. This prevents appearance quality from masking geometric defects.

The judge is presented with the input control shape, where region colors provide control-level labels:
\begin{itemize}[noitemsep,topsep=2pt]
    \item \textbf{Orange (High Control, $\tau=10$):} The output geometry must strictly preserve the shape, proportions, and placement of these primitives.
    \item \textbf{White/Gray (Weak Control, $\tau=3$):} The generator is expected to rely on the generative prior to generate a plausible geometry that matches the text prompt.
\end{itemize}

\PAR{Pairwise Evaluation.}
We conduct head-to-head comparisons across all 83 dataset assets between \methodname (local guidance with $\tau=10$ in high-control and $\tau=3$ in low-control regions) and SpaceControl %
uniform guidance baselines ($\tau=3$ and $\tau=10$). The judge evaluates control and prompt adherence, realism, and overall preference. The exact prompt layout and instructions are detailed in Fig.~\ref{fig:supp_structure_h2h_prompt}.

\subsection{Appearance Evaluation: Pairwise Evaluation}
\label{sec:supp_appearance_h2h_prompt}

For text-guided texture synthesis, we perform pairwise comparisons between \methodname (part-level condition routing) and appearance baselines (TRELLIS~\cite{Xiang_2025_CVPR}
 and GuideFlow3D~\cite{sayandsarkar_2025_guideflow3d}
 ). All methods generate textures on identical base geometry generated with \methodname to strictly isolate appearance generation. The complete pairwise appearance instructions are shown in Fig.~\ref{fig:supp_appearance_h2h_prompt}.

\subsection{Appearance Evaluation: Standalone Prompts}
\label{sec:supp_text_judge_prompt}

In addition to pairwise comparisons, we score text- and image-conditioned appearance generations on absolute $1$--$10$ scales.

\PAR{Text-Conditioned Evaluation.} Measures prompt faithfulness, color/material accuracy, texture detail, and overall appearance quality (Fig.~\ref{fig:supp_text_prompt}).

\PAR{Image-Conditioned Evaluation.} Measures reference-style fidelity, local routing accuracy, texture detail, and overall appearance quality (Fig.~\ref{fig:supp_image_prompt}).

\subsection{VLM-Judge Configuration and Validation}
\label{sec:supp_judge_settings}

\PAR{Inference.}
The \textit{GPT-5.6 Sol} model enforces a fixed default temperature of $T=1.0$. To eliminate stochastic judge noise, evaluations are conducted over three independent passes. For head-to-head comparisons:
\begin{itemize}[noitemsep,topsep=2pt]
    \item A/B presentation order is independently randomized across passes
    \item Final pairwise preferences are decided by majority vote across three passes (three-way ties marked undecided), yielding $0.898$ average inter-pass agreement.
\end{itemize}

\PAR{Validation Probes.}
We verify judge reliability with two calibration checks:
\begin{enumerate}[noitemsep,topsep=2pt]
    \item \textbf{Mismatched Controls:} Scoring outputs against incorrect geometric inputs reduced adherence scores by $4.43$ points, confirming sensitivity to geometric errors.
    \item \textbf{Identical Pairs:} Comparing identical renders yielded $100\%$ ``Equal'' ratings.
\end{enumerate}

Tab.~\ref{tab:supp_judge_settings} summarizes the overall judge configuration, inference hyperparameters, and calibration metrics.

\begin{table}[h!]
    \centering
    \caption{
        \textbf{VLM-judge configuration and evaluation settings.} Summary of model parameters, multi-pass aggregation rules, and rendering specifications.
    }
    \label{tab:supp_judge_settings}
    \small
    \begin{tabular}{p{0.40\linewidth}p{0.52\linewidth}}
        \toprule
        \textbf{Setting} & \textbf{Value / Specification} \\
        \midrule
        Judge Model & \textit{GPT-5.6 Sol} \\
        Sampling Temperature & Default ($T=1.0$) \\
        Number of Passes & 3 independent passes per condition \\
        Pairwise Aggregation & 2-of-3 majority vote (re-salted slot order) \\
        Standalone Aggregation & Mean across 3 passes \\
        Inter-Pass Agreement & Pairwise agreement: $0.898$ ($\text{Spearman}$) \\
        Rendered Input Views & 2 fixed views ($\sim 210^\circ$ azimuth separation) \\
        Evaluated Modalities & Structure (untextured GLB) \& Appearance (textured) \\
        \bottomrule
    \end{tabular}
\end{table}

\section{User Study Details}
\label{sec:supp_user_study}

\begin{figure}[!htbp]
    \centering
    \includegraphics[width=0.55\linewidth]{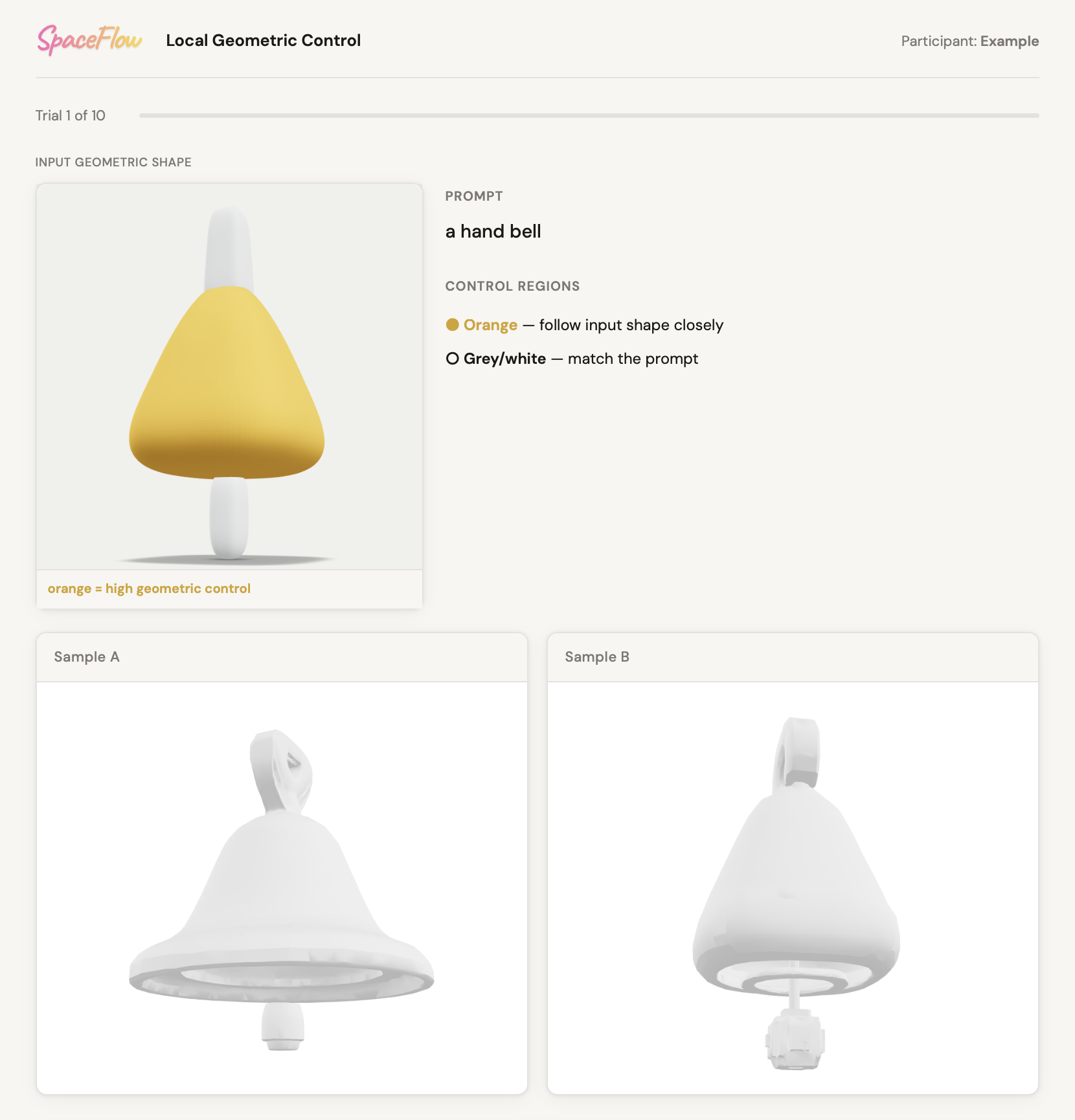}
    \caption{\textbf{User study interface screenshot.} Screenshot of the evaluation interface to study participants during a trial. The interface displays the text prompt, the input geometric control shape with color-coded region guidelines, and two randomized, anonymous outputs for comparison.}
    \label{fig:user_study_interface}
\end{figure}

This section provides additional details on the user study interface and evaluation protocol. Fig.~\ref{fig:user_study_interface} shows the interface presented to participants during the evaluation trials.

When participants entered the user study website, they saw a visual explanation of the task \methodname solves. Afterwards, for each task, participants were shown: (i)~the target text prompt (to assess realism of the generated asset); (ii)~the input geometric control shape with high-control regions highlighted in orange (to evaluate shape fidelity); and (iii)~two randomized, anonymous generated samples. Both the input shape and candidate samples were interactive 3D assets inspectable from any viewpoint.

Participants were then asked to choose between \textit{Sample A}, \textit{Sample B}, or \textit{No difference} for the following question: ``Which object looks better overall?''

\section{Additional Ablation Study}
\label{sec:additional_analyses}

We first isolate the spatial- and appearance-control components to
illustrate their individual effects. We then present additional examples
using the full \methodname pipeline, in which spatial control and
part-specific appearance conditions are applied simultaneously.

\subsection{Sensitivity to Local Control Strength}
\label{sec:local-control-continuum}

The main paper evaluates local geometric control using $\tau_{\mathrm{low}}=3$ and $\tau_{\mathrm{high}}=10$. While this setting demonstrates the contrast between high- and low-control regions, it does not characterize how the output changes between the two control endpoints. 

To quantify the effect of varying local guidance strength, we evaluate performance across $\tau_{\mathrm{low}} \in \{1, 3, 5, 7, 9\}$ on the 83-asset dataset, keeping $\tau_{\mathrm{high}} = 10$, $\alpha = 0.18$, random seeds, and the resampling schedule fixed. Following the regional Chamfer evaluation in the main text, generated mesh faces are partitioned into high- and low-control regions based on their nearest labeled input primitive surface. We compute the mean-squared Chamfer distance ($\times 10^3$) between each region and its corresponding input primitives, reporting the asset-wise mean.

\begin{figure}[!htbp]
    \centering
    \includegraphics[width=0.55\linewidth]{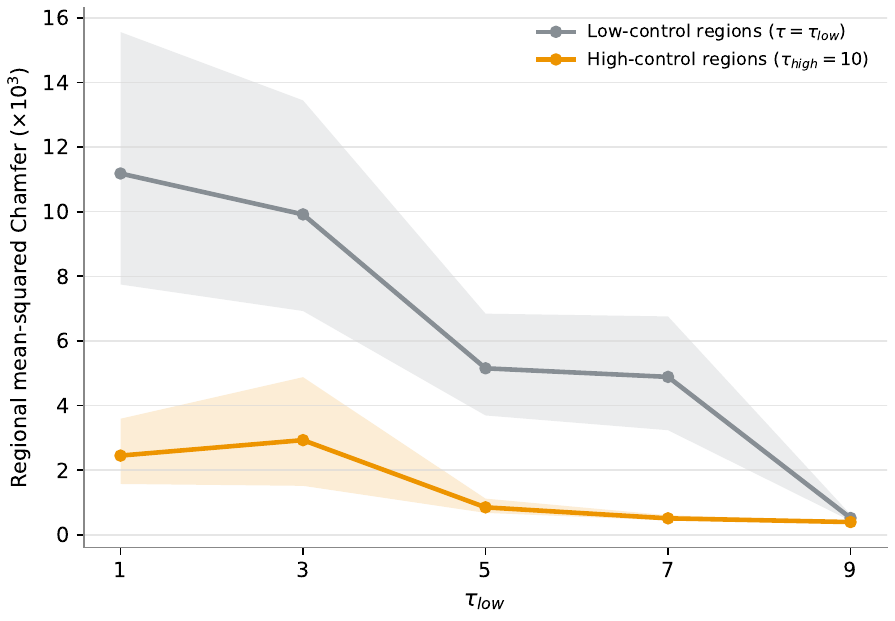}
    \caption{\textbf{Geometric deviation across local control strengths.} We fix the high-control strength at $\tau_{\mathrm{high}}=10$ and vary the low-control strength over $\tau_{\mathrm{low}}\in\{1,3,5,7,9\}$. We report the regional symmetric mean-squared Chamfer distance to the input primitive surfaces ($\times 10^{3}$; lower is better), averaged asset-wise over the 83-asset dataset. Lines show the mean and shaded bands indicate 95\% bootstrap confidence intervals. Low-control regions (gray) deviate most under weak control, and the gap to high-control regions (orange) closes as the two control strengths converge.}
    \label{fig:local-control-chamfer-sweep}
\end{figure}

\begin{figure}[!htbp]
    \centering
    \includegraphics[width=\linewidth]{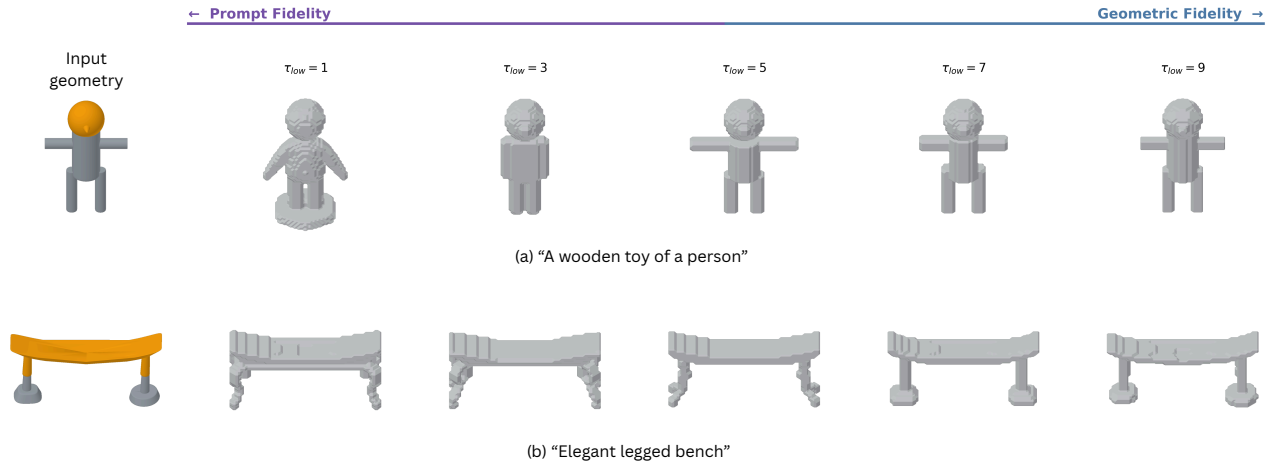}
    \caption{\textbf{Qualitative local control strength continuum.} We fix the protected primitives at $\tau_{\mathrm{high}}=10$ and vary the editable-region strength over $\tau_{\mathrm{low}}\in\{1,3,5,7,9\}$. Orange input primitives denote high control and gray input primitives denote low control. Lower values permit greater prompt-driven geometric freedom, whereas increasing $\tau_{\mathrm{low}}$ progressively strengthens adherence to the input geometry. Outputs are shown without texture to isolate geometric behavior.}
\label{fig:local-control-qualitative-sweep}
\end{figure}

As shown in Fig.~\ref{fig:local-control-chamfer-sweep}, a lower $\tau_{\mathrm{low}}$ gives the generator substantial freedom to deviate from the input primitives, while increasing $\tau_{\mathrm{low}}$ progressively constrains the shape to the input geometry until both curves converge at $\tau_{\mathrm{low}} \approx \tau_{\mathrm{high}}$. Notably, deviation in the high-control regions also decreases slightly as $\tau_{\mathrm{low}}$ increases, despite $\tau_{\mathrm{high}}$ being held constant. This reflects the global nature of the generative flow backbone: large shape modifications in low-control regions exert a slight geometric pull across part boundaries onto neighboring high-control regions. 

Additionally, the higher mean and variance observed in the high-control region at low $\tau_{\mathrm{low}}$ are partly due to a boundary effect of our metric definition. Under low control, the model can freely generate novel parts that may emerge near the boundary between regions. Because mesh faces are partitioned according to their nearest input primitive, some of these newly generated structures may be mapped to adjacent high-control primitives despite having no counterpart in the geometric input, thereby increasing the regional Chamfer distance. As $\tau_{\mathrm{low}}$ increases, unconstrained generation is reduced, lowering both the metric mean and variance.

Fig.~\ref{fig:local-control-qualitative-sweep} illustrates this behavior for two representative inputs. At low $\tau_{\mathrm{low}}$, the model has greater freedom to reinterpret the geometric input according to the prompt. At higher values, the output increasingly follows the input geometry, providing a continuous transition from prompt-driven generation to geometric fidelity.

\subsection{Spatial Feature-Distance Visualization}
\label{sec:supp_feature_distance}

\begin{figure}[!htbp]
    \centering
    \includegraphics[width=0.55\linewidth]{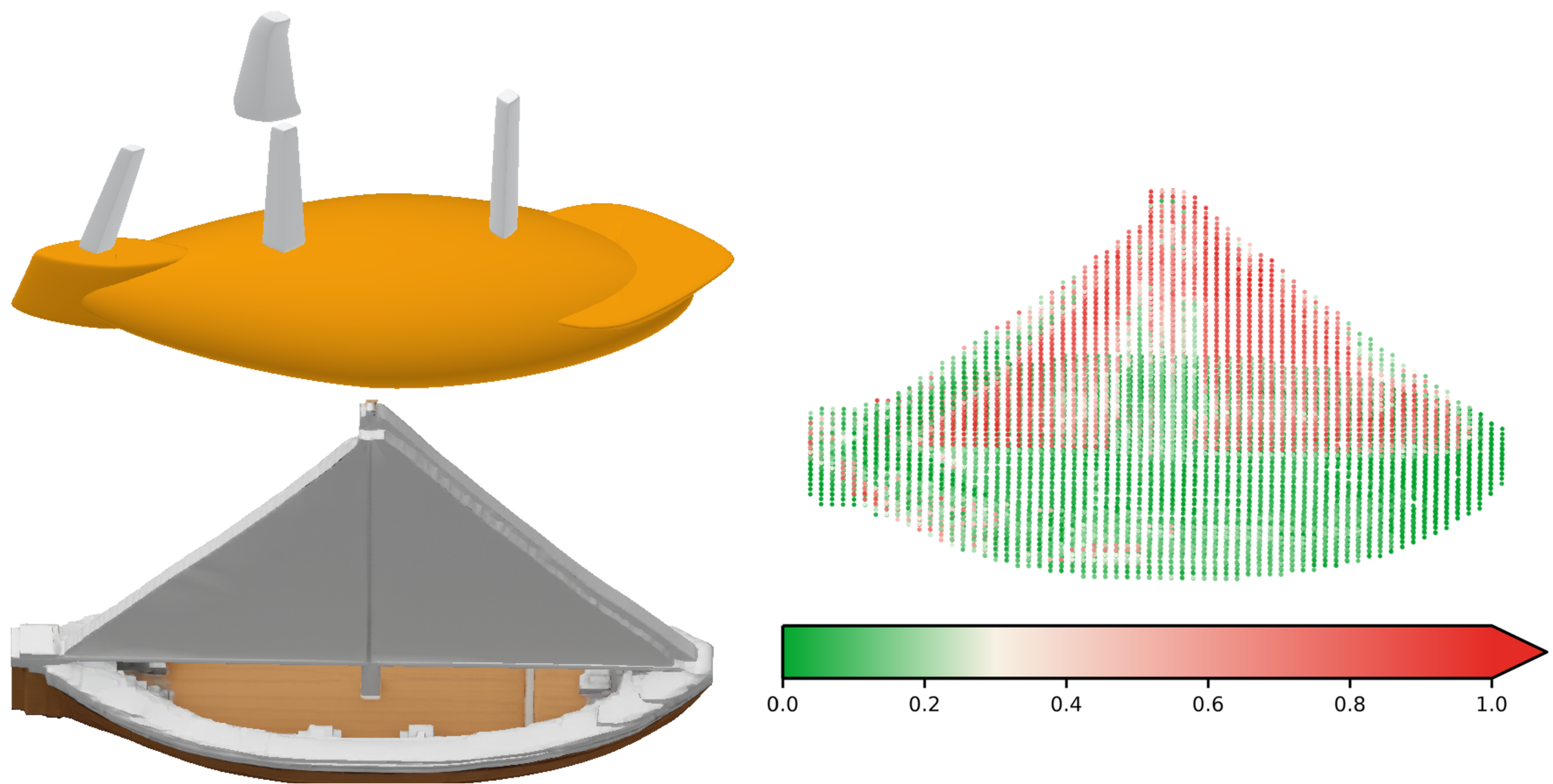}
    \caption{
        \textbf{Visualization of spatial feature distance.}
        Left: input superquadrics (top; orange denotes high control and
        gray denotes low control) and the asset generated by \methodname
        (bottom). Right: voxel-wise cosine distance between the generated
        asset's PartField features and those of the nearest spatial match
        in the uniformly controlled ($\tau=10$) reference. Feature
        deviations concentrate primarily in the low-control region.
    }
    \label{fig:cosineship}
\end{figure}

The feature-distance metric reported in the main paper measures regional deviation from an asset generated with uniform high-control. For each generated voxel, we find its nearest spatial match in the reference generated with $\tau=10$ and compute the cosine distance between their normalized PartField features. 

Fig.~\ref{fig:cosineship} visualizes this metric for a representative example. Larger distances occur primarily in the low-control region, while the high-control region remains closer to the reference, illustrating that \methodname localizes generative variation to regions where control is relaxed.

\subsection{Primitive-to-Part Assignment Visualizations}
\label{sec:supp_primitive_routing}

\begin{figure}[!htbp]
    \centering

    \newcommand{\routecell}[3]{%
        \begin{subfigure}[t]{0.24\textwidth}
            \centering
            \includegraphics[width=0.48\linewidth]{figs/routing_assets/#1}%
            \hspace{0.1em}%
            \includegraphics[width=0.48\linewidth]{figs/routing_assets/#2}%
            \caption{\scriptsize ``#3''}
        \end{subfigure}%
    }

    \routecell{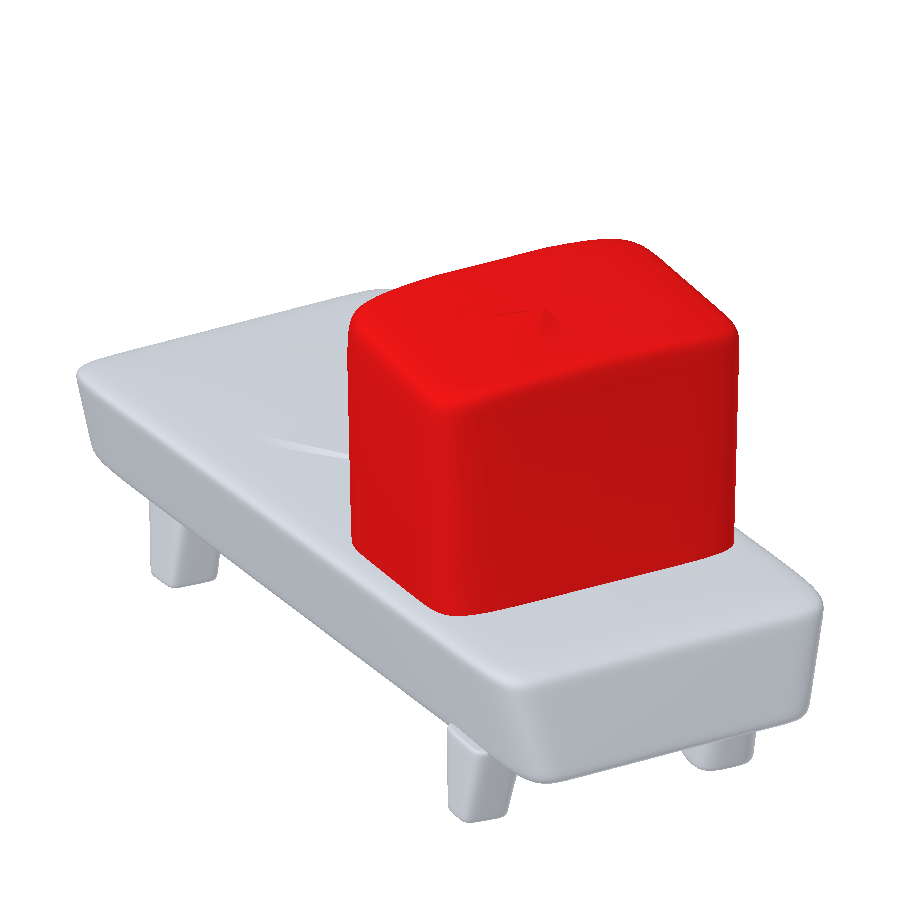}
              {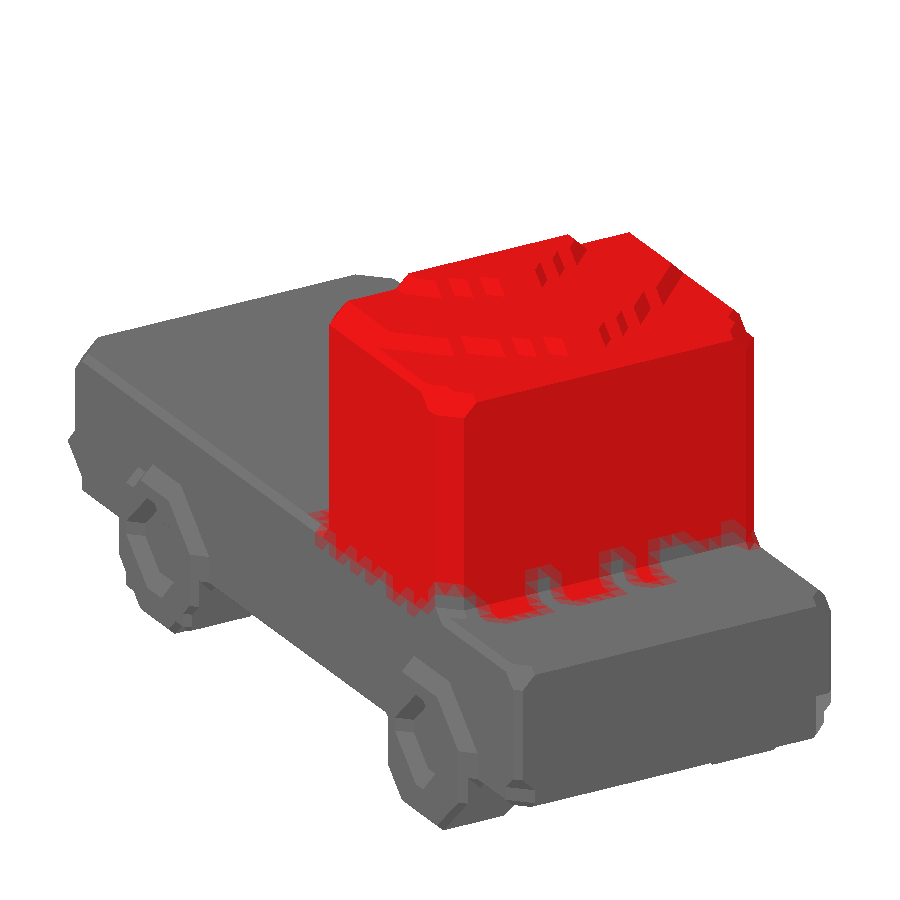}
              {red metal truck with a white cabin and blue glass}\hfill
    \routecell{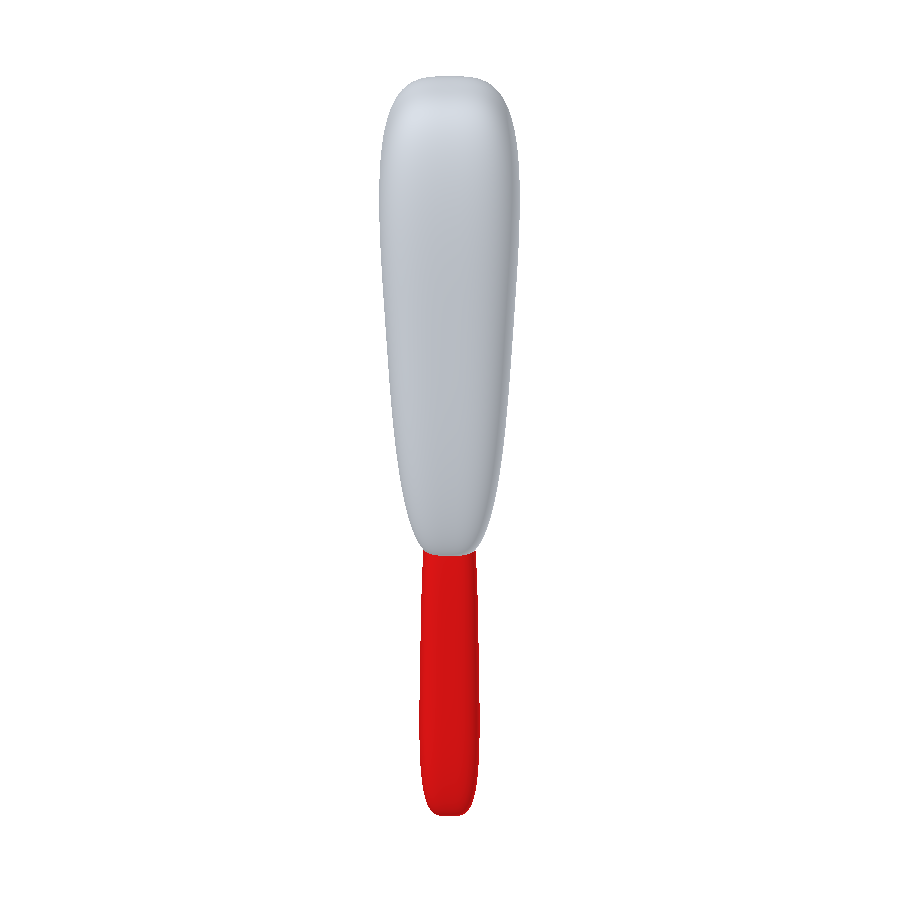}
              {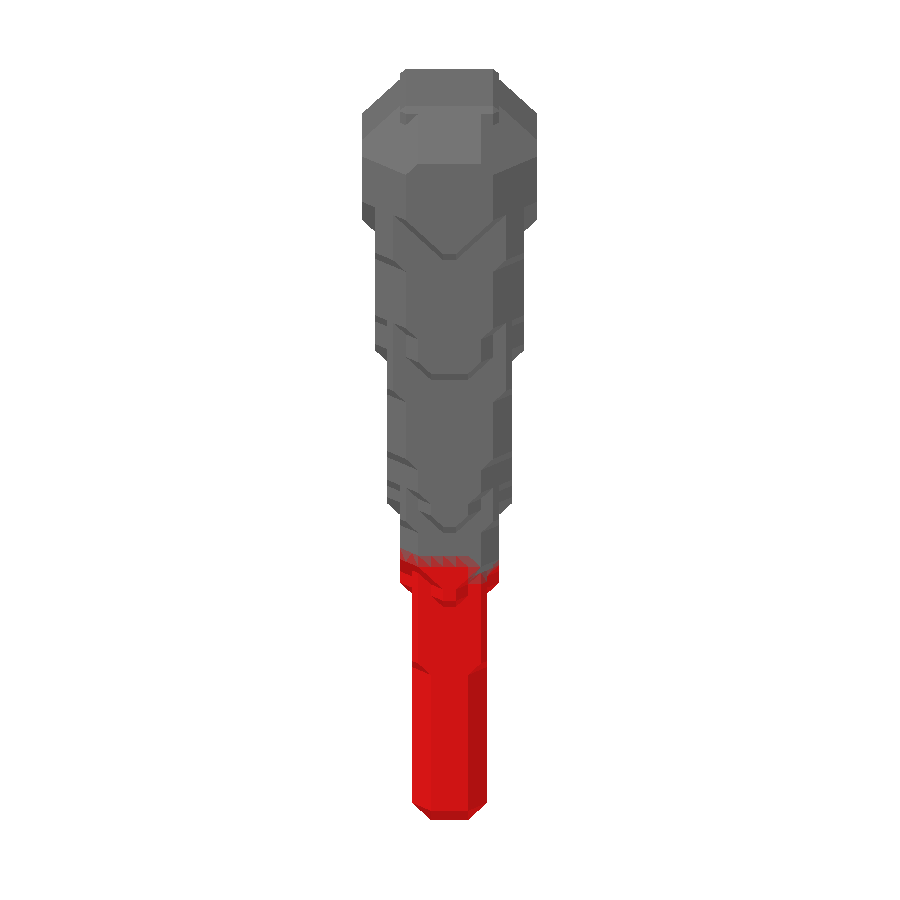}
              {light wood baseball bat with a black grip}\hfill
    \routecell{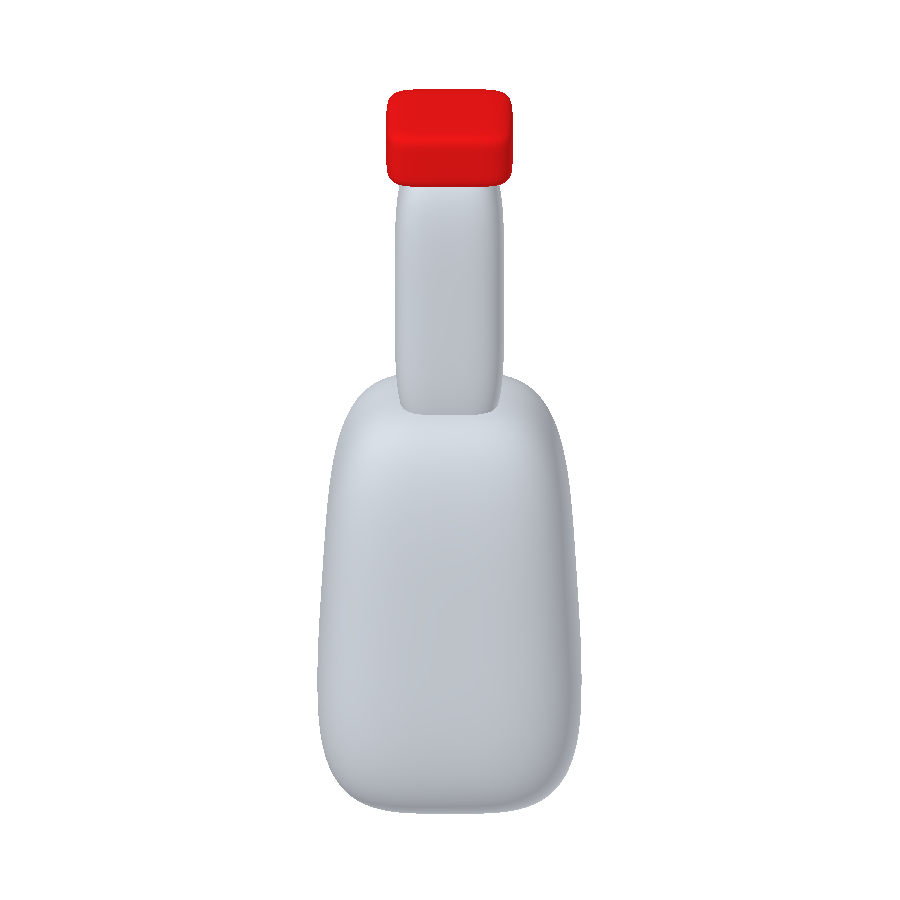}
              {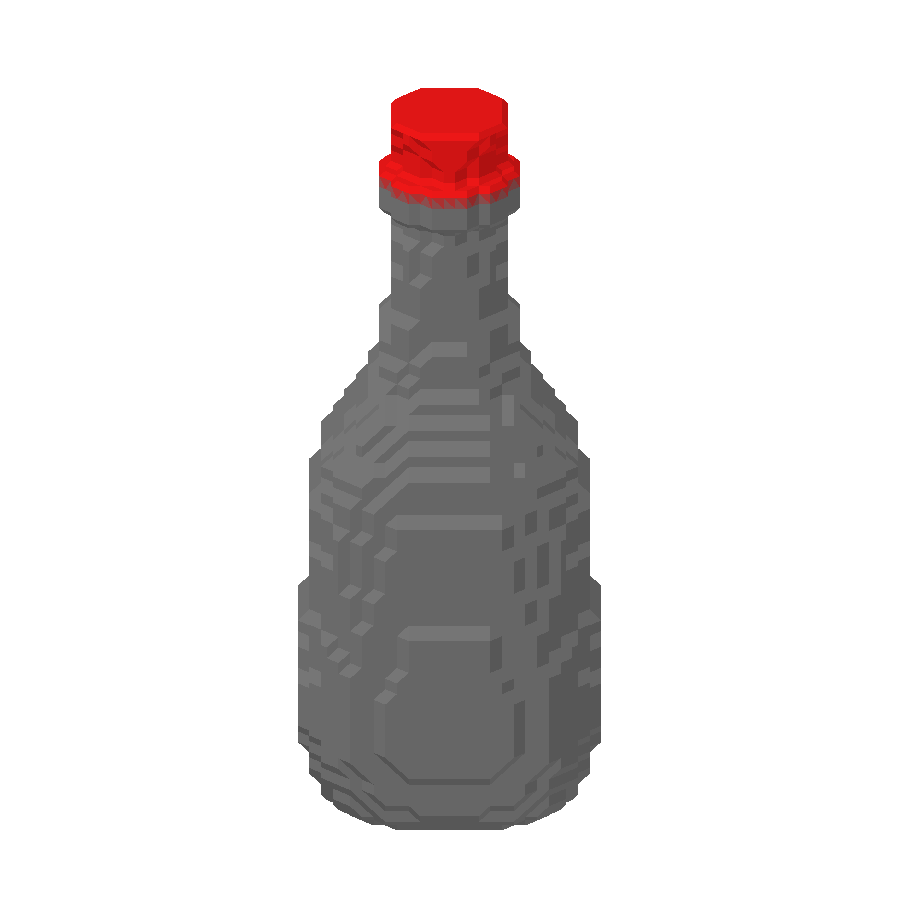}
              {green glass bottle with a brown cork stopper}\hfill
    \routecell{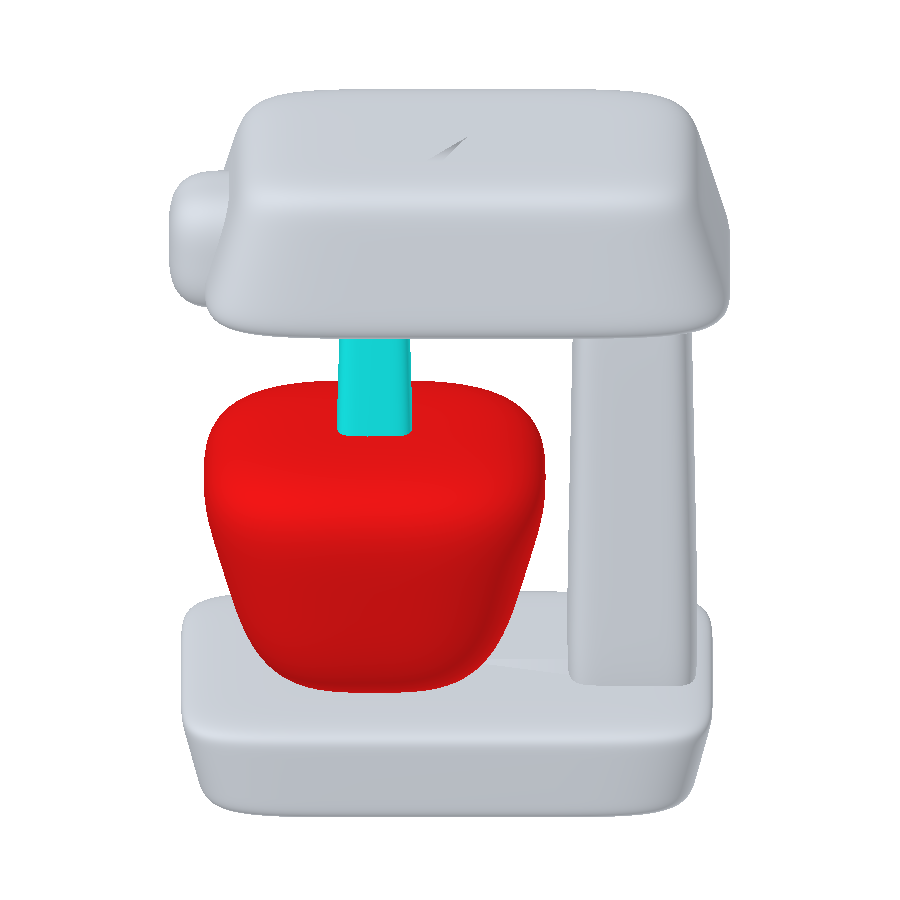}
              {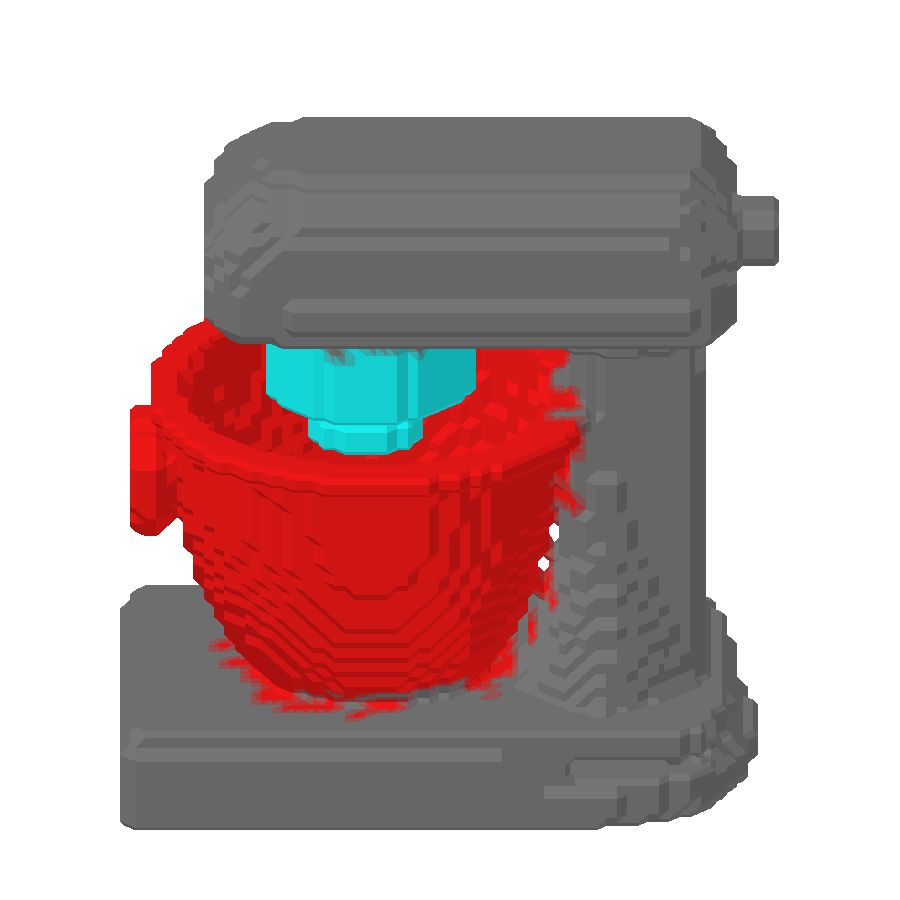}
              {cream enamel mixer with a silver metal bowl and beater}
    \par\vspace{0.8em}

    \routecell{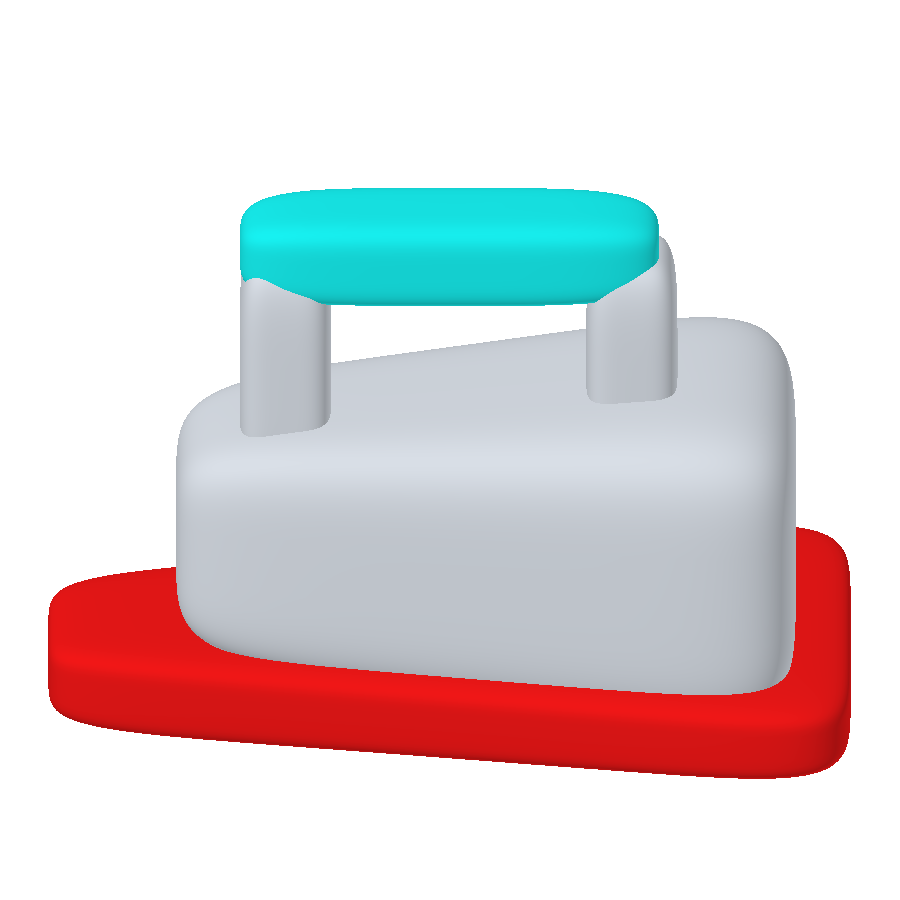}
              {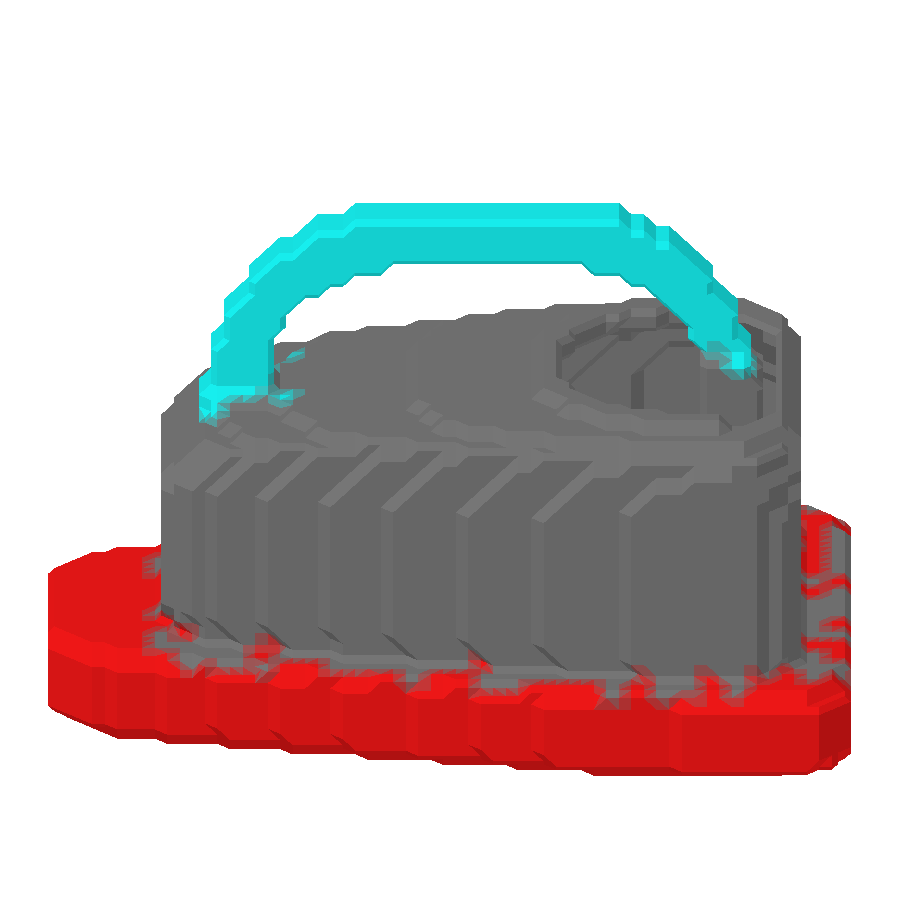}
              {blue metal steam iron with a silver soleplate and black handle}\hfill
    \routecell{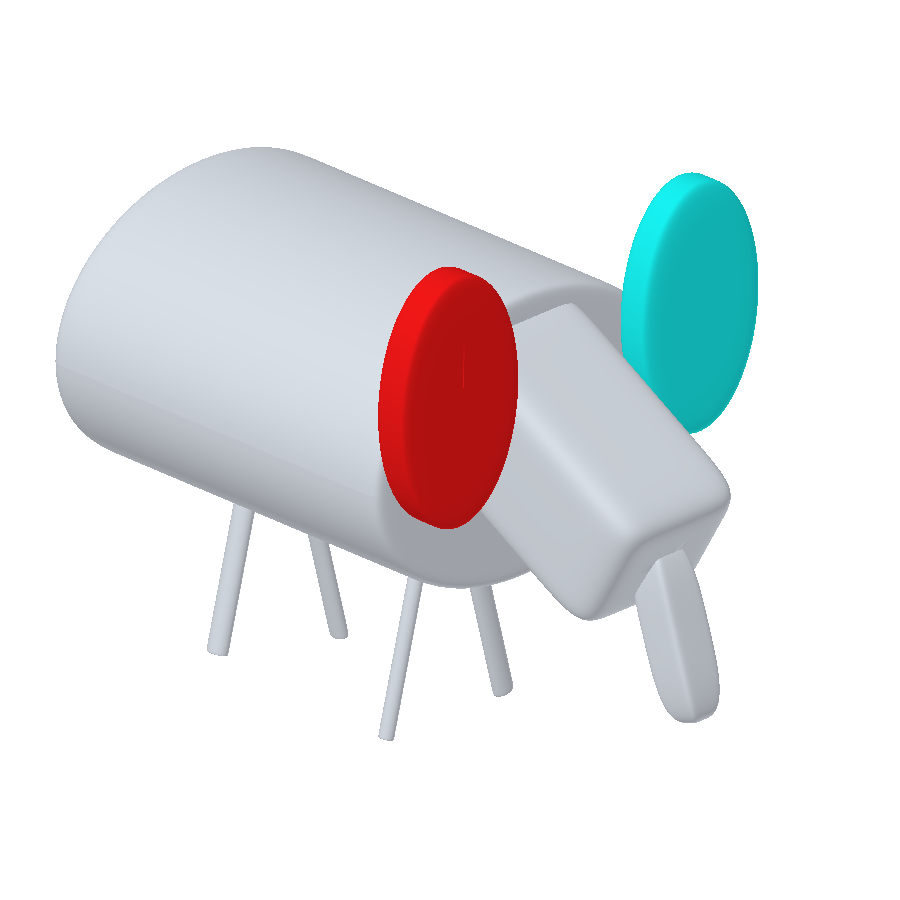}
              {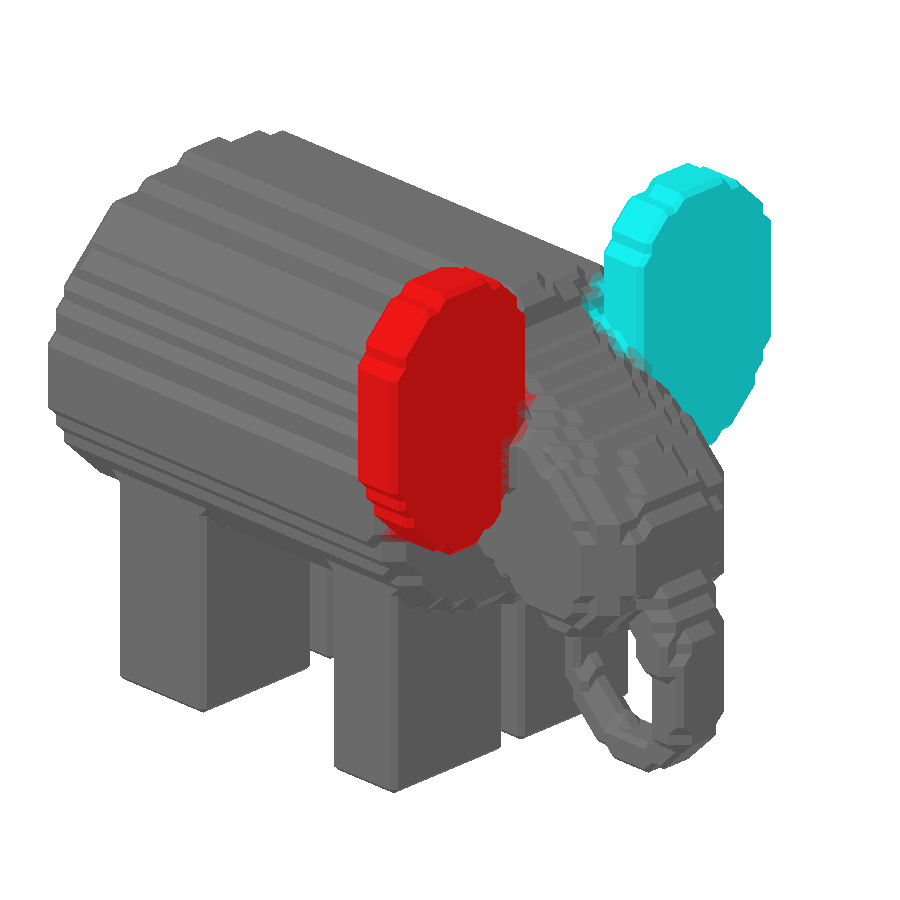}
              {blue wooden toy elephant with pink wooden ears}\hfill
    \routecell{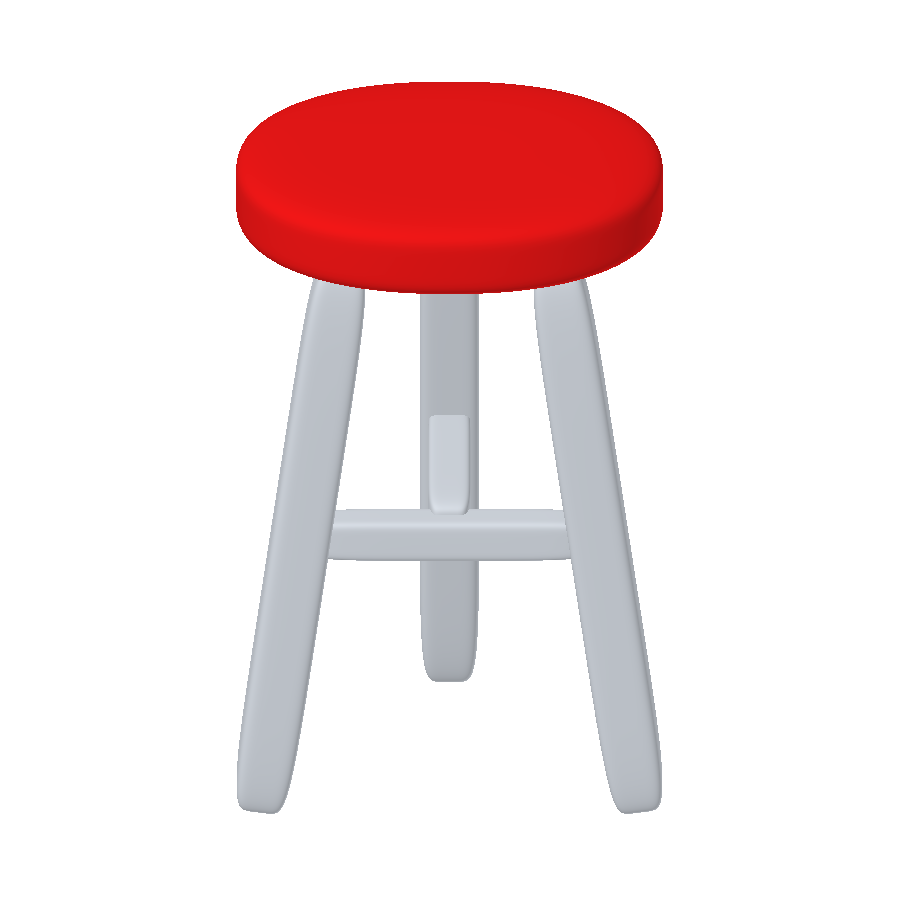}
              {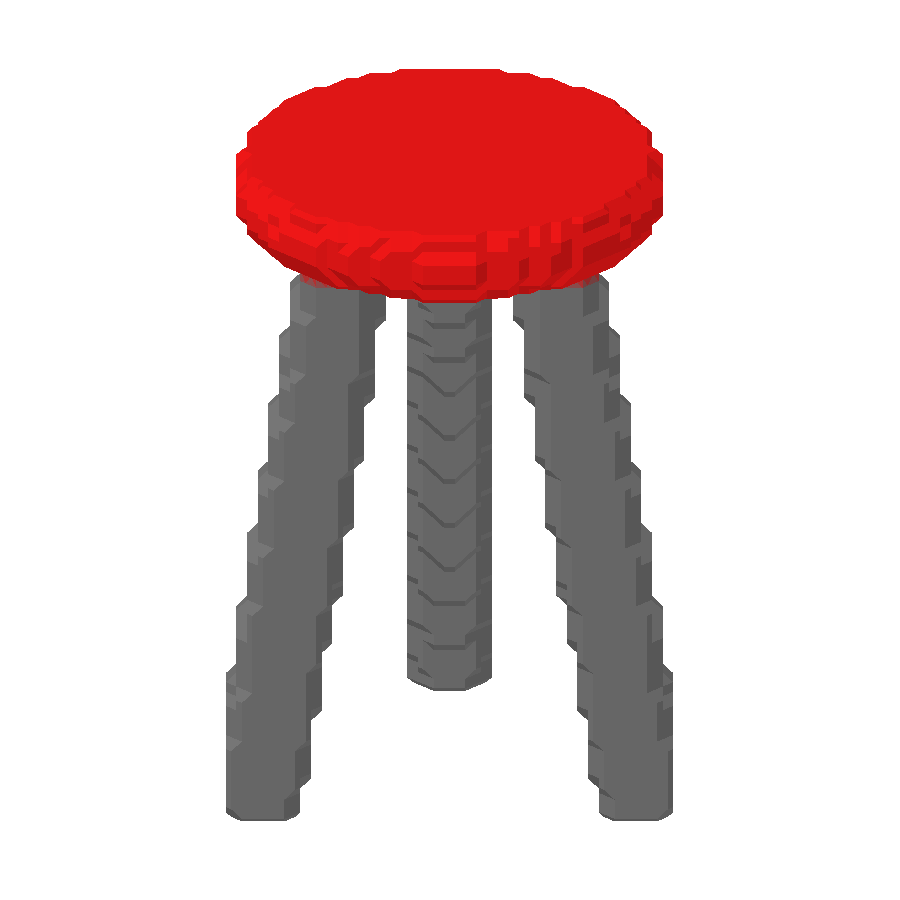}
              {silver metal stool with a green fabric seat}\hfill
    \routecell{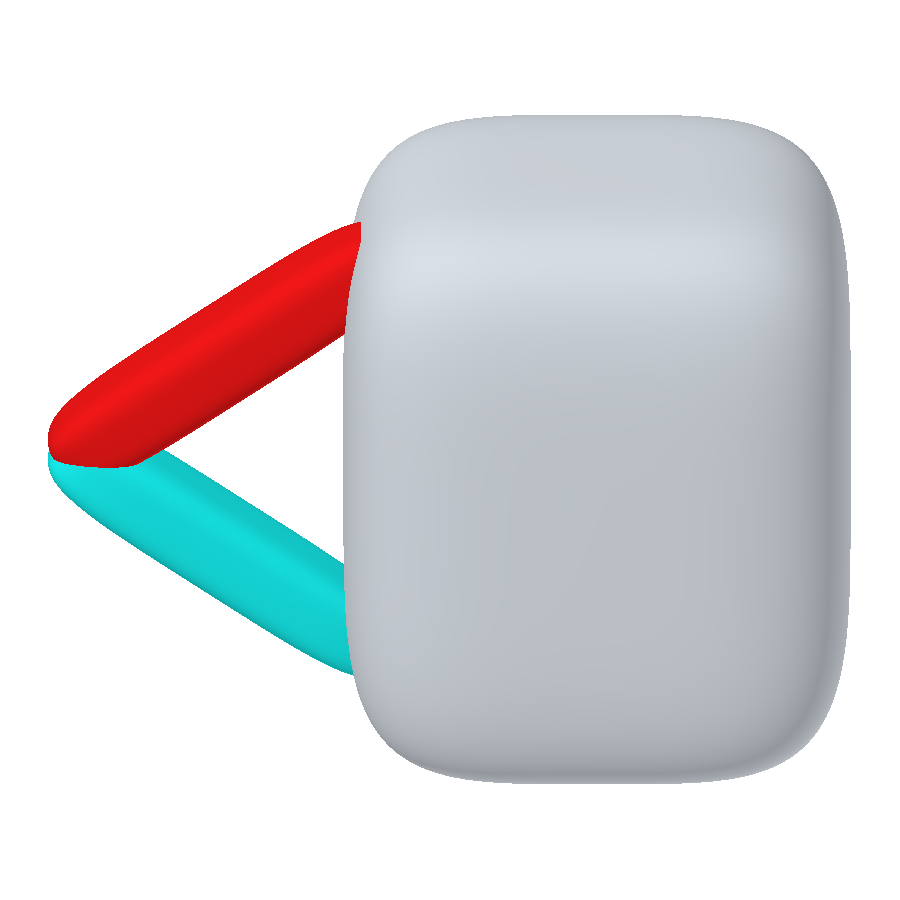}
              {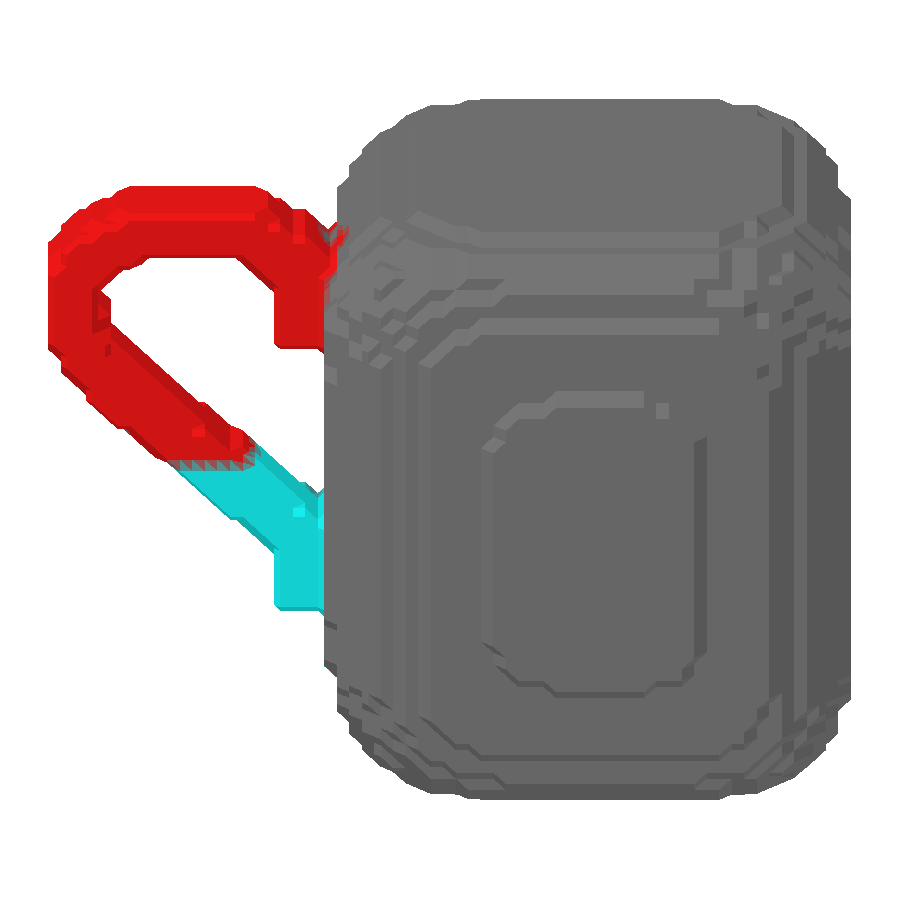}
              {white enamel mug with a blue handle}

    \vspace{0.2em}
    \caption{\textbf{Primitive-to-part appearance routing.} In each pair (a--h), we show the input structure (left) and the resulting $32^3$ routing volume (right). The mapping is defined color-wise: regions sharing the same color between the input primitives and the generated structure represent a 1-to-1 mapping to that condition (with gray assigned to the global prompt). Subcaptions provide the flattened prompt containing all localized appearance cues.}
    \label{fig:supp_primitive_routing}
\end{figure}

 In Fig.~\ref{fig:supp_primitive_routing}, we visualize local appearance routing. In each pair, the left image is the geometric input and the right image is the derived routing volume from generated structure in $32^3$ voxel space; colored regions attend to local appearance cues and gray regions attend to the global condition. We use $K_{\mathrm{PF}}=30$ PartField clusters.

\section{Limitations}
\label{sec:limitations}

\begin{figure}[!htbp]
    \centering
    \includegraphics[width=0.55\linewidth]{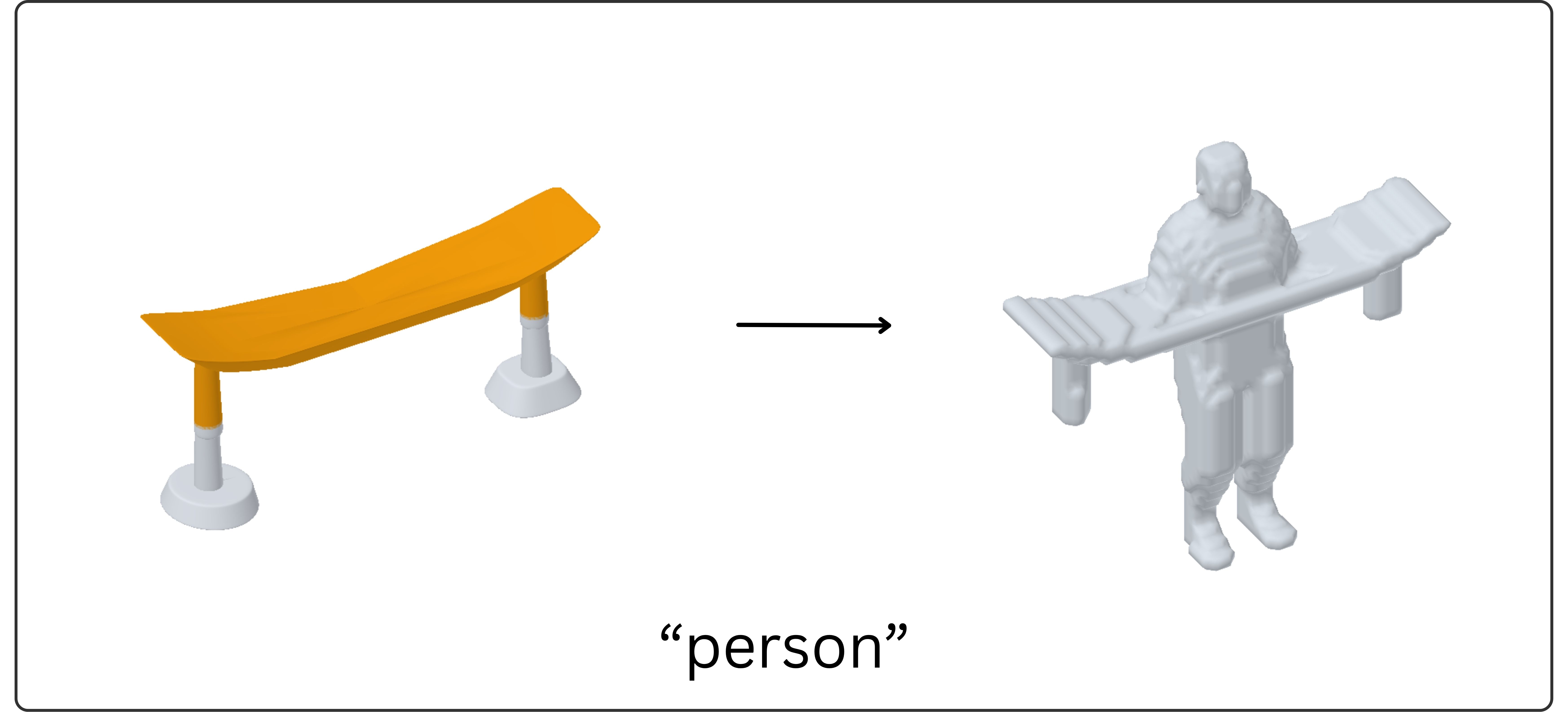}
    \caption{\textbf{Semantic-geometric conflict.} Synthesizing a ``person'' over an input ``bench'' structure (left). Lacking a joint prior for conflicting modalities, the model generates unnatural distortions by embedding the bench's high-control geometry inside the generated ``person'' (right).}
    \label{fig:geometry_limitation}
\end{figure}

\noindent\textbf{Semantic-Geometric Conflicts.}
A primary failure mode in our geometric control module occurs when the text prompt contradicts the input geometry (\eg, requesting a \emph{``standing person''} over a \emph{``bench''} structure). In such cases, the text-conditioning pathway attempts to generate human features while the geometric guidance enforces the shape of the high-control regions. Because the underlying generator lacks a joint prior for geometric and textual combinations, the model struggles to reconcile both signals. This typically results in unnatural geometric distortions as the model attempts to synthesize the text-described asset around the constrained high-control regions (see Fig.~\ref{fig:geometry_limitation}).

\begin{figure}[!htbp]
    \centering
    \newcommand{\fcrow}[4]{%
        \begin{subfigure}[b]{0.48\linewidth}
            \centering
            \includegraphics[height=#1,keepaspectratio]{figs/failure_cases/#2}
            \caption{#4 -- routing}
        \end{subfigure}\hfill
        \begin{subfigure}[b]{0.48\linewidth}
            \centering
            \includegraphics[height=#1,keepaspectratio]{figs/failure_cases/#3}
            \caption{#4 -- result}
        \end{subfigure}%
    }
    \fcrow{0.105\textheight}{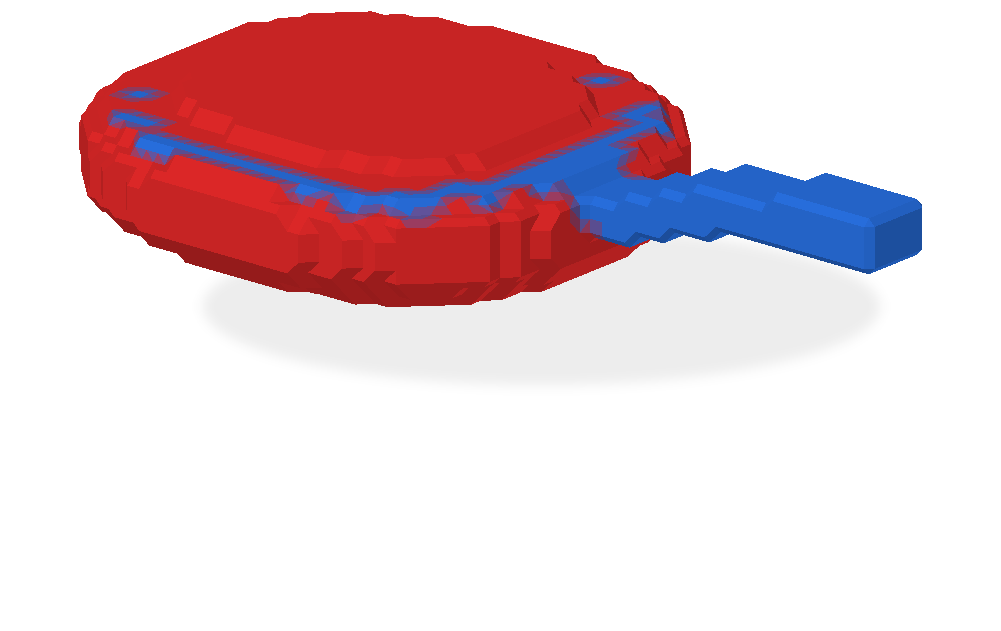}{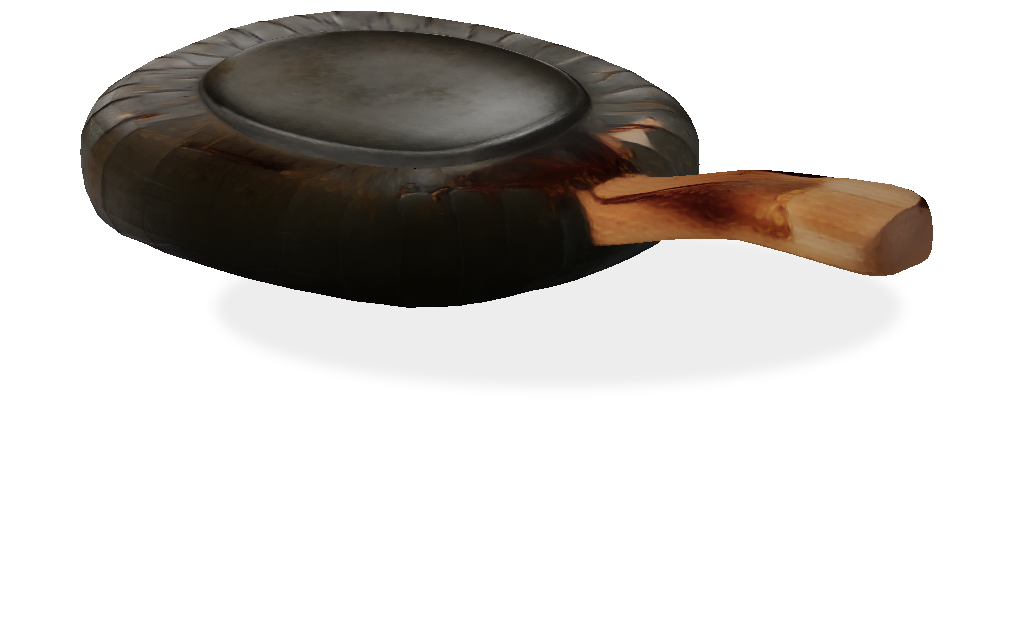}{Frying pan}

    \vspace{0.6em}

    \fcrow{0.185\textheight}{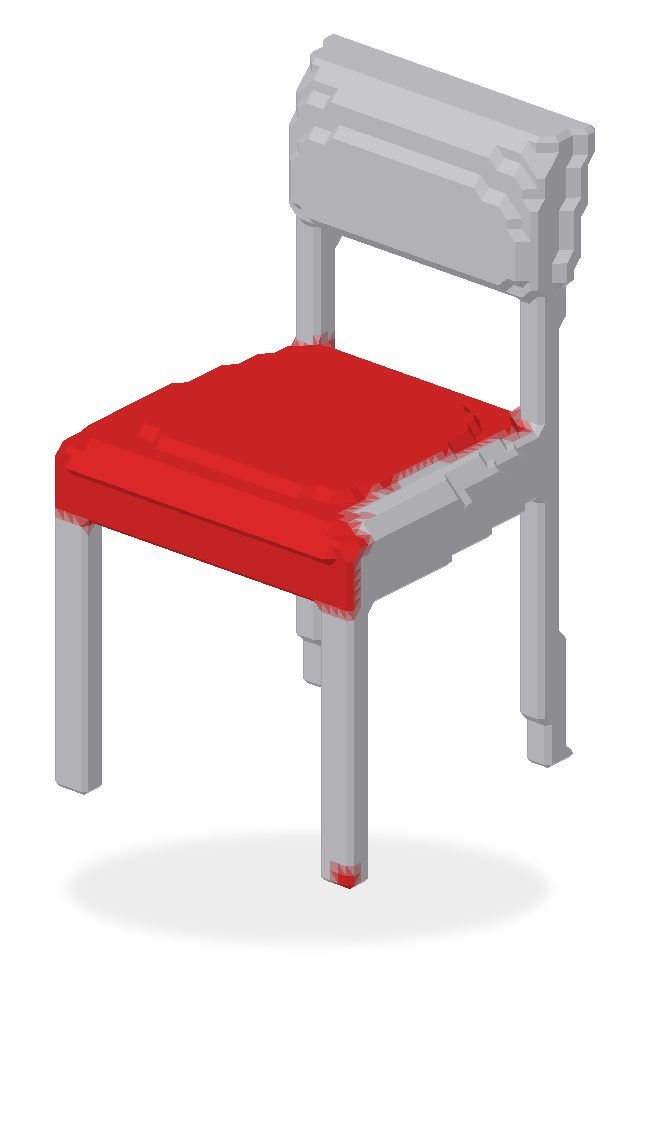}{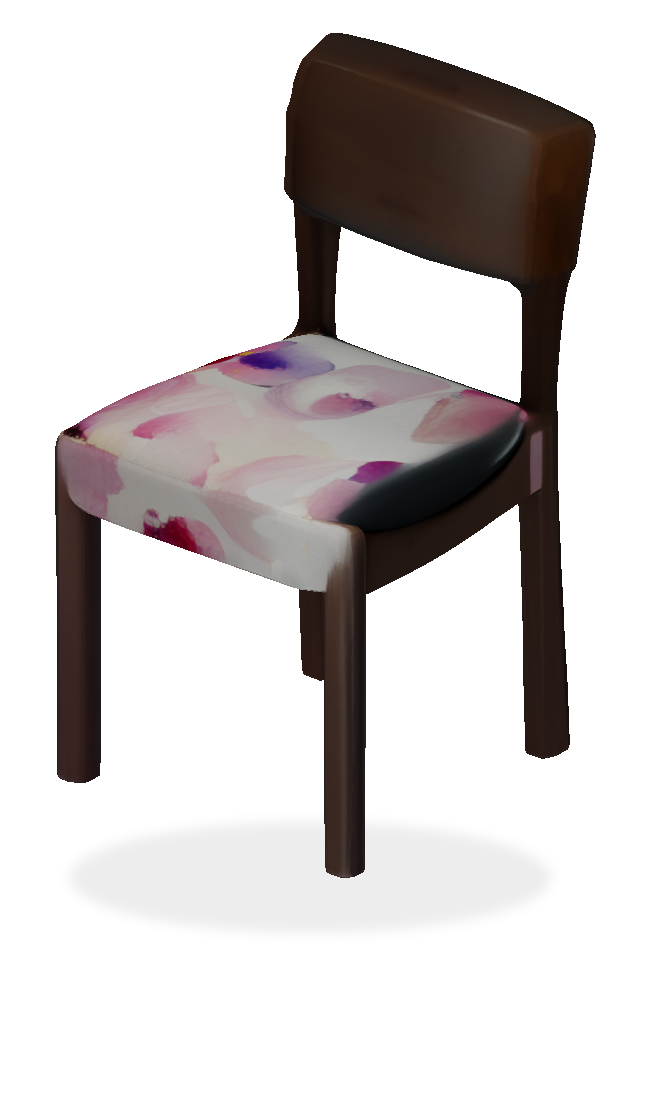}{Wood chair}
        \caption{
        \textbf{Failure cases of local appearance routing.}
        For each asset we show the derived routing volume (left; colors
        denote distinct appearance conditions) and the generated result
        (right). \textbf{(a--b)} The handle cluster wraps around the rim,
        so the wooden handle appearance bleeds onto the metal pan body.
        \textbf{(c--d)} The seat cluster covers only part of the cushion,
        leaving the rest textured by the competing condition.
    }

    \label{fig:failure_cases}
\end{figure}

\noindent\textbf{Misrouted Appearance Cues.}
Our appearance module inherits the granularity of the underlying part
decomposition: the routing volume is derived from PartField clusters in
$32^3$ voxel space, and each cluster is assigned to a single appearance
condition. When a cluster boundary does not coincide with the true
material boundary of the generated asset, the corresponding appearance
cue is applied to the wrong voxels, and the error is directly visible in
the final texture. Fig.~\ref{fig:failure_cases} shows two representative
cases. In the frying pan, the routing volume assigns the pan body and
the handle to different conditions, but a thin band of handle voxels
leaks along the rim; the wooden handle appearance consequently bleeds
onto the metal body. In the chair, the seat cluster does not fully cover
the cushion surface, so only part of the seat receives the intended
upholstery condition while the remaining voxels are textured by the
competing condition, yielding a visibly discontinuous seat. Both
failures are localized to the routing stage rather than the generator:
the geometry is faithful to the input primitives, and only the
region-to-condition assignment is incorrect. Finer part decompositions
or soft, distance-weighted routing boundaries are a natural direction
for mitigating this effect.

\section{Additional Qualitative Results}
\label{sec:supp_qualitative}

\begin{figure}[!htbp]
    \centering
    \includegraphics[width=1\linewidth]{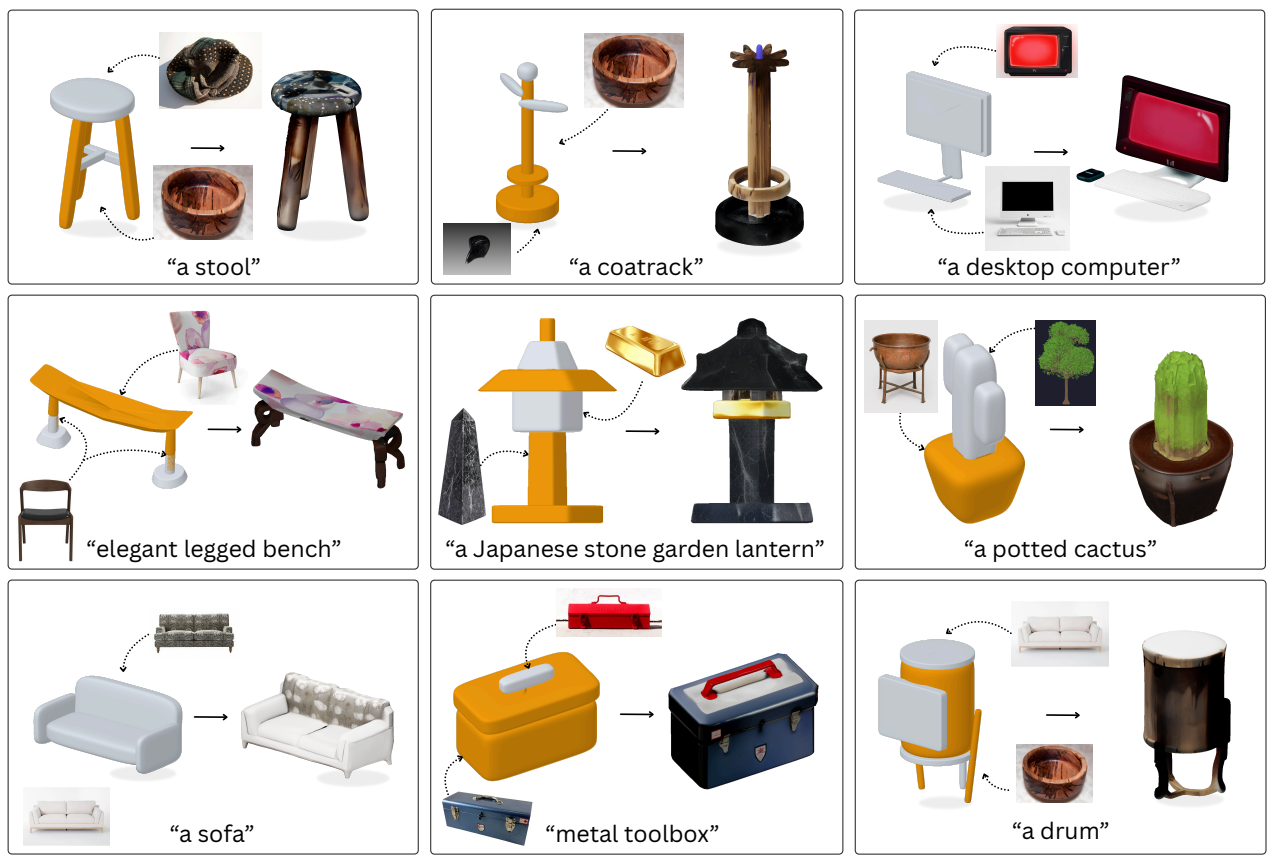}
    \caption{
    \textbf{Additional qualitative results for image-conditioned appearance control.} Each panel shows an input geometric structure (left), the object-level text prompt (bottom), and two reference images specifying the desired material, texture or style for the object or for particular parts. Dotted arrows indicate the part-level associations between the reference images and the corresponding target regions. We recommend zooming in to inspect the fine-grained structural and appearance details; for example, in the metal-toolbox example, the generated asset includes the small front latches visible in the reference image and renders them with a consistent metallic appearance. These examples illustrate how \methodname transfers reference-driven appearance cues to the corresponding regions while retaining the overall structure of the input object.}
    \label{fig:imageconditioned-examples}
\end{figure}

Fig.~\ref{fig:imageconditioned-examples} shows additional qualitative results for image-conditioned appearance control. In each example, the geometric input is paired with two reference images that specify the material or style for designated parts (indicated by arrows). Across diverse object categories, \methodname faithfully transfers these appearance cues to the corresponding regions while preserving the underlying geometry.

\begin{figure}[!htbp]
    \centering
    \newcommand{\expcell}[3]{%
        \begin{subfigure}[t]{0.485\textwidth}
            \centering
            \includegraphics[width=0.49\linewidth]{figs/table_experiments_sq/#1}%
            \hfill
            \includegraphics[width=0.49\linewidth]{figs/table_experiments/#2}%
            \caption{``#3''}
        \end{subfigure}%
    }

    \expcell{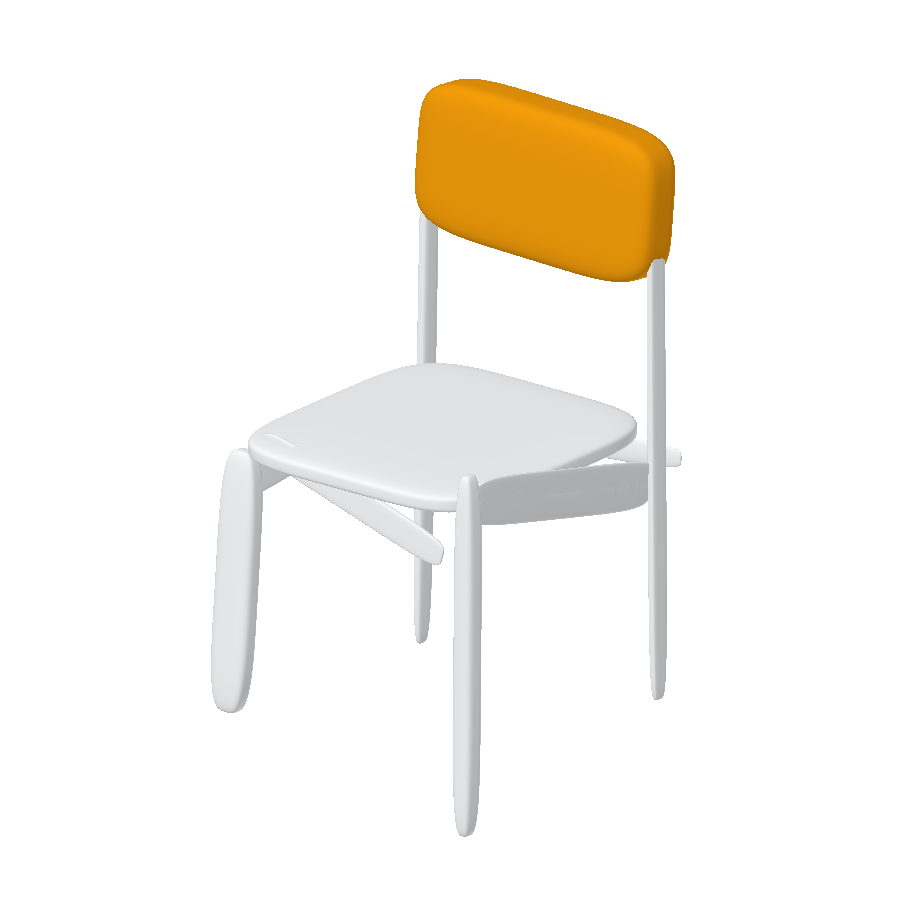}{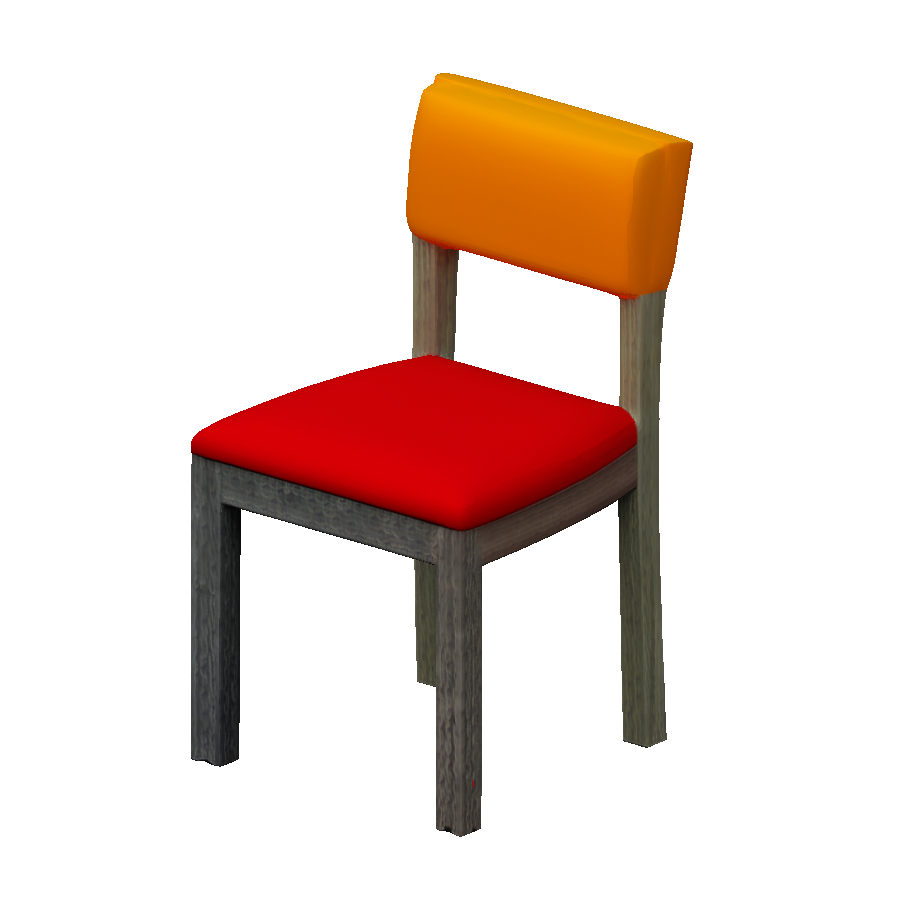}{wood chair with straight legs and a red cushioned seat and orange backrest}\hfill
    \expcell{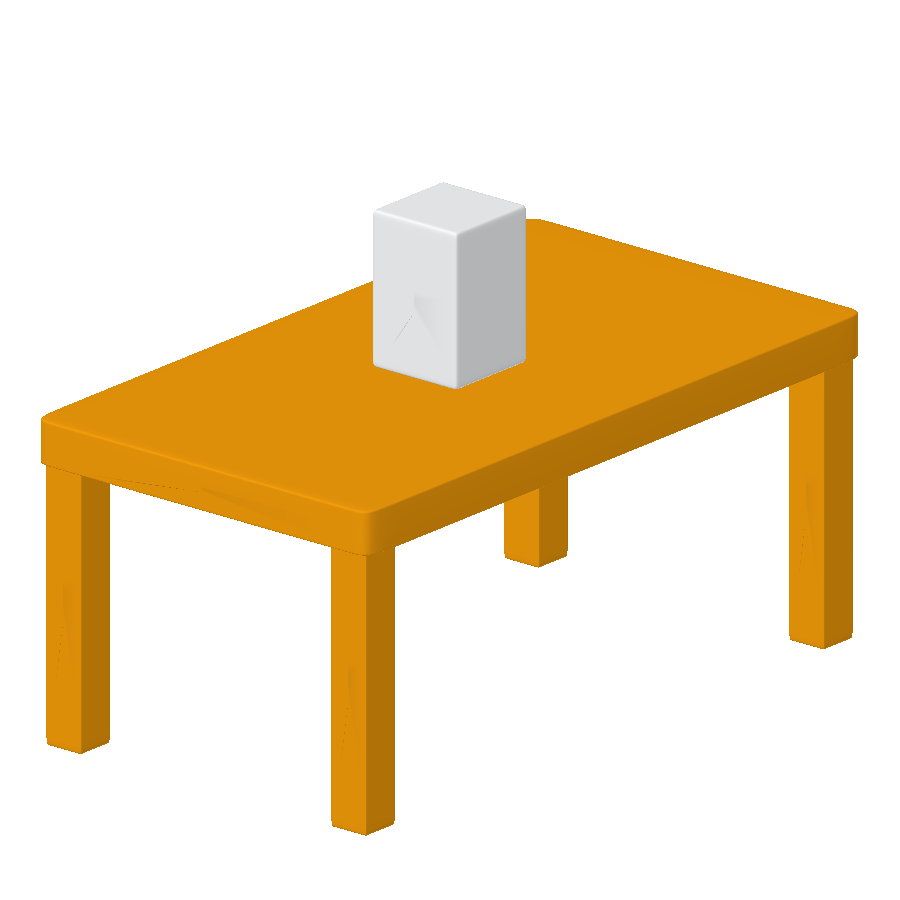}{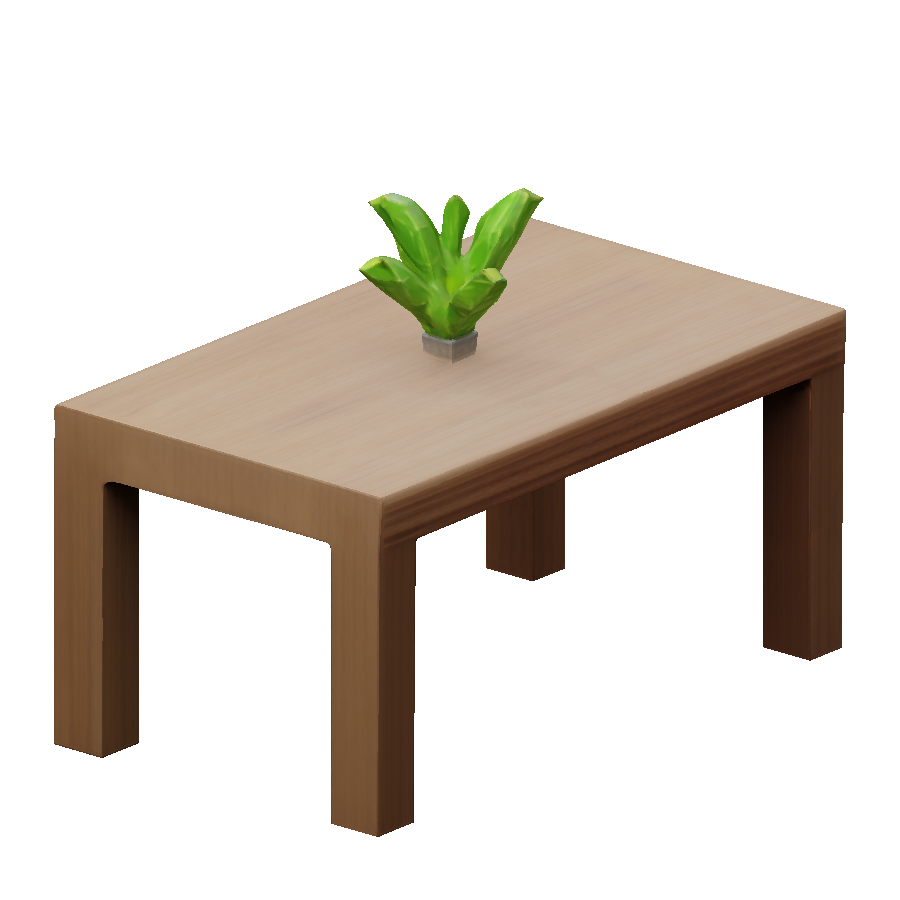}{A table with a plant}
    
    \vspace{0.9em}

    \expcell{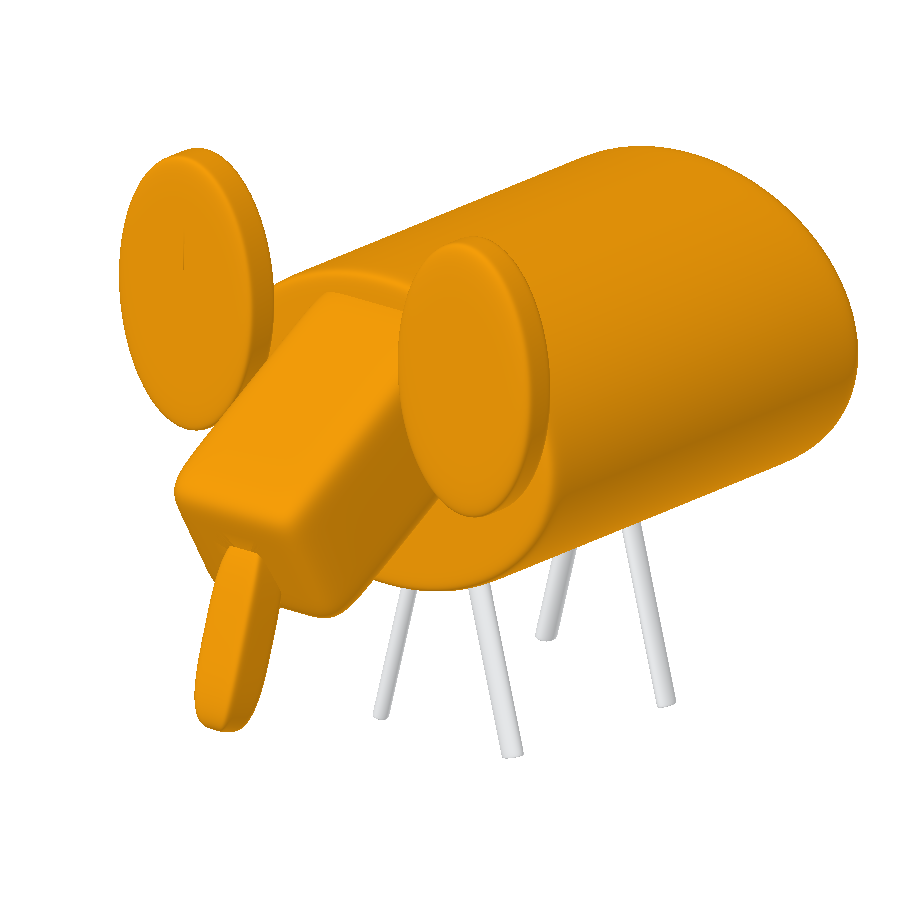}{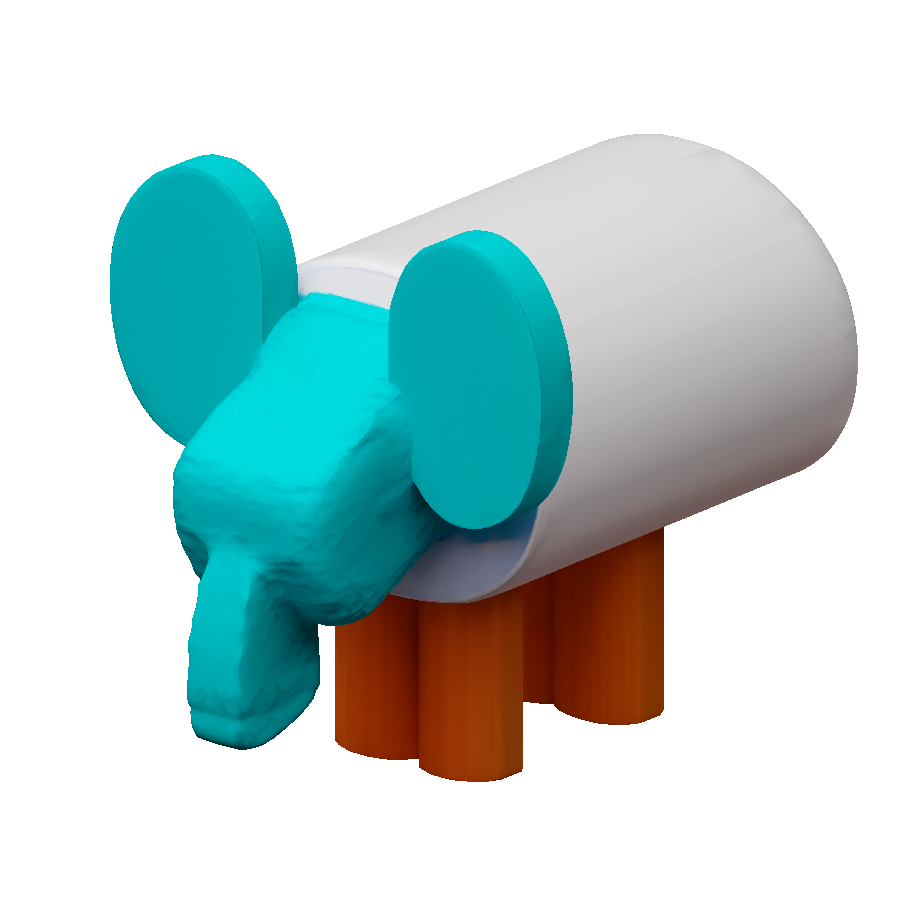}{white elephant toy with orange legs and cyan head}\hfill
    \expcell{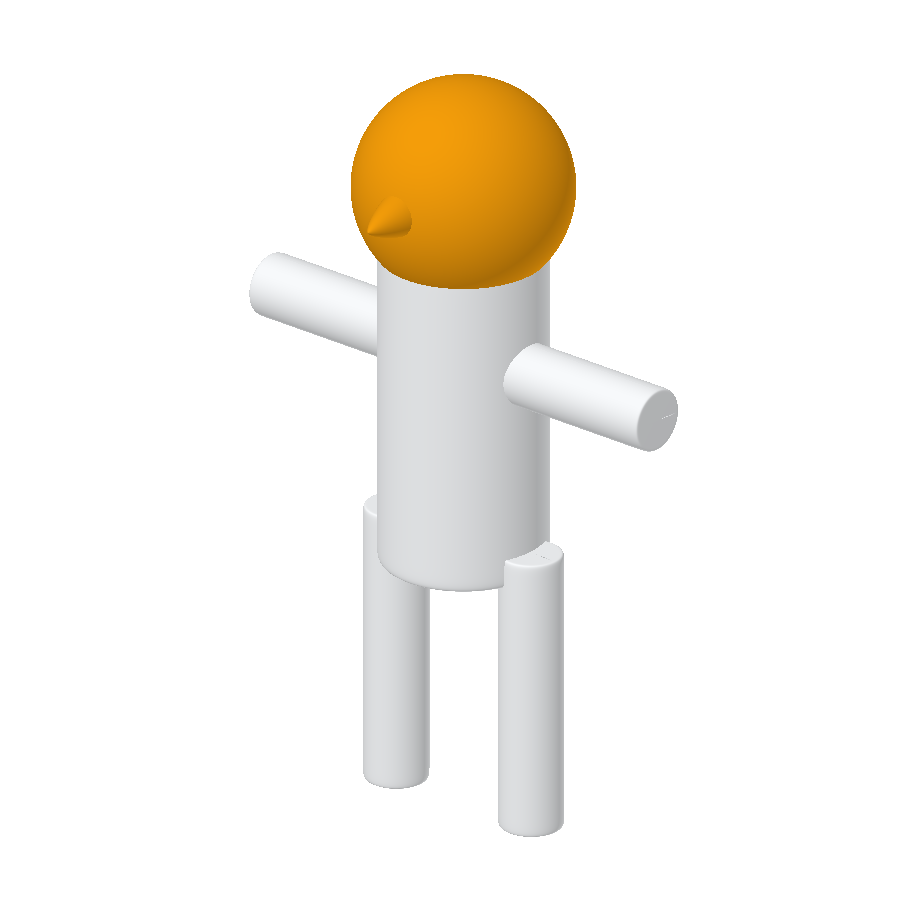}{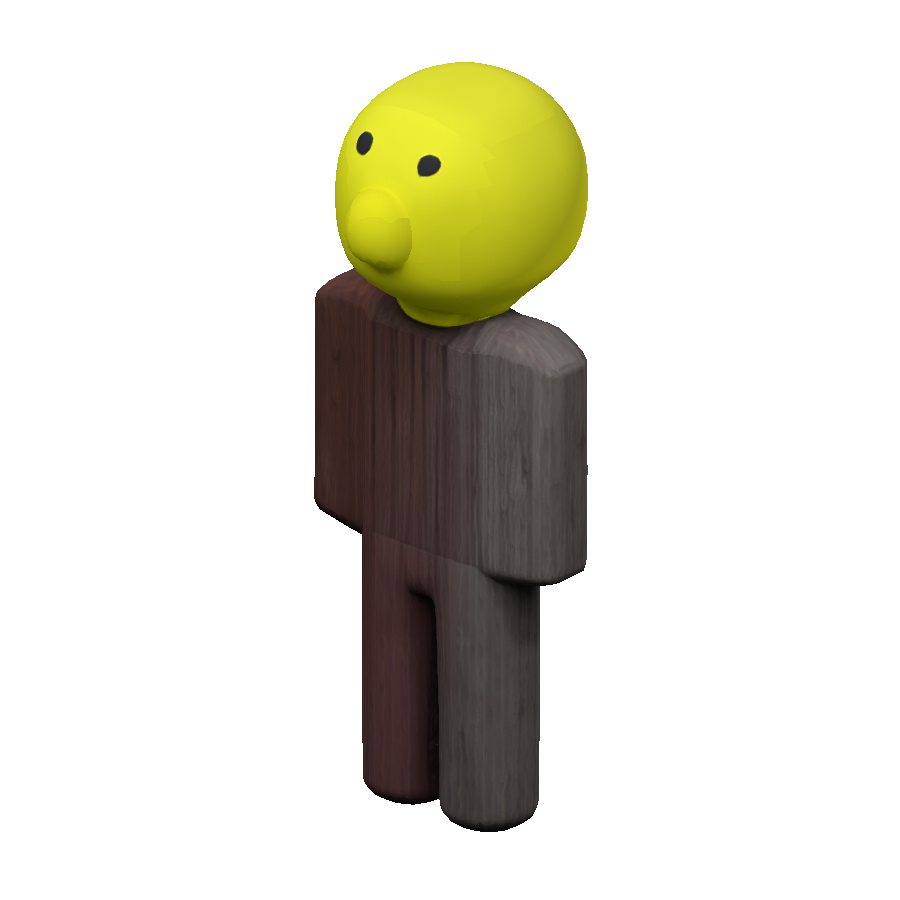}{A wooden toy of a person with a yellow plastic head}
    
    \vspace{0.9em}

    \expcell{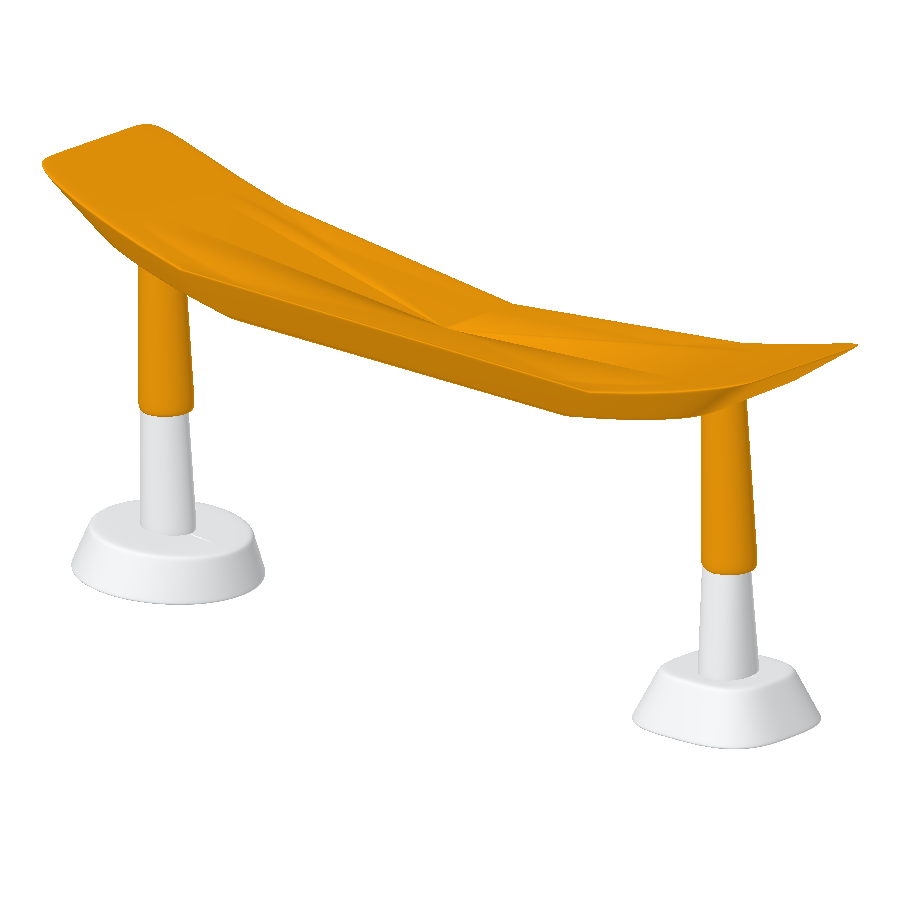}{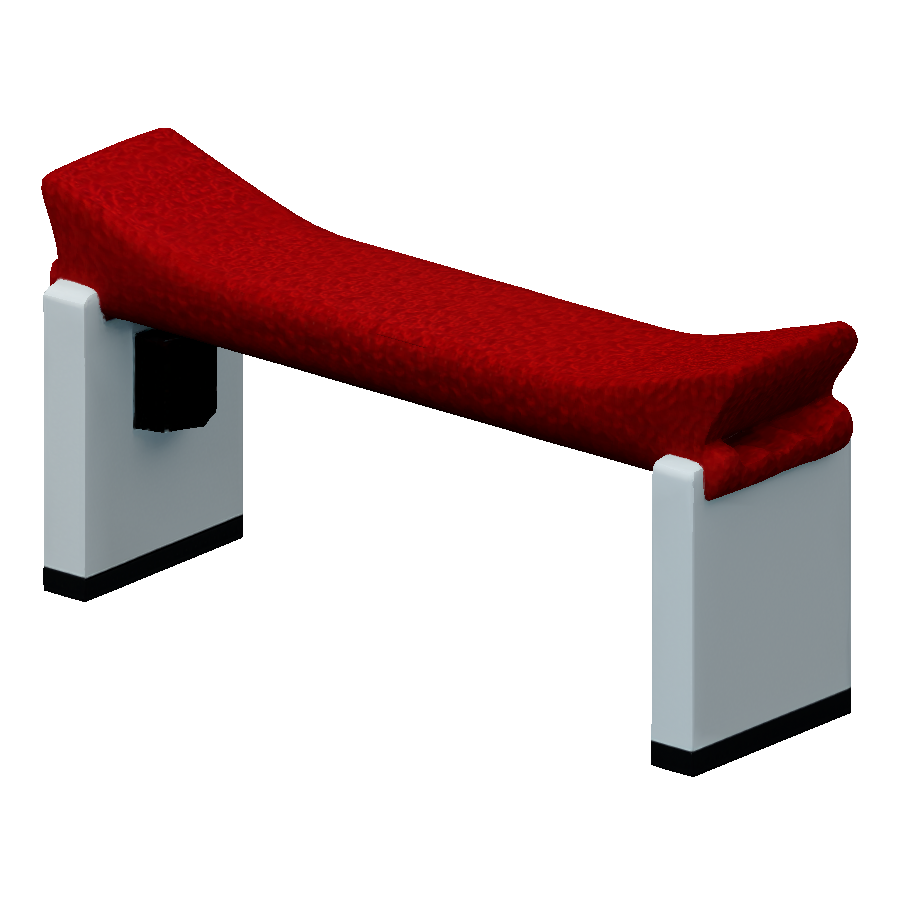}{A white bench with a red velvet seat and a black bottom on the legs}\hfill
    \expcell{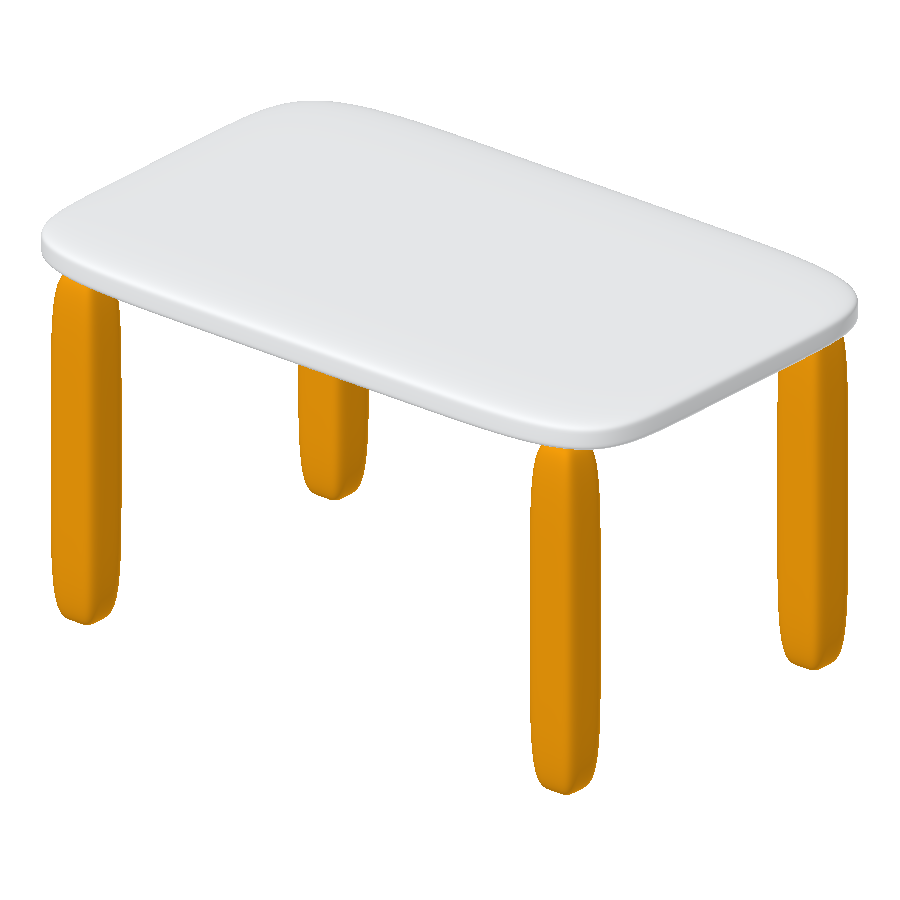}{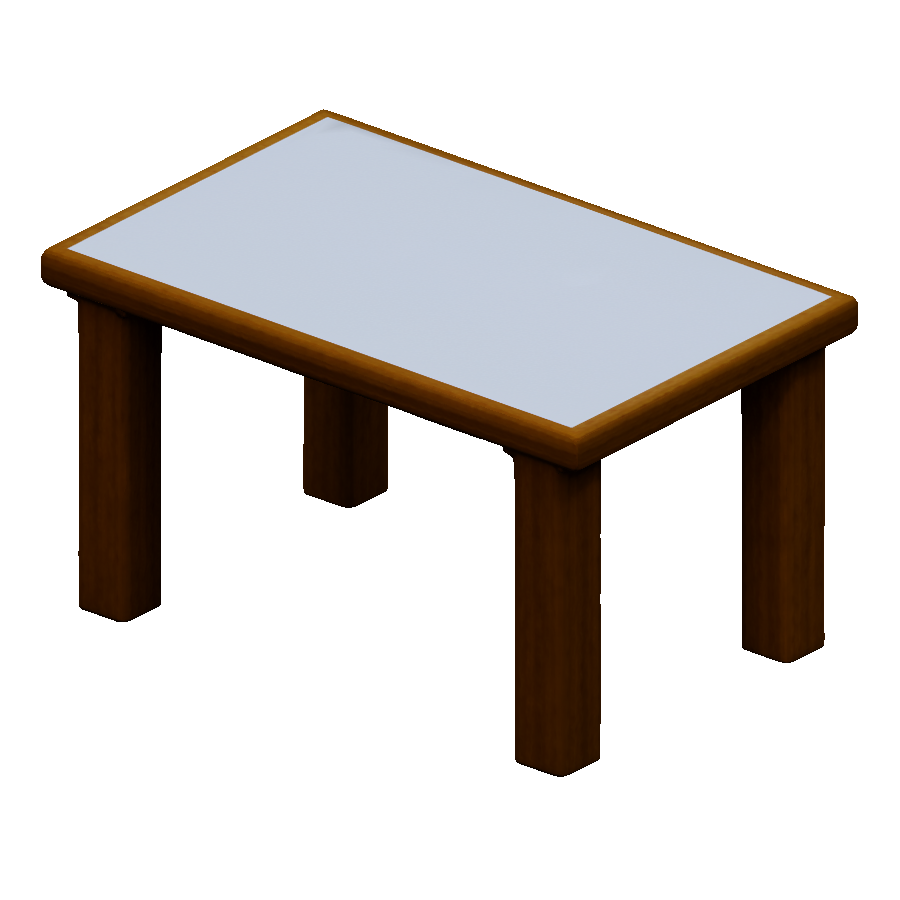}{A glass table with wooden legs}

    \vspace{0.9em}

    \expcell{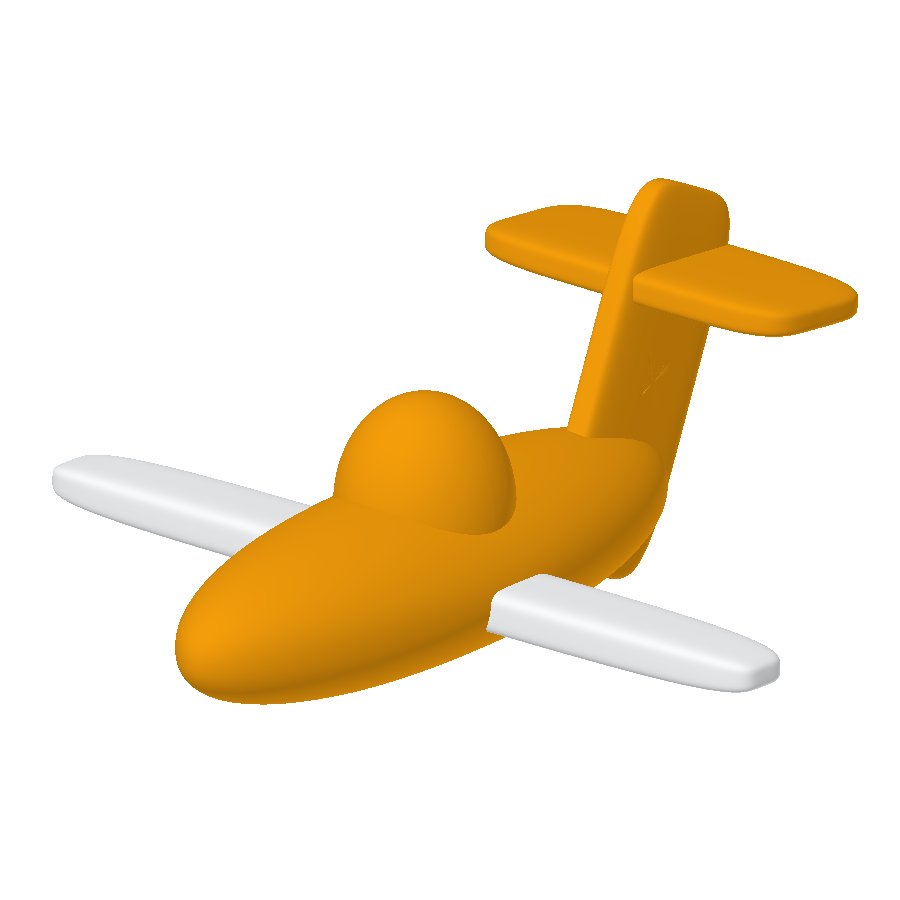}{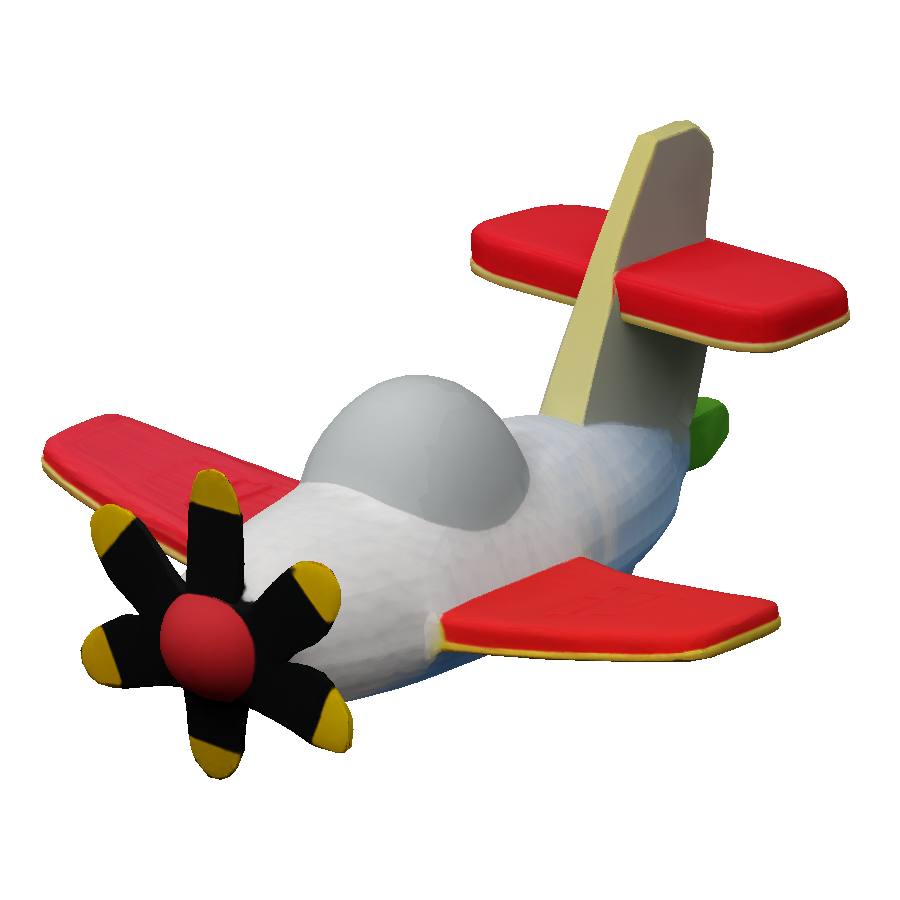}{A plane with the main wings red}\hfill
    \expcell{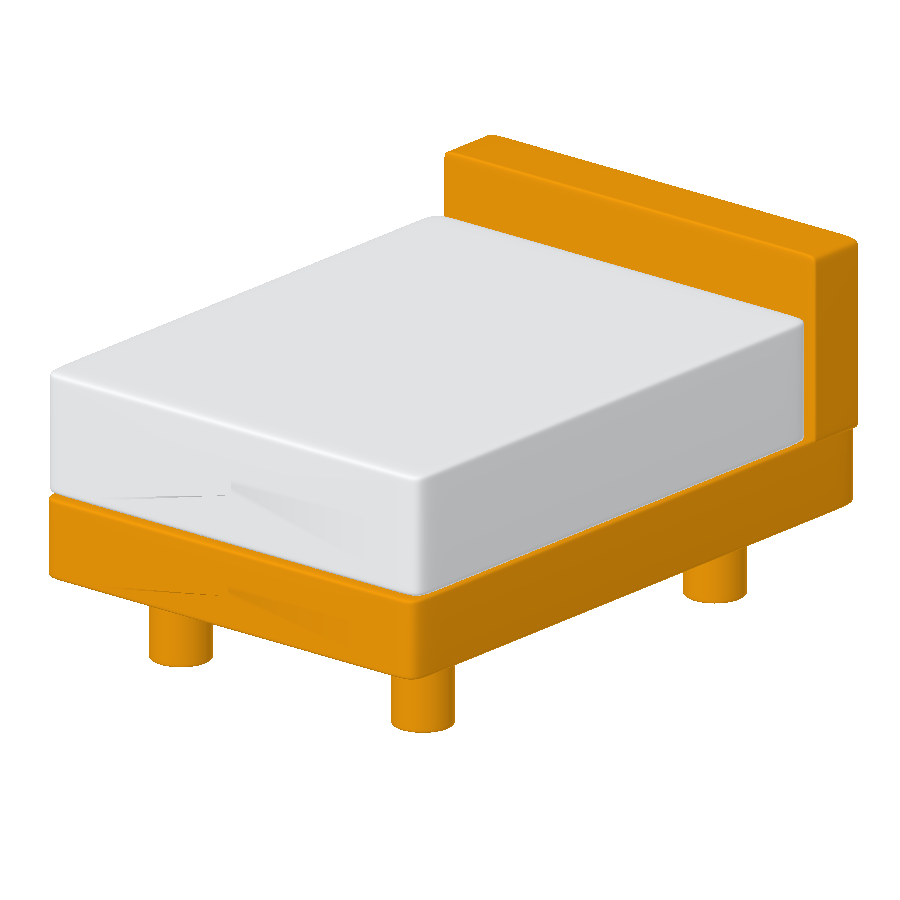}{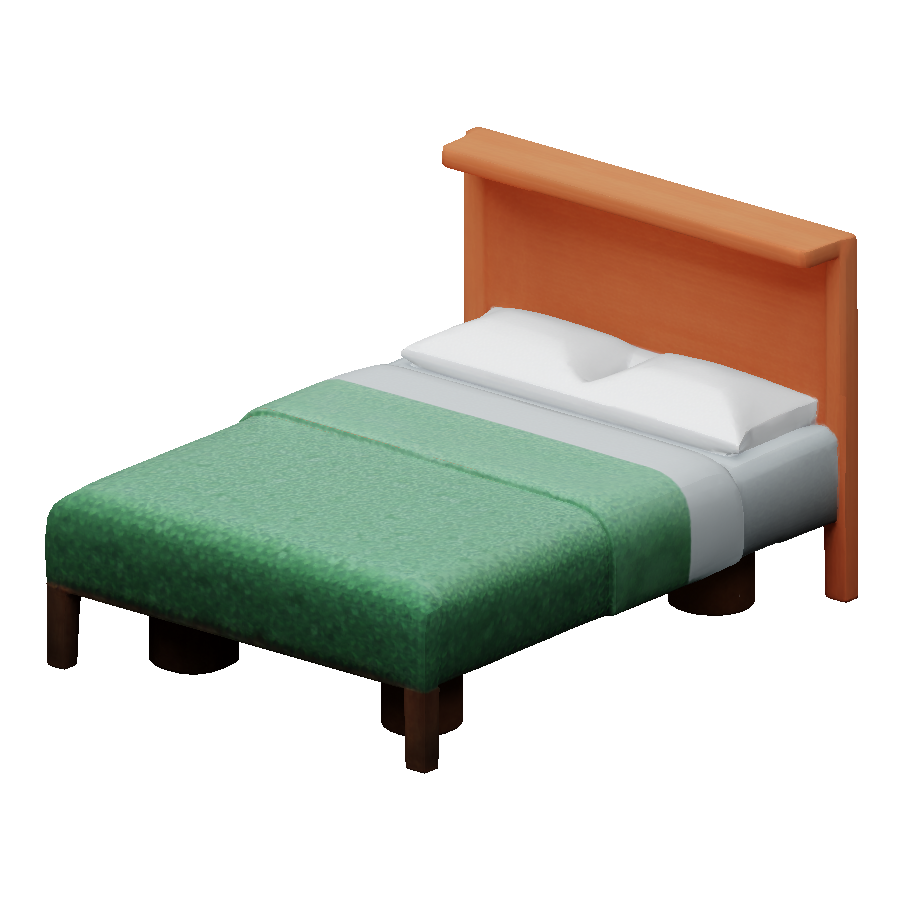}{A bed with a green velvet duvet, pillows and an oak bed frame}

    \caption{
        \textbf{Extended qualitative gallery (Part 1).} Text-conditioned \methodname
        results across diverse object categories. Each cell pairs the input
        superquadric scaffold (left; orange = high-control, gray = low-control
        regions) with the generated asset (right) rendered from the same
        viewpoint.
    }
    \label{fig:supp_more_qualitative_1}
\end{figure}

\begin{figure}[!htbp]
    \centering
    \newcommand{\expcell}[3]{%
        \begin{subfigure}[t]{0.485\textwidth}
            \centering
            \includegraphics[width=0.49\linewidth]{figs/table_experiments_sq/#1}%
            \hfill
            \includegraphics[width=0.49\linewidth]{figs/table_experiments/#2}%
            \caption{``#3''}
        \end{subfigure}%
    }

    \expcell{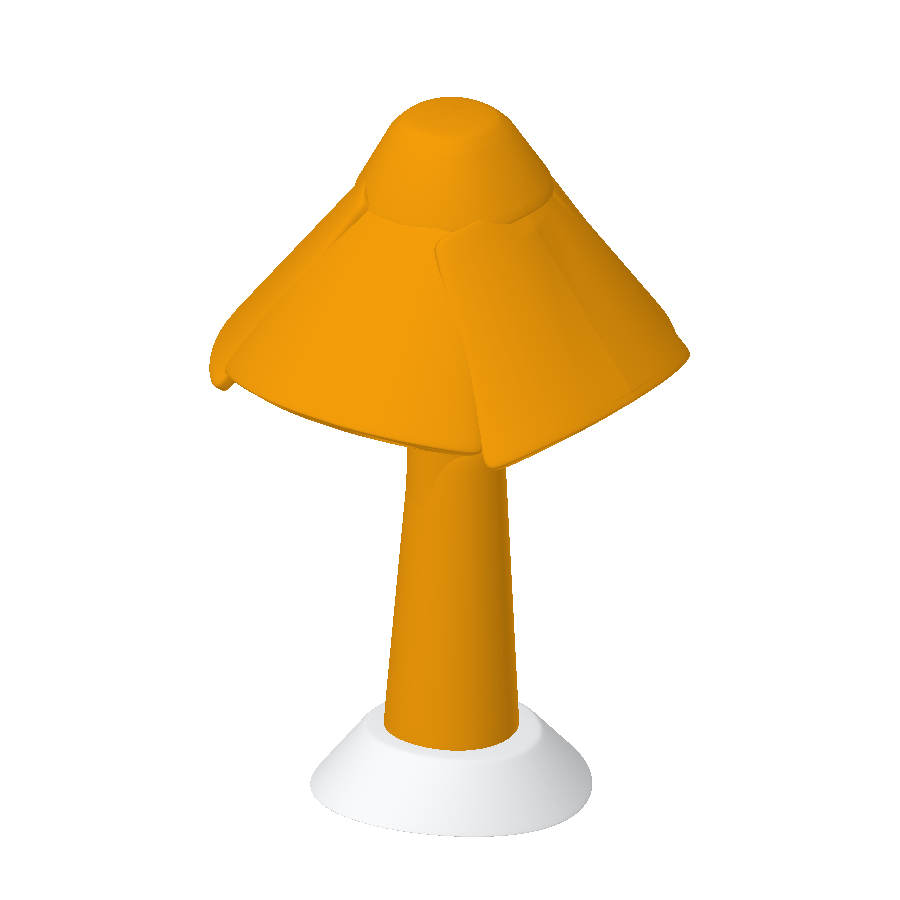}{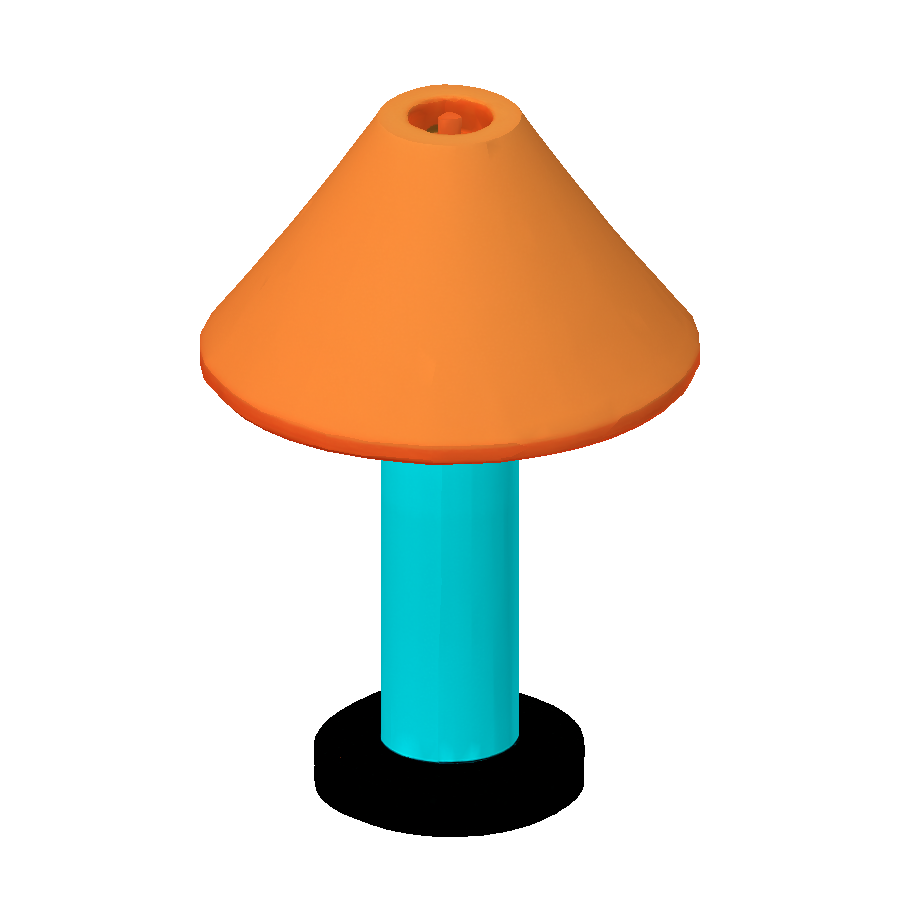}{orange lamp with a cyan pole and a rectangular black base}\hfill
    \expcell{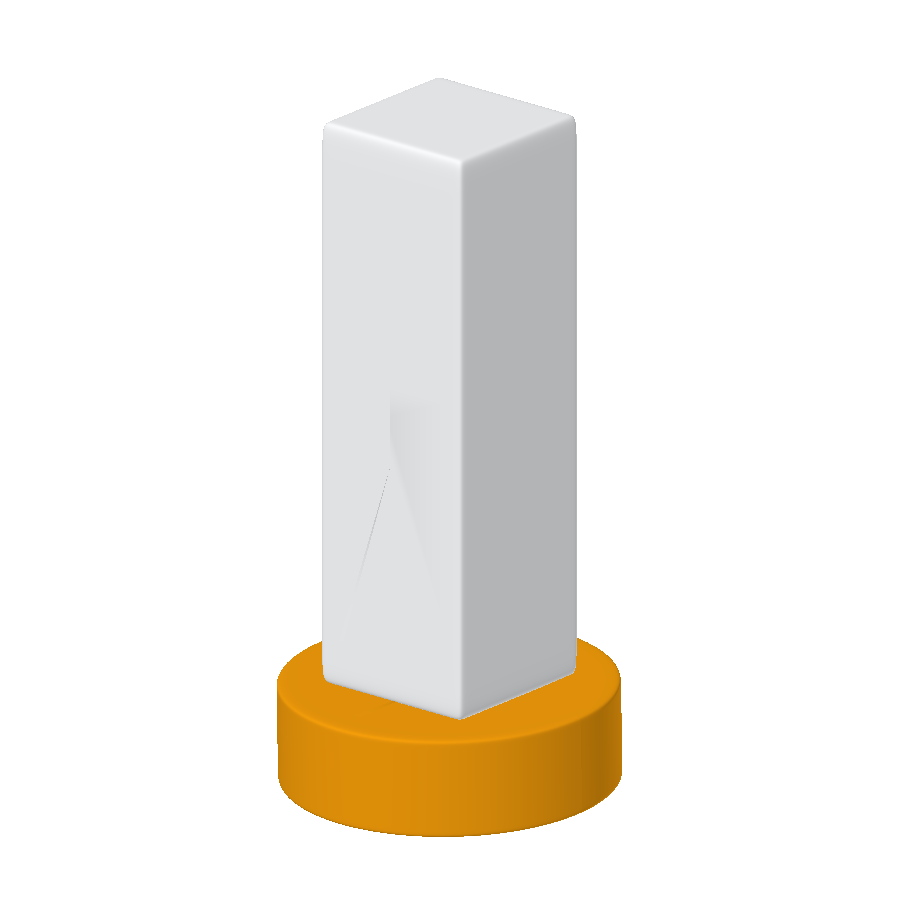}{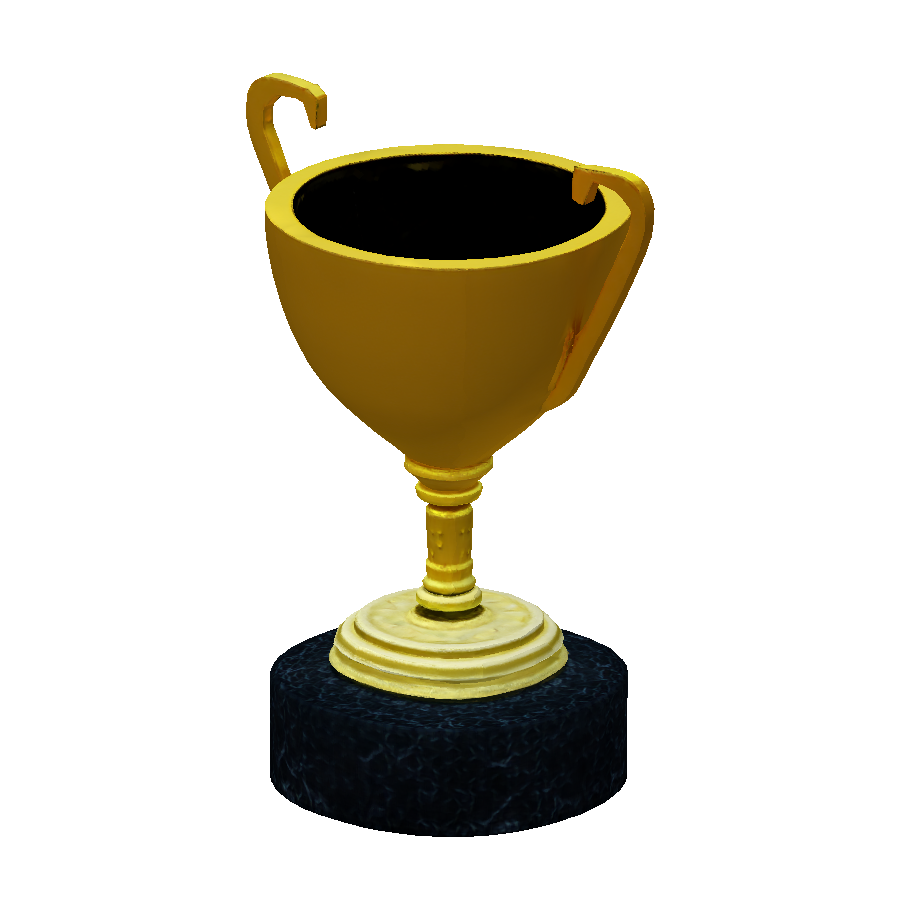}{A gold trophy with a black marble base}

    \vspace{0.9em}

    \expcell{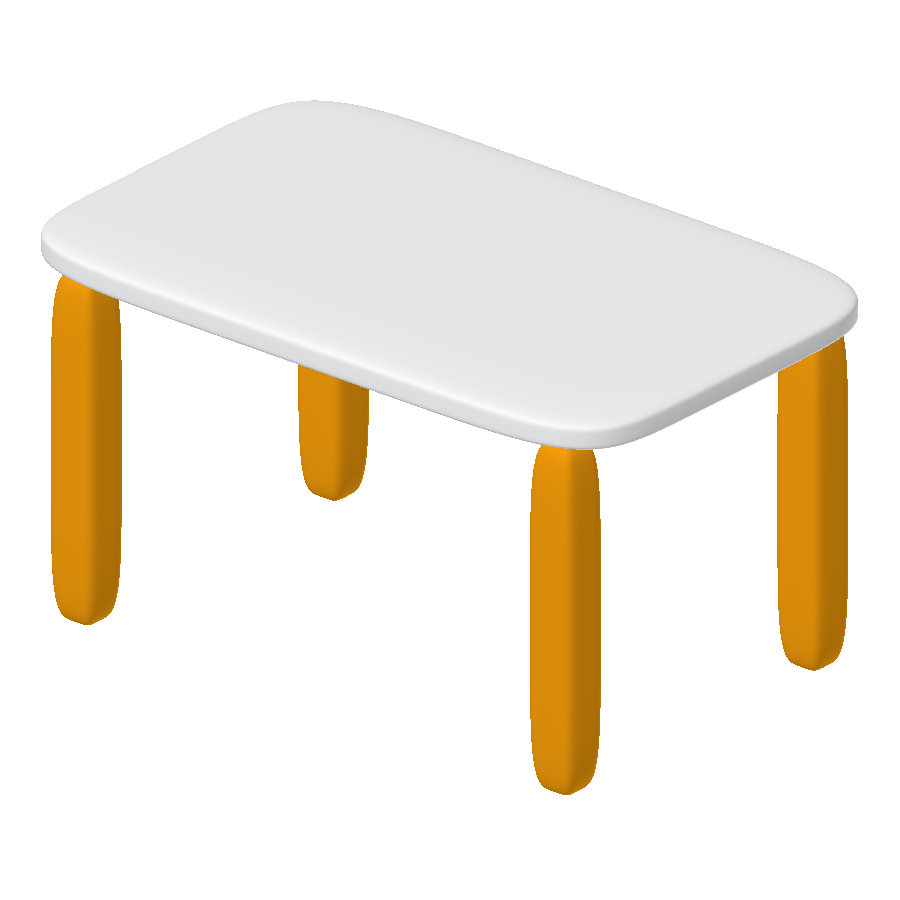}{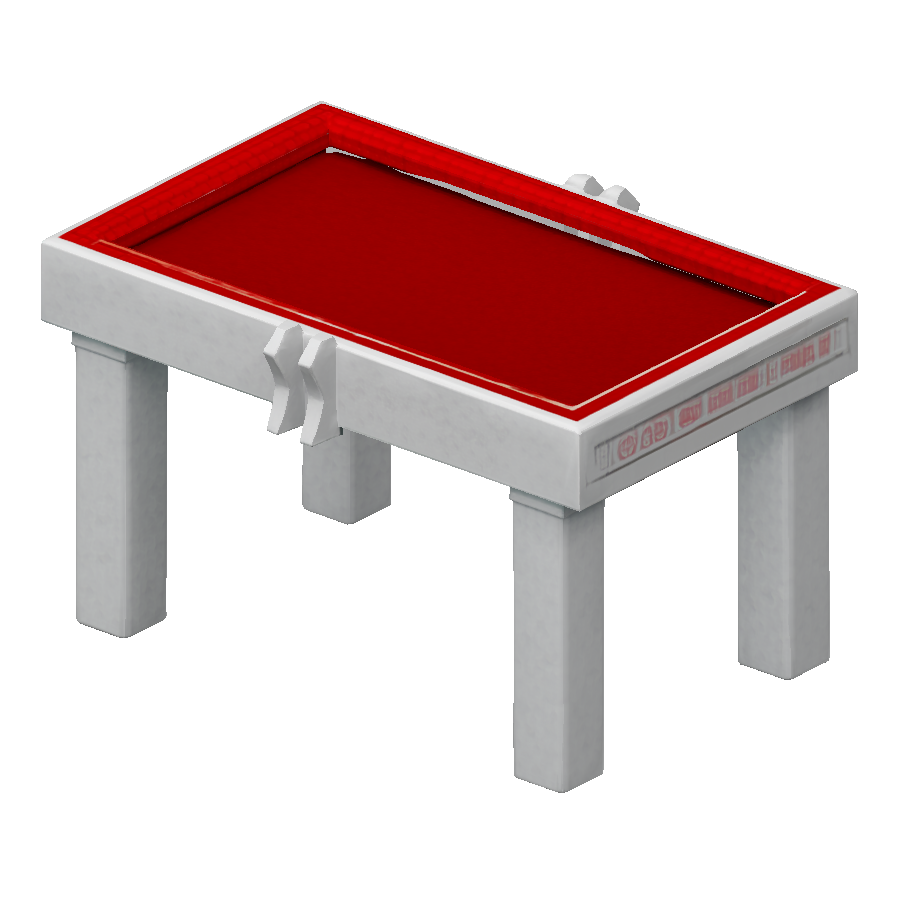}{A white marble pool table with red billiard cloth}\hfill
    \expcell{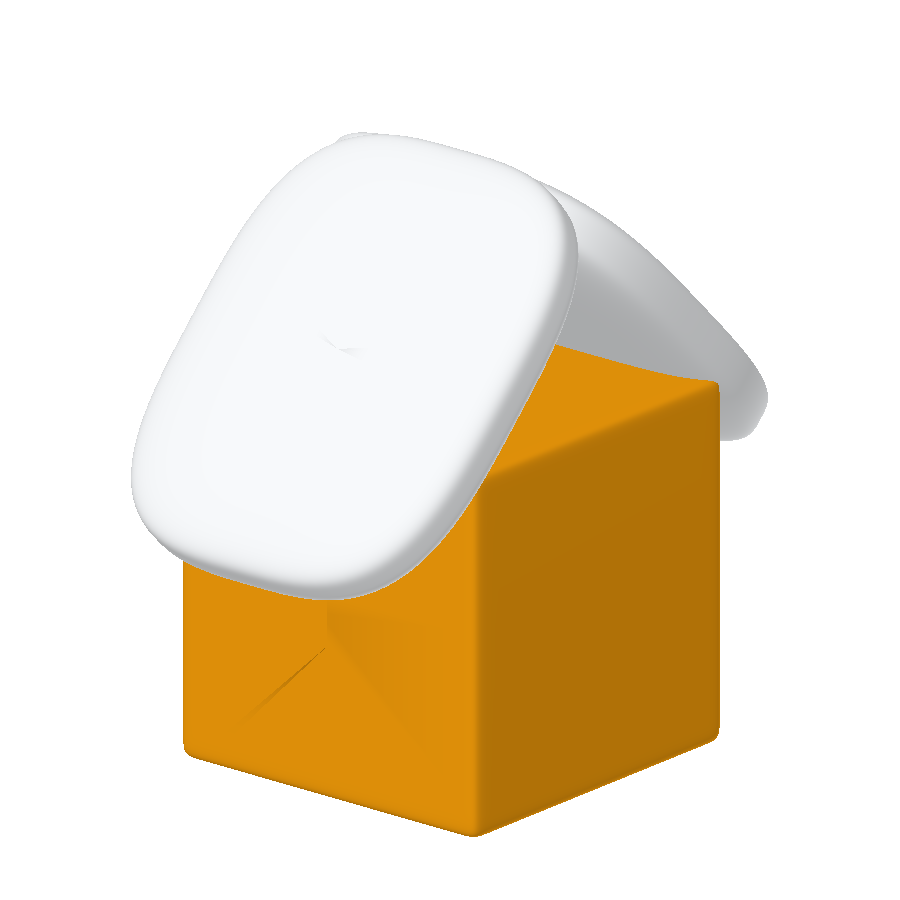}{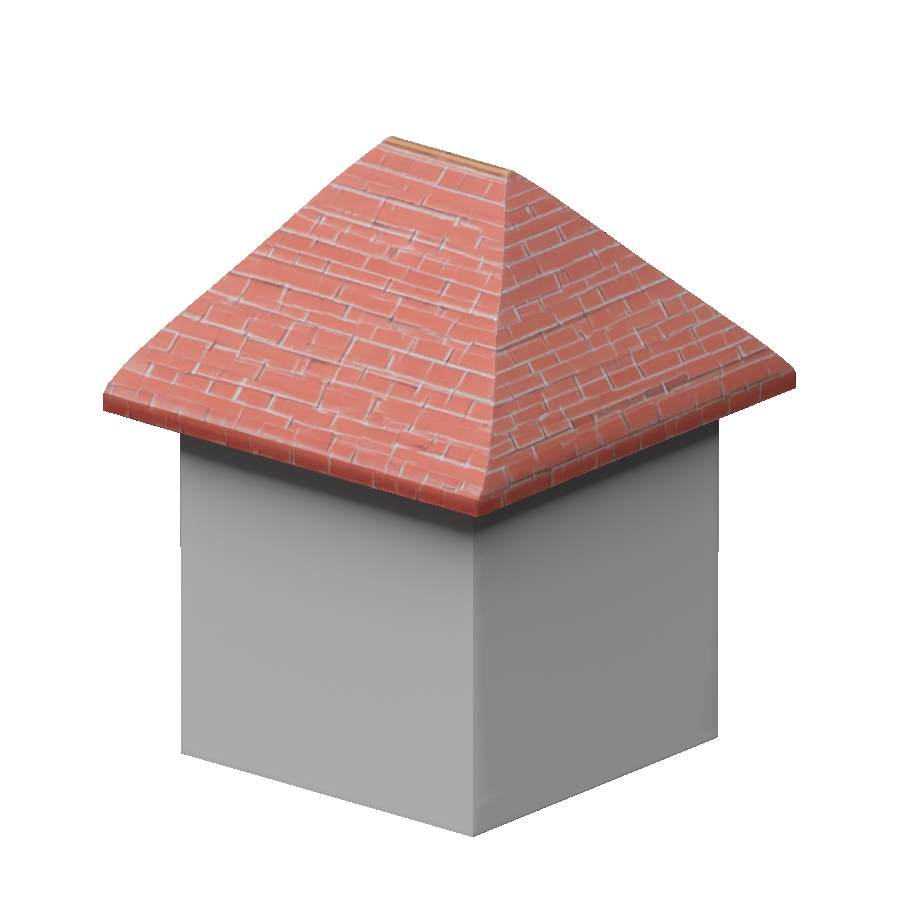}{A white house with a triangular brick roof}

    \vspace{0.9em}

    \expcell{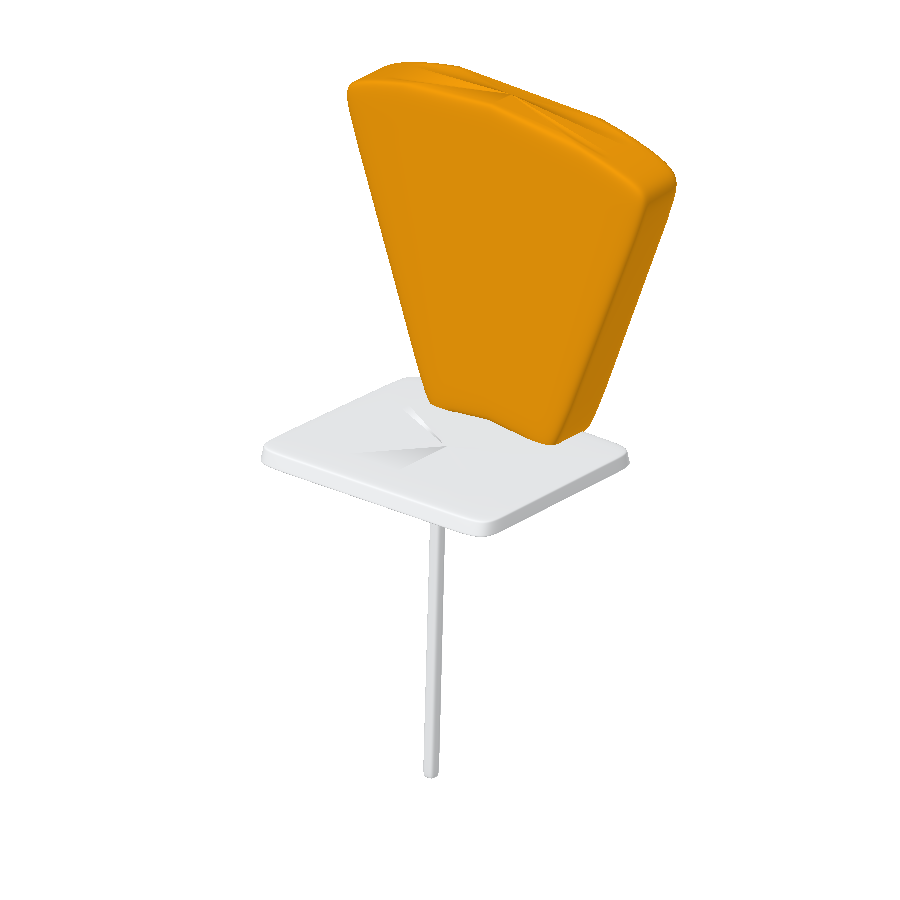}{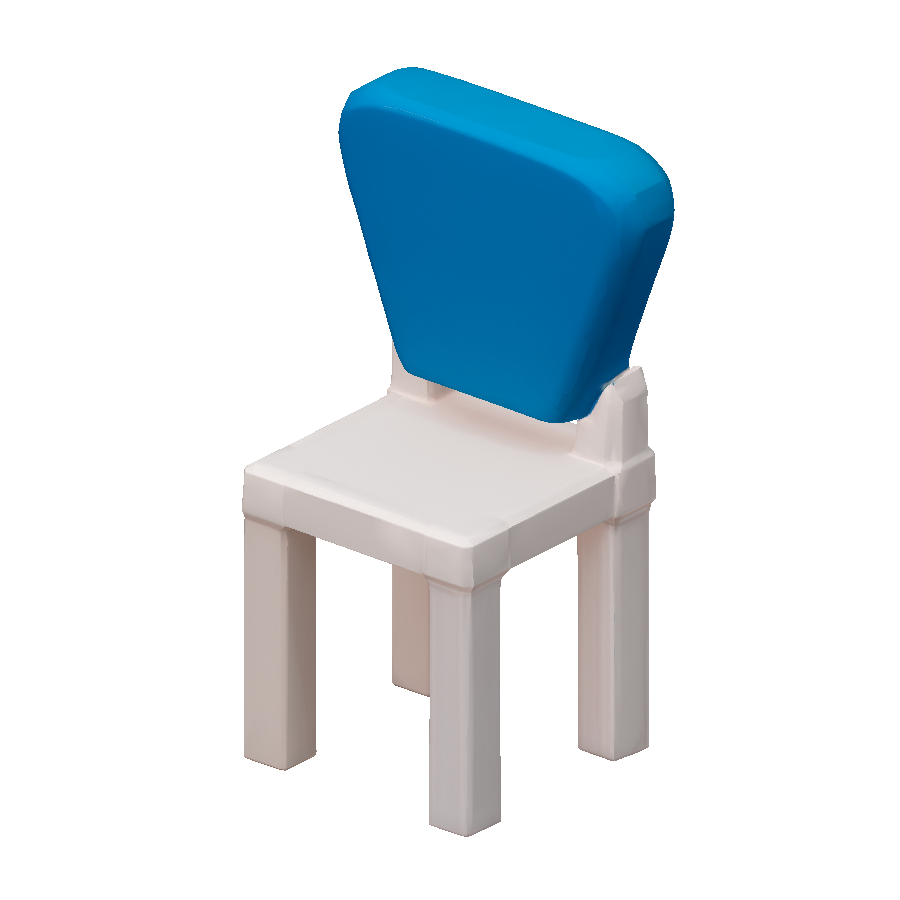}{white chair with a blue backrest}\hfill
    \expcell{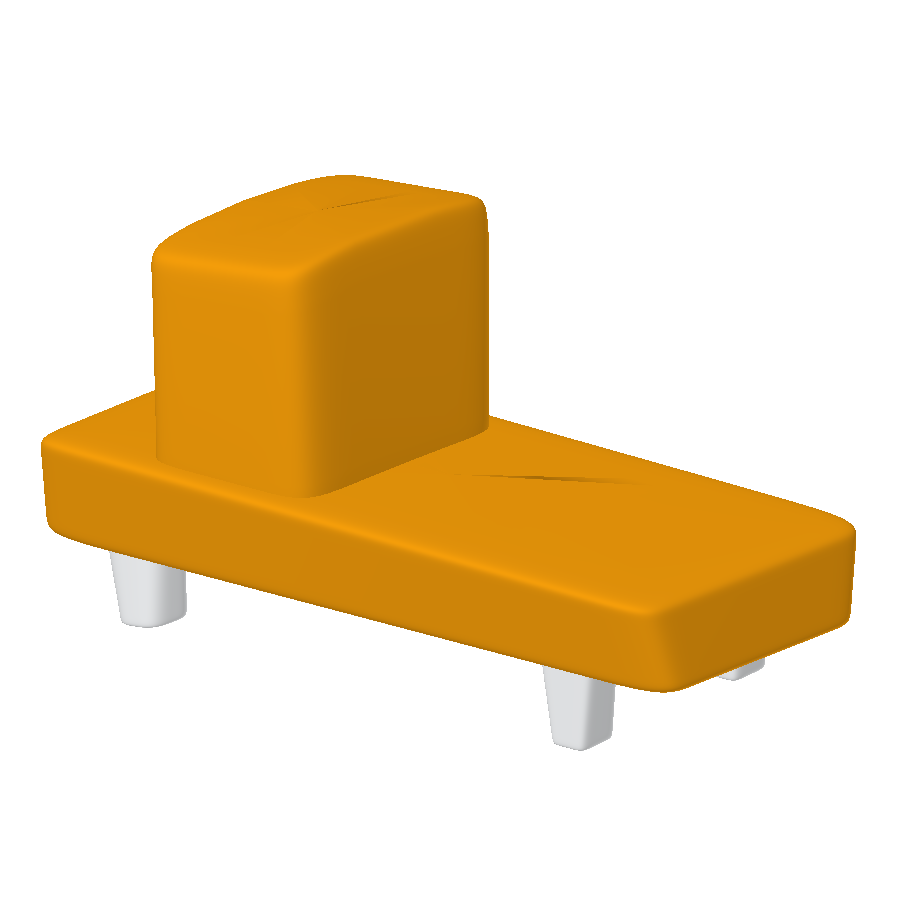}{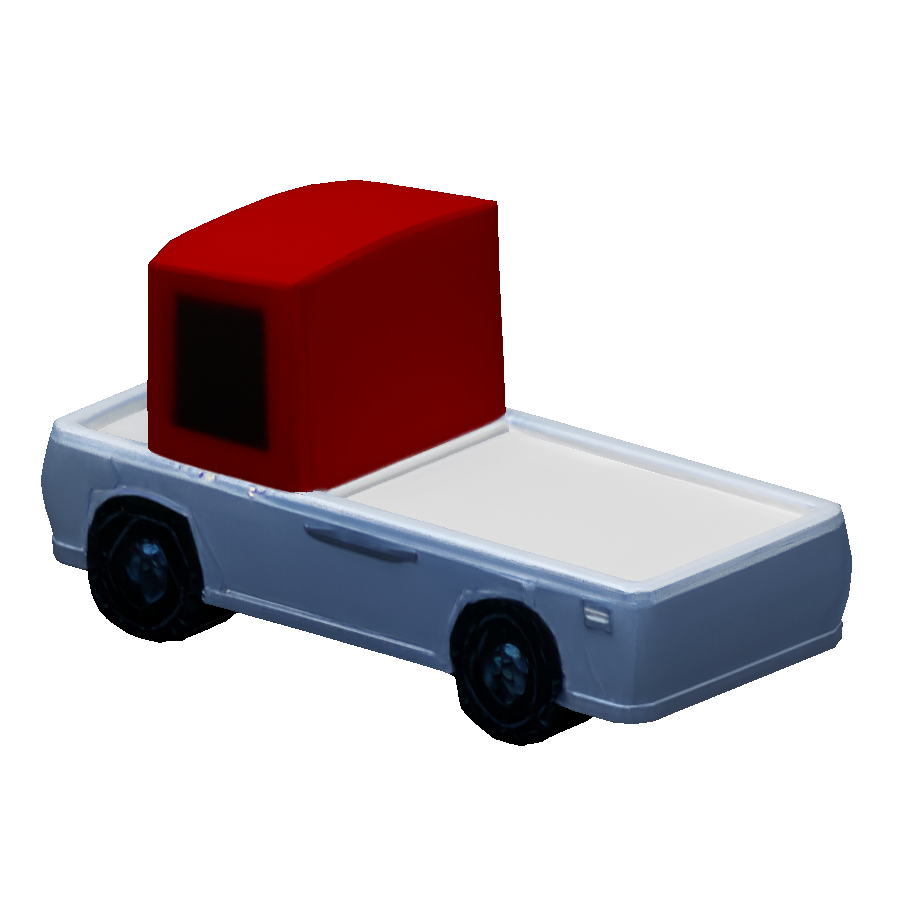}{A silver pick up truck with a red cab and red wheels}

    \vspace{0.9em}

    \expcell{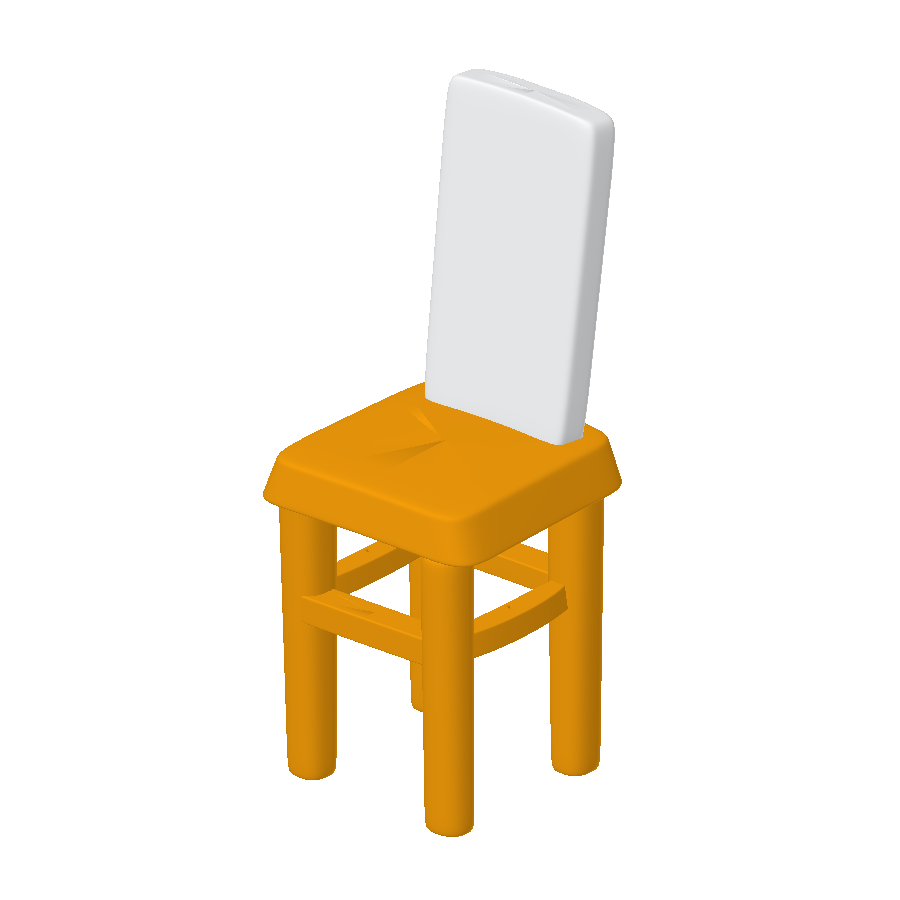}{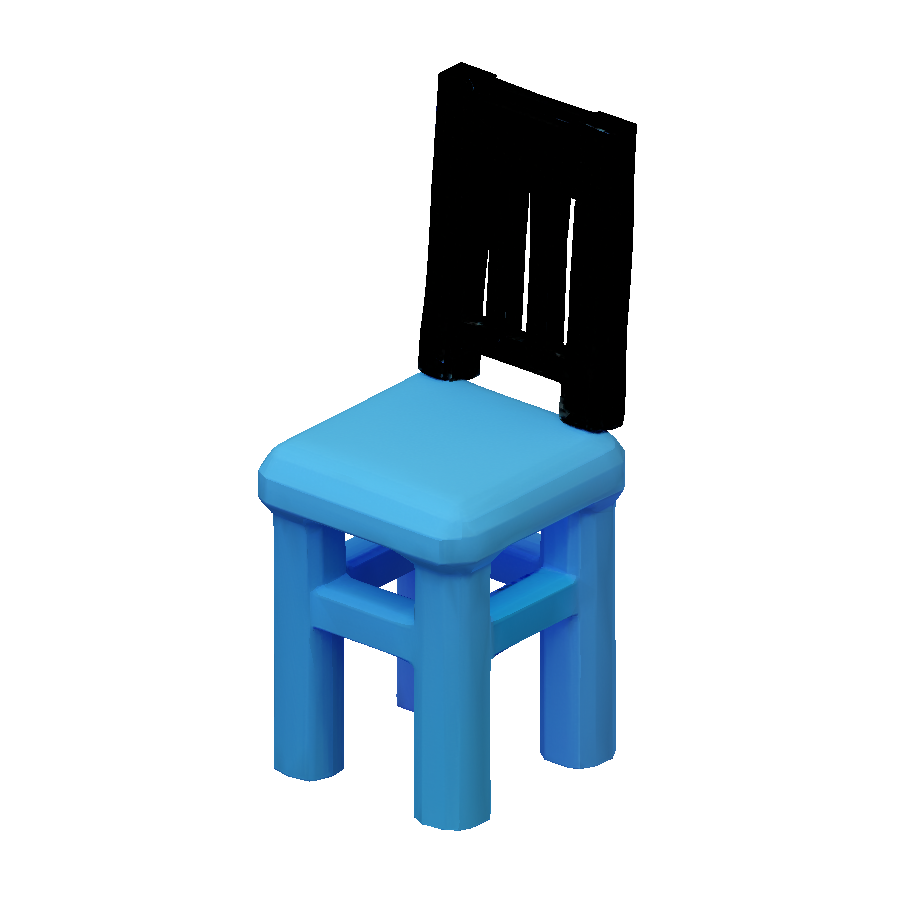}{blue classic chair with a black slat back}\hfill
    \expcell{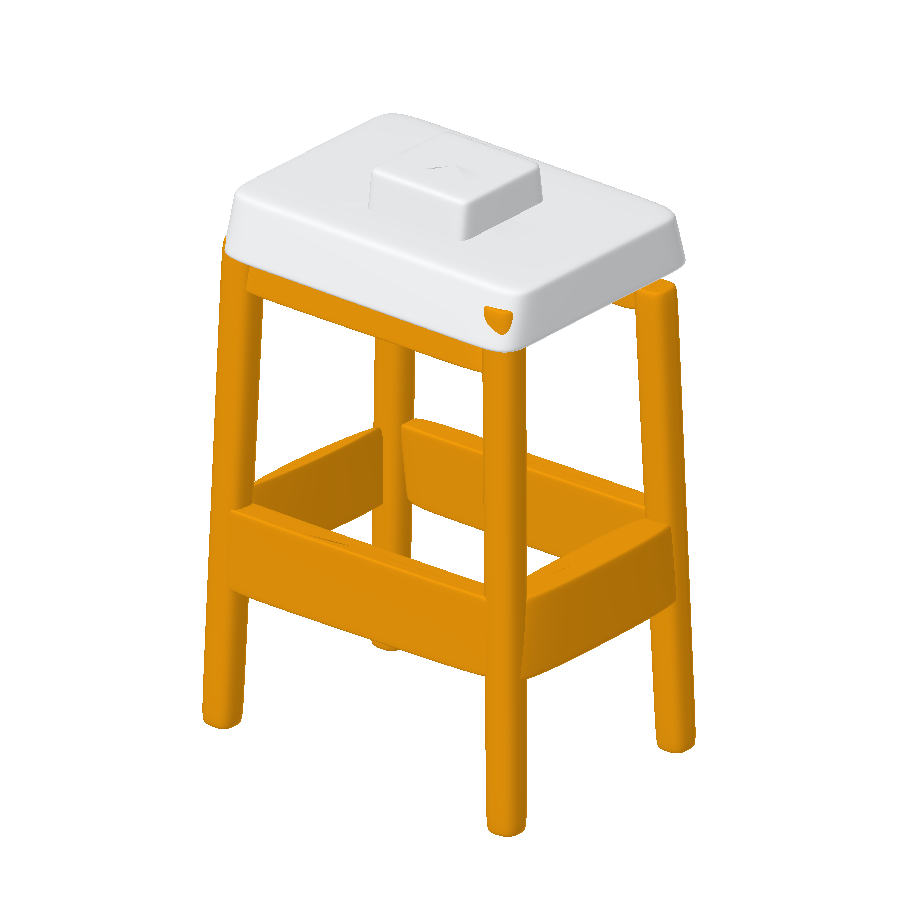}{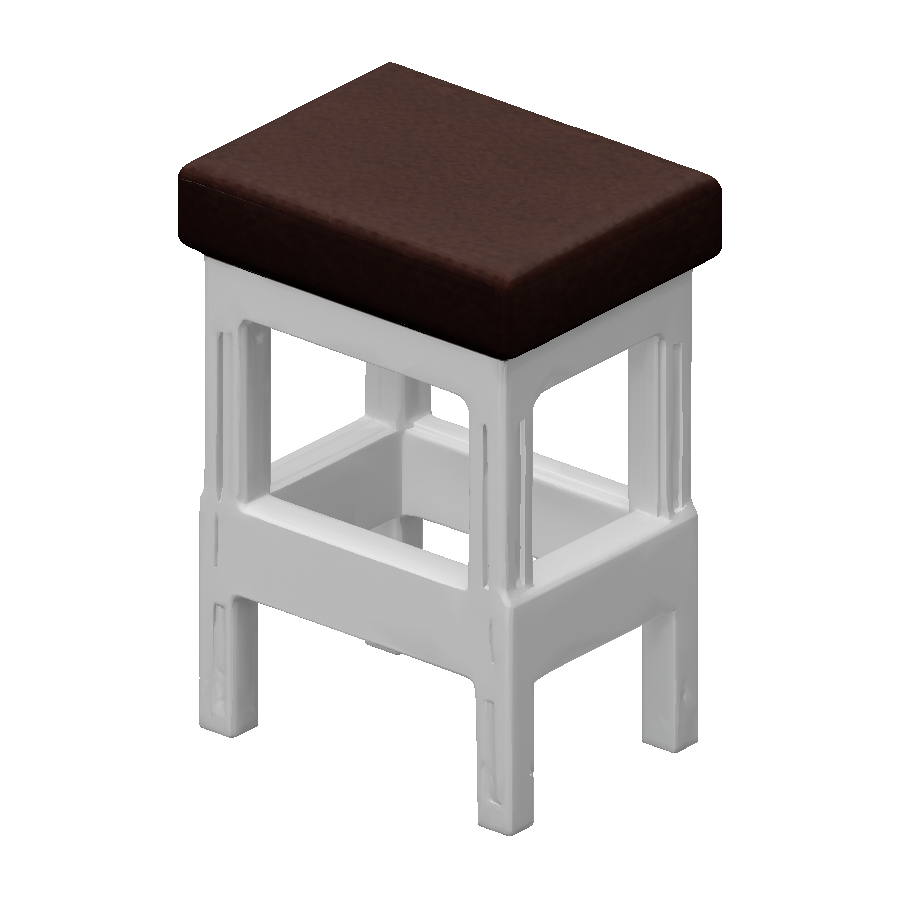}{A brown leather stool with white horizontal slats connecting the legs}

    \caption{
        \textbf{Extended qualitative gallery (Part 2).} Additional text-conditioned
        \methodname results. Each cell pairs the input superquadric scaffold
        (left; orange = high-control, gray = low-control regions) with the
        generated asset (right) rendered from the same viewpoint.
    }
    \label{fig:supp_more_qualitative_2}
\end{figure}

Additionally, Figs.~\ref{fig:supp_more_qualitative_1} and~\ref{fig:supp_more_qualitative_2} show a gallery of text-conditioned qualitative examples of \methodname spanning diverse object categories, varying numbers of geometric primitives, and different localized color and material requests. For each object, we pair the input geometric structure (with high-control regions in orange and low-control regions in gray) with the generated asset rendered from the same viewpoint. For interactive examples, we refer to the attached HTML gallery, which includes a broader set of examples of joint geometric and appearance control. In addition to that, we share an accompanying video that demonstrates the complete workflow and control capabilities in action using our own UI.

\section{Implementation Details}
\label{sec:supp_implementation}

Our implementation is developed in PyTorch  and builds upon the SpaceControl\footnote{\url{https://github.com/spacecontrol3d/spacecontrol}} and GuideFlow3D\footnote{\url{https://github.com/GradientSpaces/GuideFlow3D}}
codebases. We use the publicly released TRELLIS-XL\footnote{\url{https://github.com/microsoft/TRELLIS}} for all pretrained encoders, flow transformers, and decoders, which remain frozen because \methodname is training-free. CLIP\footnote{\url{https://github.com/openai/CLIP}} and DINOv2\footnote{\url{https://github.com/facebookresearch/dinov2}} encode the
text and reference-image conditions, respectively. All experiments were conducted on a single NVIDIA GeForce RTX 5060 Ti GPU. The structure and appearance models were evaluated using the inference settings reported in Tab.~\ref{tab:supp_structure_settings} and \ref{tab:supp_appearance_settings}.

\begin{figure}[!htbp]
\footnotesize
\begin{promptbox}
(*@\prompttag{[SYSTEM INSTRUCTION]}@*)
You are an expert judge of locally controlled 3D shape generation. First you are shown an (*@\prompthighlight{INPUT~CONTROL~SHAPE}@*) built from coarse primitives. Its colours are region LABELS, not a texture target:
  - (*@\prompthighlight{ORANGE}@*) marks high-control regions. The generated object must follow these parts closely in (*@\prompthighlight{SHAPE}@*).
  - (*@\prompthighlight{WHITE~/~GREY}@*) marks weak-control regions. The generator is free to reshape these into whatever the text prompt needs, and doing so is the (*@\prompthighlight{DESIRED}@*) behaviour, not an error.

Then you are shown the generated object's (*@\prompthighlight{STRUCTURE~ONLY}@*): the raw generated geometry, rendered untextured as a plain uniform grey shape. It carries no colour or material of its own, so the orange/white labels of the control shape have no counterpart in it and you must never look for one. Work out which part of the output corresponds to which part of the control shape from (*@\prompthighlight{POSITION~and~GEOMETRY}@*) alone.

The two region types are judged by opposite standards, and this is the single most important thing to get right:
  - An orange region that changed shape is a (*@\prompthighlight{FAILURE}@*).
  - A white/grey region that changed shape is a (*@\prompthighlight{SUCCESS}@*). The whole point is that the generator turns those coarse blobs into real, detailed parts. Never treat 'the output no longer looks like the input primitives here' as an error in a white/grey region.

Judge (*@\prompthighlight{GEOMETRY~ONLY}@*): proportions, part placement, silhouette, structural plausibility. Completely (*@\prompthighlight{IGNORE}@*) colour, material, texture sharpness, lighting, shadows and background. You are shown two rendered views of each object. They are the (*@\prompthighlight{SAME~object}@*) from two camera angles roughly 210 degrees apart, (*@\prompthighlight{NOT~two~different~objects}@*). A feature may therefore be visible in only one of the two views; judge the object as a whole and never penalise a feature merely for being out of frame in the other view.

Sample A and Sample B were generated from the (*@\prompthighlight{SAME~prompt}@*) and the (*@\prompthighlight{SAME~input~control~shape}@*). Compare them on geometry alone.

For each question choose (*@\prompthighlight{exactly~one}@*) of: (*@\prompthighlight{A, B, Equal}@*). Use Equal only when there is genuinely no meaningful difference - do not use it to avoid deciding.

- q1_control_prompt: Which object best matches the prompt in the white/grey weak-control regions while also closely following the orange high-control input shape?
- q2_realism: Which object looks more realistic as a real-world object?
- q3_overall: Which object is better overall as a controlled 3D generation?

Return ONLY a single JSON object, no prose, no markdown fences, with exactly these string keys:
{"q1_control_prompt": "A|B|Equal", "q2_realism": "A|B|Equal", "q3_overall": "A|B|Equal"}

(*@\prompttag{[USER DATA]}@*)
Target description: [Shape Prompt]
INPUT CONTROL SHAPE (two views):
[Image: Superquadric Proxy View 1] [Image: Superquadric Proxy View 2]
Sample A (two views), untextured grey geometry: 
[Image: Sample A View 1] [Image: Sample A View 2]
Sample B (two views), untextured grey geometry: 
[Image: Sample B View 1] [Image: Sample B View 2]
\end{promptbox}
\vspace{-3mm}
\caption{\textbf{Prompt template for pairwise structure evaluation.} The judge evaluates two untextured geometries against the input control shape, with randomized A/B sample ordering across passes.}
\label{fig:supp_structure_h2h_prompt}
\end{figure}

\begin{figure}[!htbp]
\begin{promptbox}
(*@\prompttag{[SYSTEM INSTRUCTION]}@*)
You are an expert judge of 3D appearance and texture generation.

You are shown a generated 3D object that was textured to satisfy a (*@\prompthighlight{TARGET~DESCRIPTION}@*). The description may name different materials for different parts; getting each material onto the correct part is the central thing being tested.

Judge (*@\prompthighlight{APPEARANCE~ONLY}@*): colour, material, and how cleanly the texture sits on the surface. Completely (*@\prompthighlight{IGNORE}@*) the underlying geometry, the object's proportions, lighting, shadows, background and camera angle. If the shape itself is odd, that is not an appearance failure. You are shown two rendered views of each object. They are the (*@\prompthighlight{SAME~object}@*) from two camera angles roughly 210 degrees apart, (*@\prompthighlight{NOT~two~different~objects}@*). A feature may therefore be visible in only one of the two views; judge the object as a whole and never penalise a feature merely for being out of frame in the other view.

Sample A and Sample B were generated from the (*@\prompthighlight{SAME~prompt}@*). They also share the same underlying geometry, so do not look for shape differences - look at colour, material and texture.

For each question choose (*@\prompthighlight{exactly~one}@*) of: (*@\prompthighlight{A, B, Equal}@*). Use Equal only when there is genuinely no meaningful difference - do not use it to avoid deciding.

- q1_prompt_faithfulness: Which object's appearance better matches the target description, including getting the (*@\prompthighlight{right~material~onto~the~right~part}@*)?
- q2_texture_quality: Which object has the cleaner, sharper, less artifacted surface appearance?
- q3_overall: Considering both faithfulness to the requested appearance and surface texture quality, which object is preferred overall?

Return (*@\prompthighlight{ONLY}@*) a single JSON object, no prose, no markdown fences, with exactly these string keys:
{"q1_prompt_faithfulness": "A|B|Equal", "q2_texture_quality": "A|B|Equal", "q3_overall": "A|B|Equal"}

(*@\prompttag{[USER DATA]}@*)
Target description: [Global object prompt and part-specific requests]
Sample A (two views):
[Image: Sample A Textured View 1] [Image: Sample A Textured View 2]
Sample B (two views):
[Image: Sample B Textured View 1] [Image: Sample B Textured View 2]
\end{promptbox}
\vspace{-3mm}
\caption{\textbf{Prompt template for pairwise appearance evaluation.} Blinded comparison of textured outputs on identical geometry to measure prompt faithfulness and surface quality.}
\label{fig:supp_appearance_h2h_prompt}
\end{figure}

\begin{figure}[!htbp]
\begin{promptbox}
(*@\prompttag{[SYSTEM INSTRUCTION]}@*)
You are an expert judge of 3D texture / appearance transfer. You are shown two rendered views of a single textured 3D object (*@\prompthighlight{SAME~OBJECT}@*) from different camera angles, not different objects. The two views look at (*@\prompthighlight{OPPOSITE sides}@*), so a textured region may appear in only ONE view -- judge the object as a whole and do NOT penalise a feature merely for being out of frame in the other view. Base your scores on colour, material and pattern transfer onto the correct parts; (*@\prompthighlight{IGNORE}@*) lighting, shadows, background and which side faces the camera. You may also first be shown a 'structure reference': clean, untextured renders of the same geometry from the same angles, provided only so you can recognise the object's parts -- (*@\prompthighlight{do NOT score the geometry itself}@*). Imagine the full 3D object and judge how the texture behaves across its surfaces.

The texture was generated to satisfy a (*@\prompthighlight{TEXT}@*) description. Judge how well the rendered object matches that description.

Score (*@\prompthighlight{EACH}@*) of the following criteria as an integer from 1 to 10, using the (*@\prompthighlight{FULL}@*) range and these anchors: 
    1-2  = the target colour/material/pattern is absent or plainly wrong; 
    3-4  = barely present or mostly wrong;
    5-6  = roughly right but with clear, obvious errors;
    7-8  = clearly matches the target, only minor issues;
    9-10 = near-perfect match.

Judge the object on its own merits, not relative to anything. Avoid defaulting to the extremes (all 1s or all 9s) -- reserve those for cases that truly warrant them.

1. prompt_faithfulness: How well the texture matches what the prompt asked for, (*@\prompthighlight{INCLUDING per-part / local requests}@*) (e.g. 'orange backrest', 'red cushion'). Penalise missing or misplaced local colours/materials.
2. color_material_accuracy: Correctness of colours and material feel (matte / glossy / wood / metal / fabric) relative to the prompt.
3. texture_detail_quality: Sharpness and cleanliness of local texture detail; absence of noise, blur, smearing or obvious artifacts.
4. overall: Overall texture quality considering all of the above.

Return (*@\prompthighlight{ONLY}@*) a single JSON object, no prose, no markdown fences, with exactly these integer keys:
{"prompt_faithfulness": <1-10>, "color_material_accuracy": <1-10>, "texture_detail_quality": <1-10>, "overall": <1-10>}

(*@\prompttag{[USER DATA]}@*)
Target description: [Global object prompt and all part-specific color/material requests]
Structure reference - the untextured geometry the texture was applied to (same shape and camera angles as the output below; for understanding the object's parts, NOT to be scored):
[Image: Untextured Render View 1] [Image: Untextured Render View 2]
Rendered output view(s) of the result:
[Image: Textured Render View 1] [Image: Textured Render View 2]
\end{promptbox}
\vspace{-3mm}
\caption{\textbf{Prompt template for text-conditioned standalone VLM evaluation.} Complete system instructions and input layout used to score part-wise text appearance generation.}
\label{fig:supp_text_prompt}
\end{figure}

\begin{figure}[!htbp]
\begin{promptbox}
(*@\prompttag{[SYSTEM INSTRUCTION]}@*)
You are an expert judge of 3D texture / appearance transfer. You are shown one or two rendered views of a single textured 3D object (the (*@\prompthighlight{SAME~object}@*) from different camera angles, not different objects). The two views look at (*@\prompthighlight{OPPOSITE~sides}@*), so a textured region may appear in only ONE view -- judge the object as a whole and do NOT penalise a feature merely for being out of frame in the other view. Base your scores on colour, material and pattern transfer onto the correct parts; (*@\prompthighlight{IGNORE}@*) lighting, shadows, background and which side faces the camera. You may also first be shown a 'structure reference': clean, untextured renders of the same geometry from the same angles, provided only so you can recognise the object's parts -- (*@\prompthighlight{do NOT score the geometry itself}@*). Imagine the full 3D object and judge how the texture behaves across its surfaces.

The texture was generated to imitate (*@\prompthighlight{REFERENCE~STYLE~IMAGES}@*), and the (*@\prompthighlight{TARGET~DESCRIPTION}@*) states which reference applies to which part: the (*@\prompthighlight{GLOBAL}@*) reference image gives the overall appearance and each (*@\prompthighlight{LOCAL}@*) reference image gives the style of a specific part. Judge how faithfully the rendered object reproduces that target, using the description and the reference images together (this is also what local_routing_accuracy checks). The reference images show a material/style, (*@\prompthighlight{NOT~the~target~shape}@*) -- ignore the shape in the reference.

Score (*@\prompthighlight{EACH}@*) of the following criteria as an integer from 1 to 10, using the (*@\prompthighlight{FULL}@*) range and these anchors:
  1-2  = the target colour/material/pattern is absent or plainly wrong;
  3-4  = barely present or mostly wrong;
  5-6  = roughly right but with clear, obvious errors;
  7-8  = clearly matches the target, only minor issues;
  9-10 = near-perfect match.
Judge the object on its own merits, not relative to anything. Avoid defaulting to the extremes (all 1s or all 9s) -- reserve those for cases that truly warrant them.

1. style_fidelity: How faithfully the output reproduces the reference style image's colour, material and pattern/motif.
2. local_routing_accuracy: Whether the correct reference texture landed on the correct part of the object (when multiple reference images / per-part references are given).
3. texture_detail_quality: Sharpness and cleanliness of local texture detail; absence of noise, blur, smearing or obvious artifacts.
4. overall: Overall texture quality considering all of the above.

Return (*@\prompthighlight{ONLY}@*) a single JSON object, no prose, no markdown fences, with exactly these integer keys:
{"style_fidelity": <1-10>, "local_routing_accuracy": <1-10>, "texture_detail_quality": <1-10>, "overall": <1-10>}

(*@\prompttag{[USER DATA]}@*)
Target description: [Short text description mapping target parts to the reference styles]
Global reference image (overall appearance): [Image: Global Reference Style]
Local reference image (specific part): [Image: Local Reference Style 1] ...
Structure reference - the untextured geometry the texture was applied to (same shape and camera angles as the output below; for understanding the object's parts, NOT to be scored):
[Image: Untextured Render View 1] [Image: Untextured Render View 2]
Rendered output view(s) of the result:
[Image: Textured Render View 1] [Image: Textured Render View 2]
\end{promptbox}
\vspace{-3mm}
\caption{\textbf{Prompt template for image-conditioned standalone VLM evaluation.} Layout provided to the VLM judge to score local reference image style transfer and spatial routing accuracy.}
\label{fig:supp_image_prompt}
\end{figure}

\end{document}